\documentclass[wcp]{jmlr}

\usepackage{longtable}

\usepackage{booktabs}
\usepackage{comment}

\jmlrvolume{101}
\jmlryear{2019}
\jmlrworkshop{ACML 2019}

\title[High-Resolution Network for Photorealistic Style Transfer]{High-Resolution Network for Photorealistic Style Transfer}

  \author{\Name{Ming Li} \Email{limingcv@gmail.com}\\
  \addr College of Computer and Cybersecurity, Hainan University, Haikou, China 
  \AND
  \Name{Chunyang Ye} \Email{cyye@hainanu.edu.cn}\\
  \addr College of Computer and Cybersecurity, Hainan University, Haikou, China
  \AND
  \Name{Wei Li} \Email{liwei.aiuniverse@gmail.com
  }\\
  \addr College of Chemical Engineering, Fuzhou University, Fuzhou, China
 }

\begin{document}

\maketitle


\begin{abstract}

Photorealistic style transfer aims to transfer the style of one image to
another, but preserves the original structure and detail outline of the content
image, which makes the content image still look like a real shot after the
style transfer. Although some realistic image styling methods have been
proposed, these methods are vulnerable to lose the details of the content image and
produce some irregular distortion structures. In this paper, we use a
high-resolution network as the image generation network. Compared to other
methods, which reduce the resolution and then restore the high resolution, our
generation network maintains high resolution throughout the process. By
connecting high-resolution subnets to low-resolution subnets in parallel and
repeatedly multi-scale fusion, high-resolution subnets can continuously receive
information from low-resolution subnets. This allows our network to discard
less information contained in the image, so the generated images may have a
more elaborate structure and less distortion, which is crucial to the
visual quality. We conducted extensive experiments and compared the results
with existing methods. The experimental results show that our model is
effective and produces better results than existing methods for photorealistic
image stylization. Our source code with PyTorch framework will be publicly
available at \url{https://github.com/limingcv/Photorealistic-Style-Transfer}.

\end{abstract}


\section{Introduction}

The main purpose of photorealistic image stylization (also known as color style
transfer) is to transfer the style of color distributions
\cite{jing2017neural}. Given a reference style, an input image can be transfered to 
make it look like it is in different
lighting, time of day or weather, or it has been decorated with art with
different intents.
A successful stylization should keep the semantic content of the input image and 
the output image should look like a real photo made by the camera.

\begin{figure}[bt]
	\centering
	\subfigure[Content]{
			\includegraphics[width=0.18\linewidth]{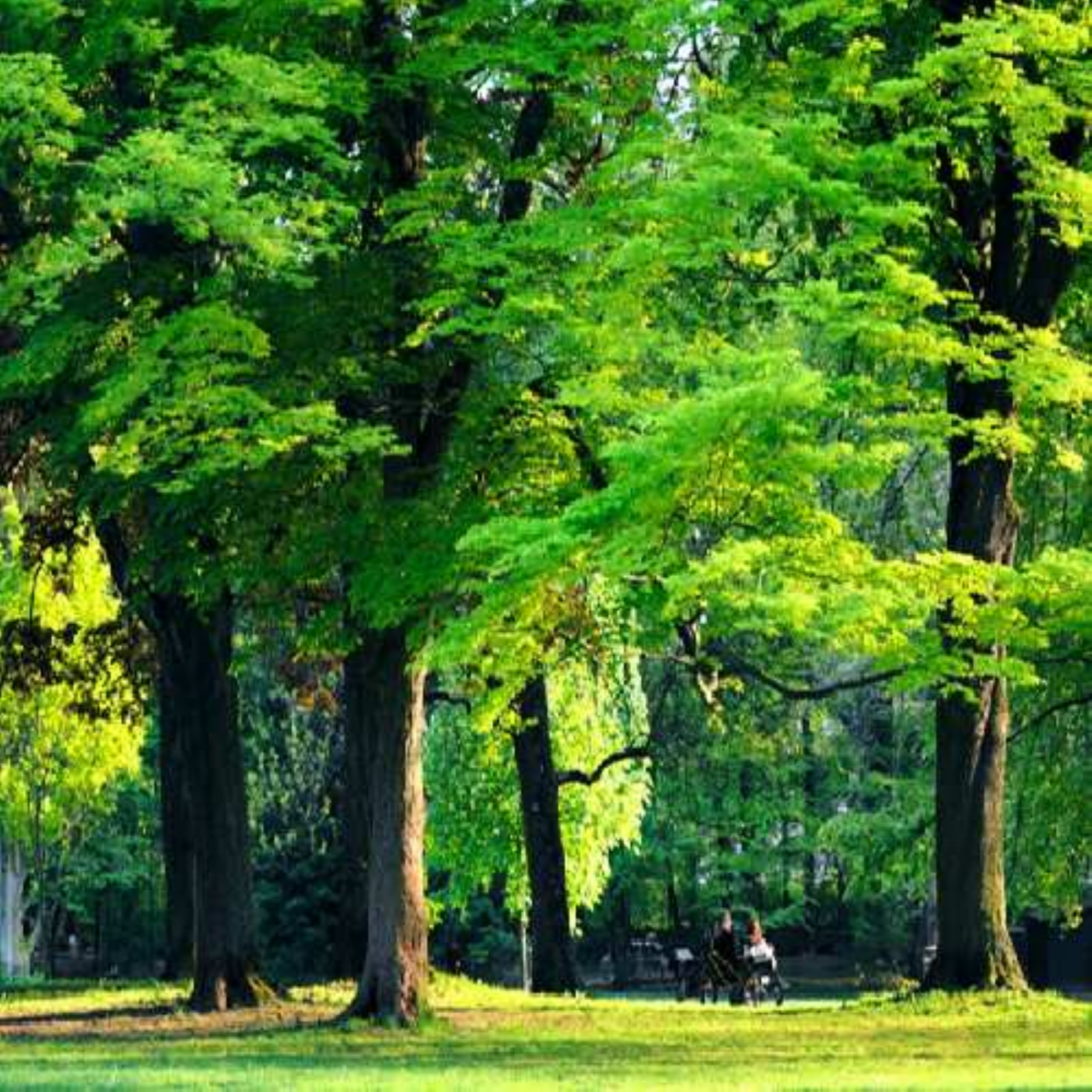}
			\label{1.a}
	}%
	\subfigure[Style]{
			\includegraphics[width=0.18\linewidth]{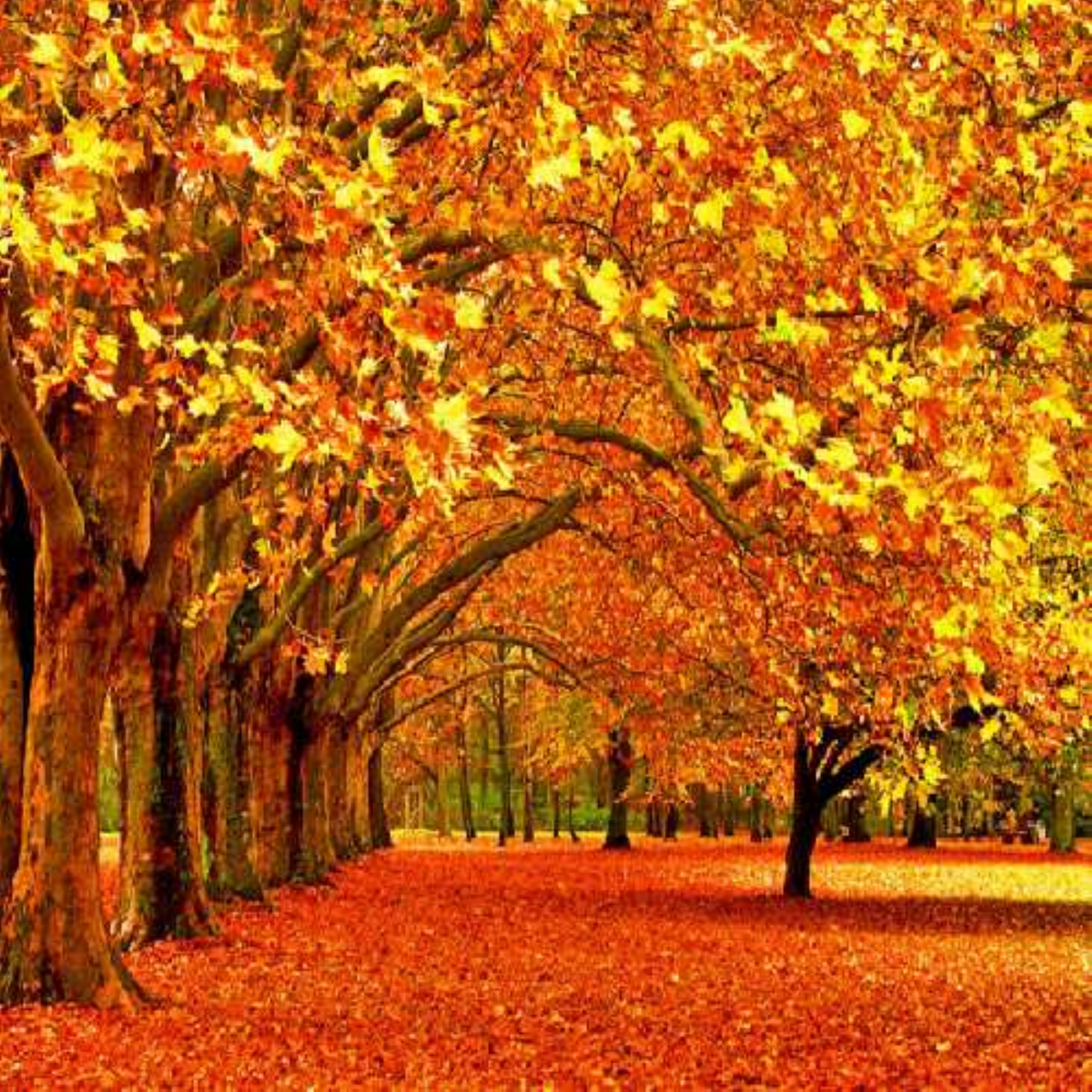}
			\label{1.b}
	}%
	\subfigure[Neural Style]{
			\includegraphics[width=0.18\linewidth]{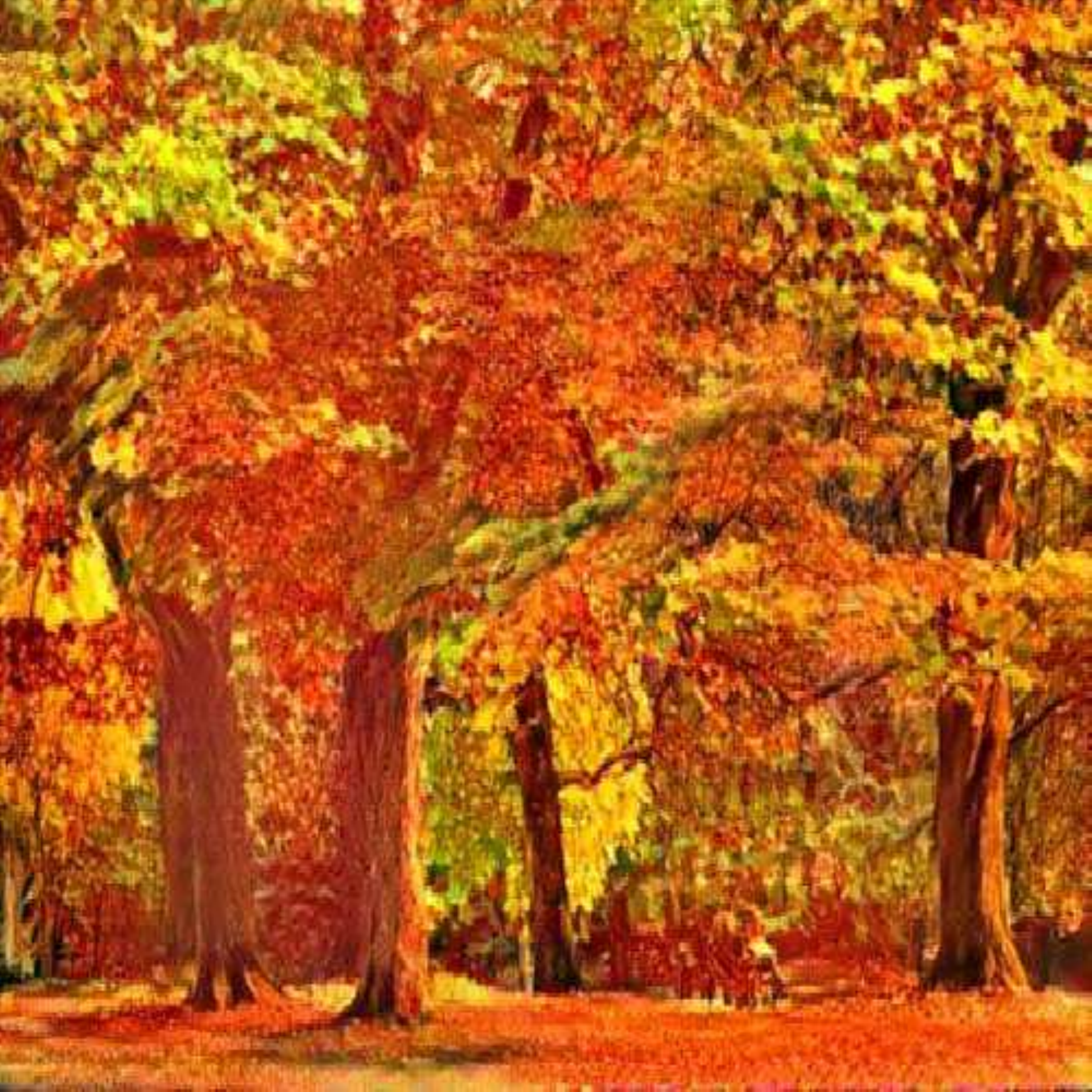}
			\label{1.c}
	}%
	\subfigure[CNNMRF]{
			\includegraphics[width=0.18\linewidth]{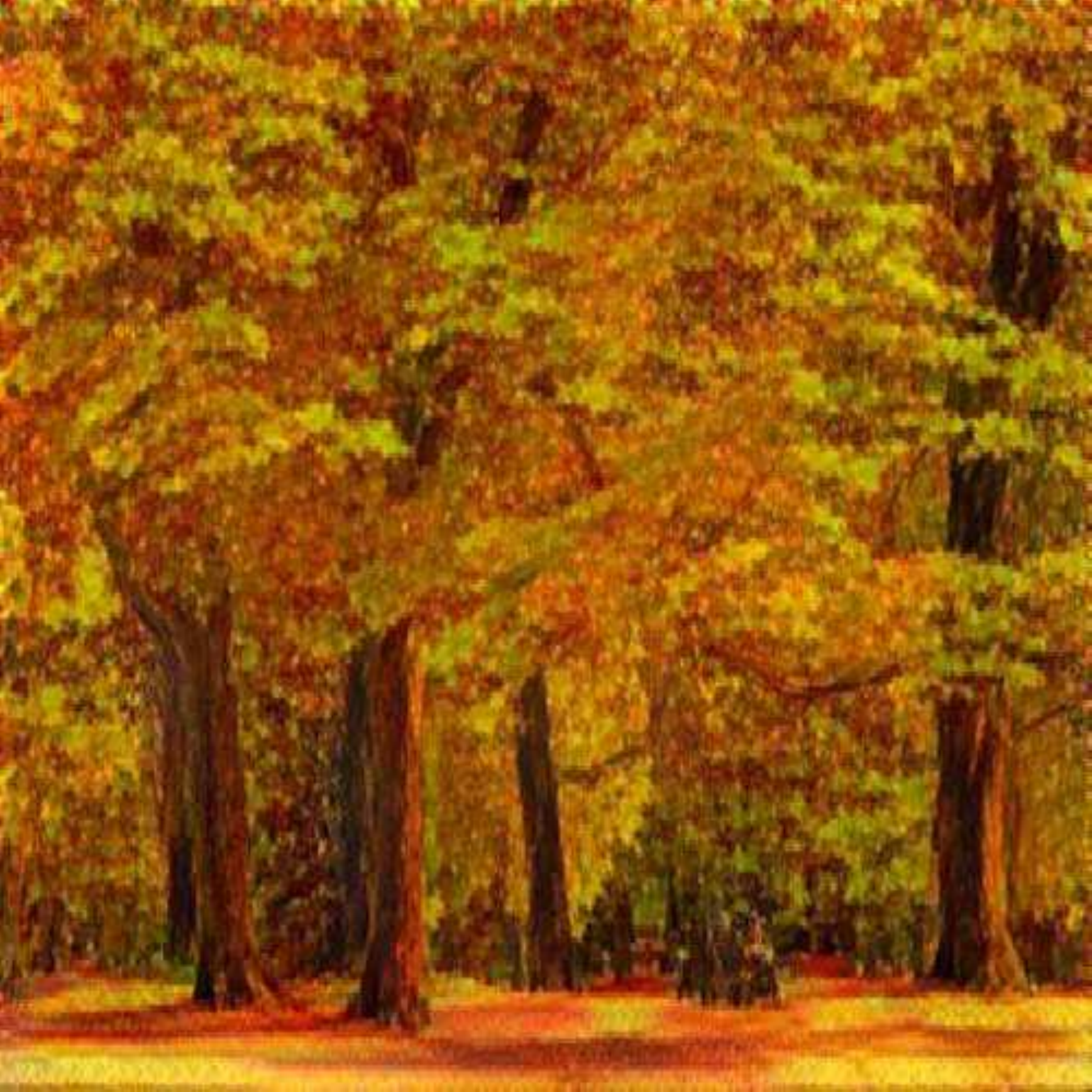}
			\label{1.d}
	}%
	\subfigure[Ours]{
			\includegraphics[width=0.18\linewidth]{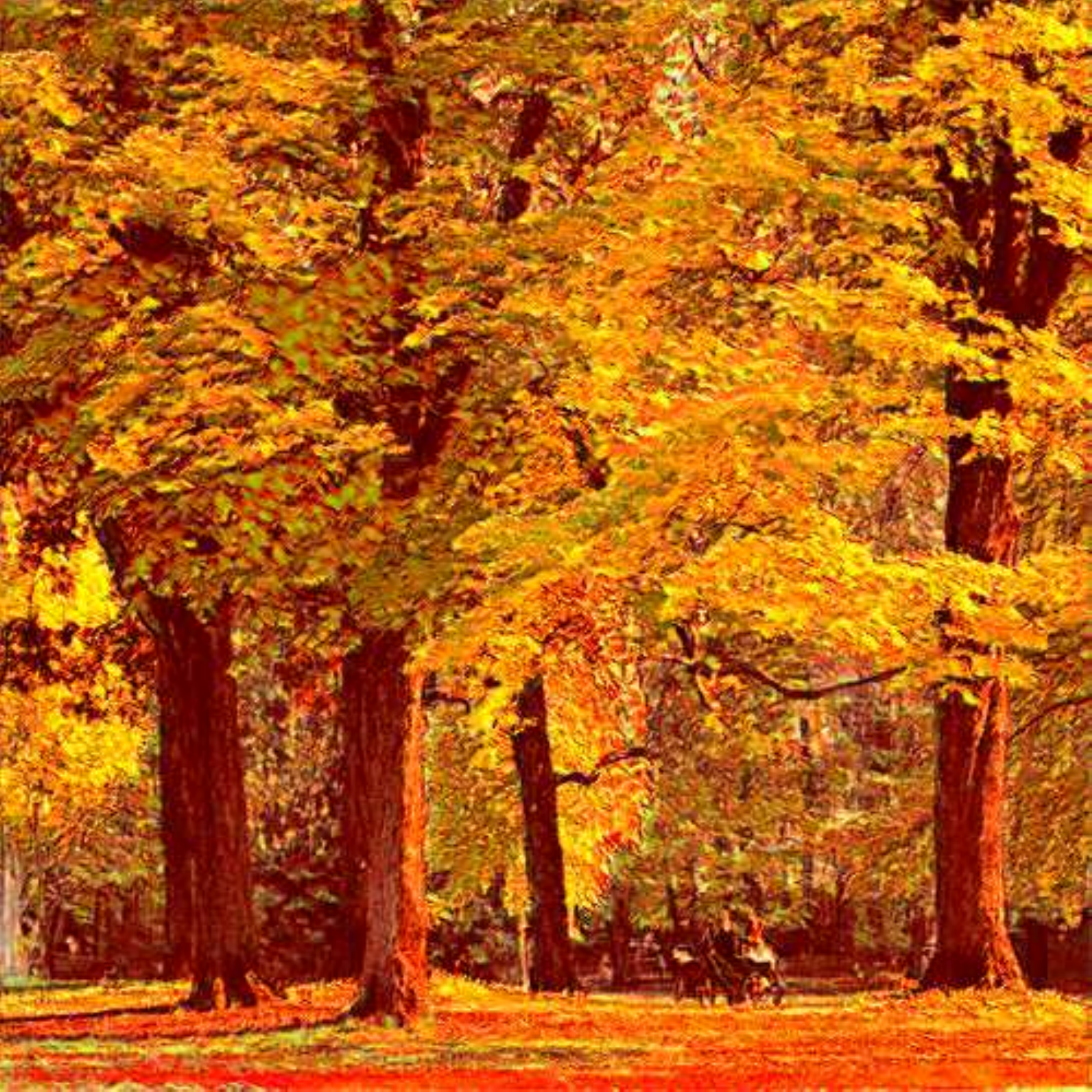}
			\label{1.e}
	}%

	\subfigure[Content]{
			\includegraphics[width=0.18\linewidth]{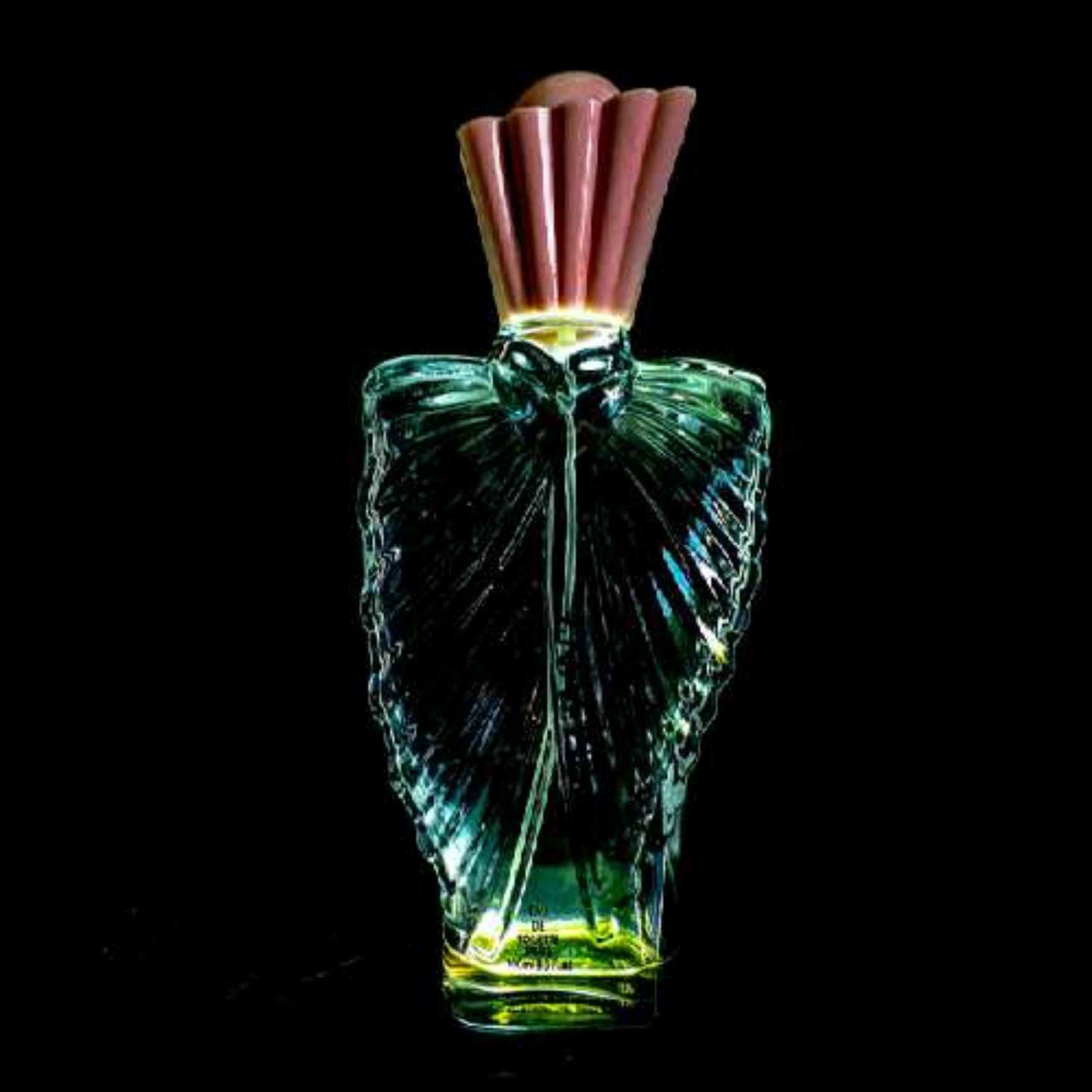}
			\label{1.f}
	}%
	\subfigure[Style]{
			\includegraphics[width=0.18\linewidth]{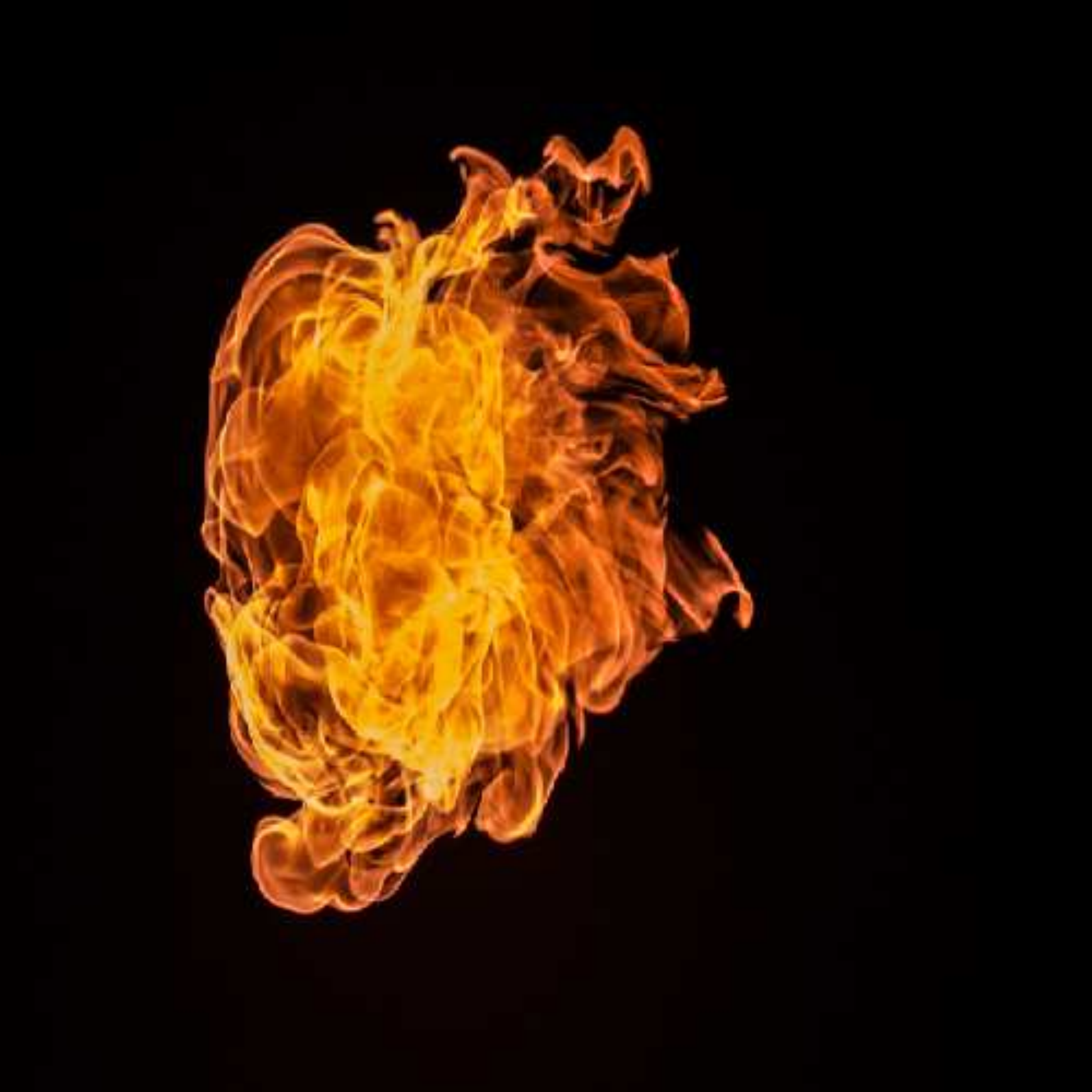}
			\label{1.g}
	}%
	\subfigure[Reinhard]{
			\includegraphics[width=0.18\linewidth]{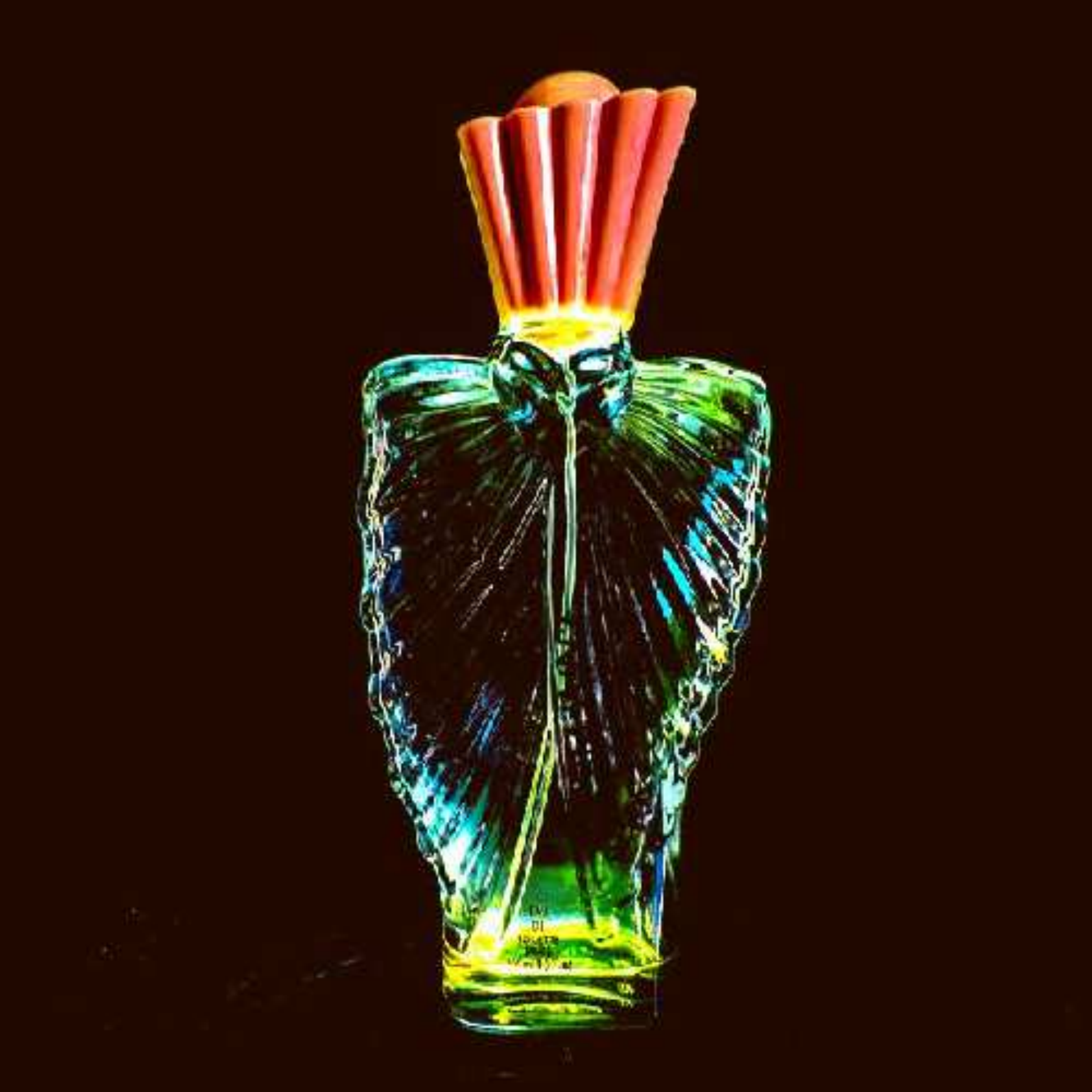}
			\label{1.h}
	}%
	\subfigure[Pitié]{
			\includegraphics[width=0.18\linewidth]{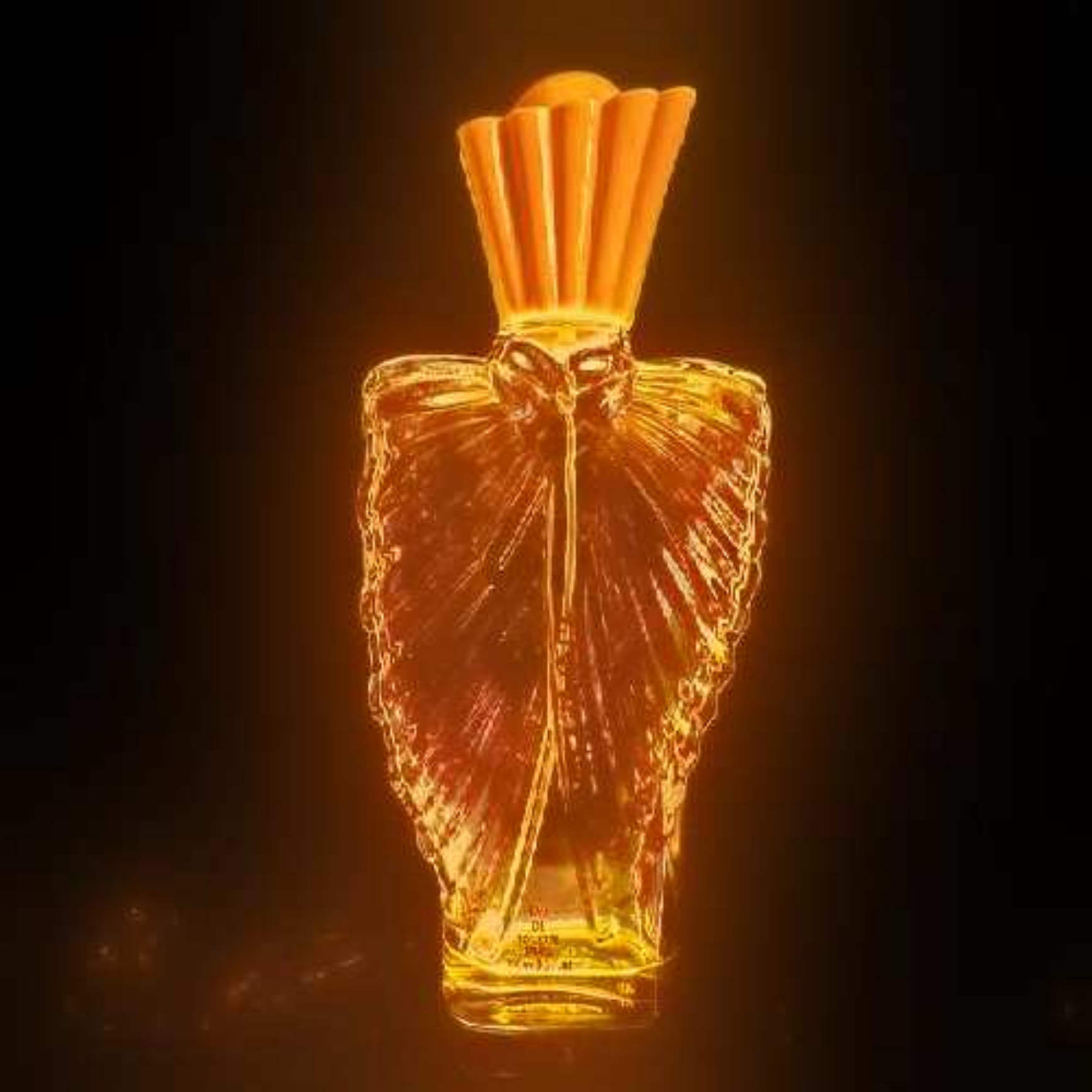}
			\label{1.i}
	}%
	\subfigure[Ours]{
			\includegraphics[width=0.18\linewidth]{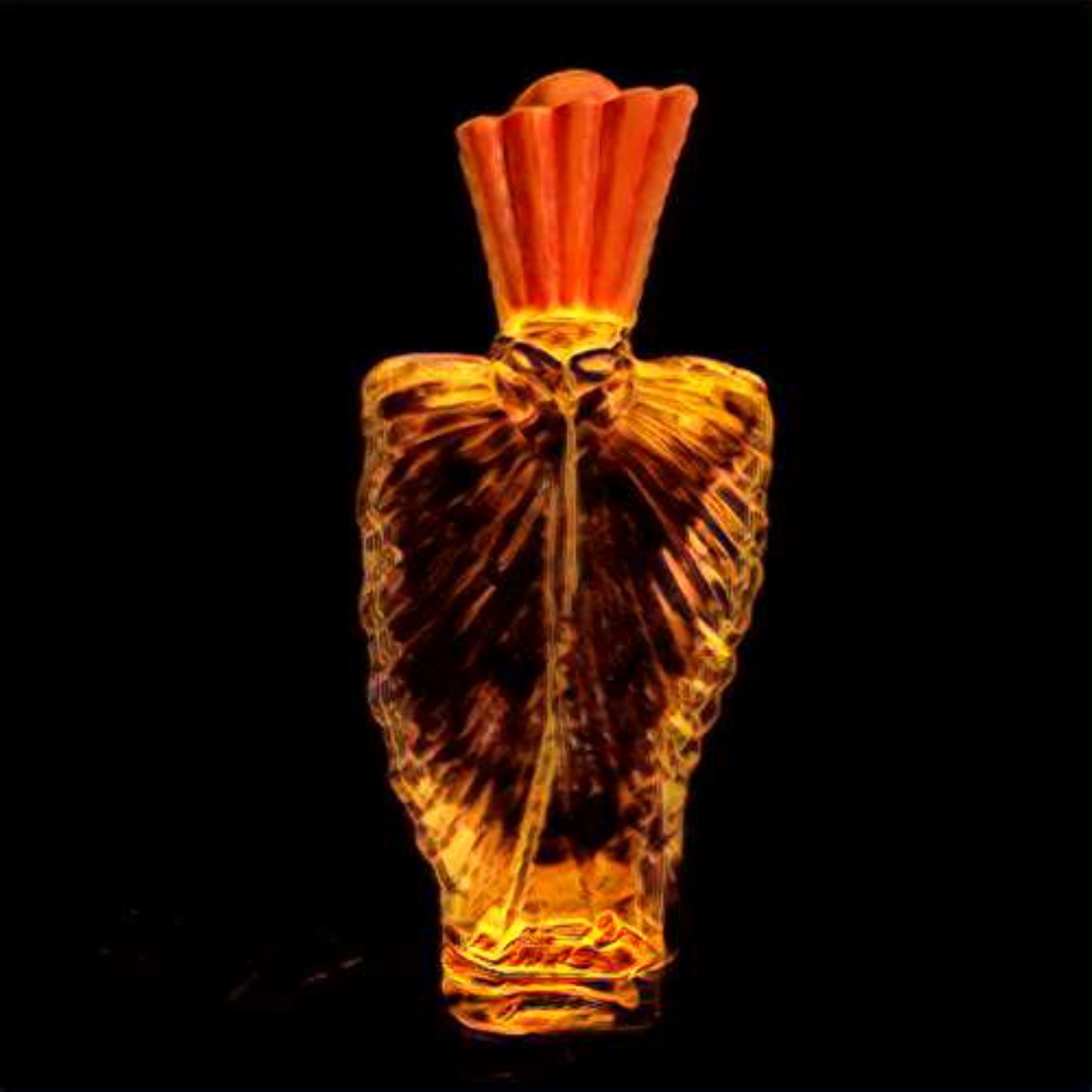}
			\label{1.j}
	}%
        \caption{Compared with our results, the results \ref{1.b}
        of \cite{gatys2016image} (Neural Style) and \ref{1.c} of
        \cite{li2016combining} (CNNMRF) produced severe distortions in visual
        effects and led to loss of semantic information of content images. Our
       approach provides greater flexibility in delivering spatially varying color
       variations and produces better results than \cite{reinhard2001color} and
       \cite{pitie2005n}.}\label{fig1} 
   	
\end{figure}

Conventional realistic stylized methods are usually based on tonal matching or
color matching
\cite{reinhard2001color,pitie2005n,sunkavalli2010multi,bae2006two}, but
unfortunately these methods can only be used in special scenes, or the images
after stylization look not real.
Recently, \cite{gatys2015texture,gatys2016image} show that the gram matrix in
feature map of convolution neural nets (CNN) can represent the style of an
image and propose the neural style transfer algorithm for image stylization.
Although the method has an amazing performance in the transfer of artistic
image styles, the application to photorealistic style transfer often results in
generated images with less semantic information and distortions in the image
than the content image as shown in Fig. \ref{1.c}. 

Since CNN may lose some low-level information contained in
the image during the downsampling process, some unattractive
distortion structures and irregular artifacts usually exist in the stylized results.
In order to maintain the consistency of the fine structure during the
stylization process, \cite{li2017laplacian} suggests adding additional
constraints on low-level features in pixel space. An additional
Laplace loss (defined as the square Euclidean distance between the
Laplacian filter response of the content image and the stylized result) is introduced.
Although Li's algorithm maintains fine structure and details in the stylization
process with good performance, it still lacks considerations such as semantics,
depth, and changes in brush strokes, as shown in Fig. \ref{1.d}.

Some style transfer algorithms also apply a spatial invariant transfer function to
process the image, but these algorithms can deal with simple styles transfer
such as global color shift and tone curves only. For example, to convert the
input and style images to a decorrelated color space, \cite{reinhard2001color}
proposed to match the mean and standard deviation between the input and style
images. \cite{pitie2005n} also proposed an algorithm to
transfer a full 3D color histogram based on a series of 1D histograms. These
methods however have limited ability to match complex styles (c.f.
Section~\ref{sec:exps})
\cite{luan2017deep} proposed a two-stage optimization procedure to
first stylise a given photo with
non-photorealistic style transfer algorithm and then penalise image distortions
by adding a photorealism regularization. However, it usually produces
inconsistent styles with obvious artifacts and is computationally expensive.

\begin{figure}[bt]
	\begin{center}
		\includegraphics[width=1\textwidth]{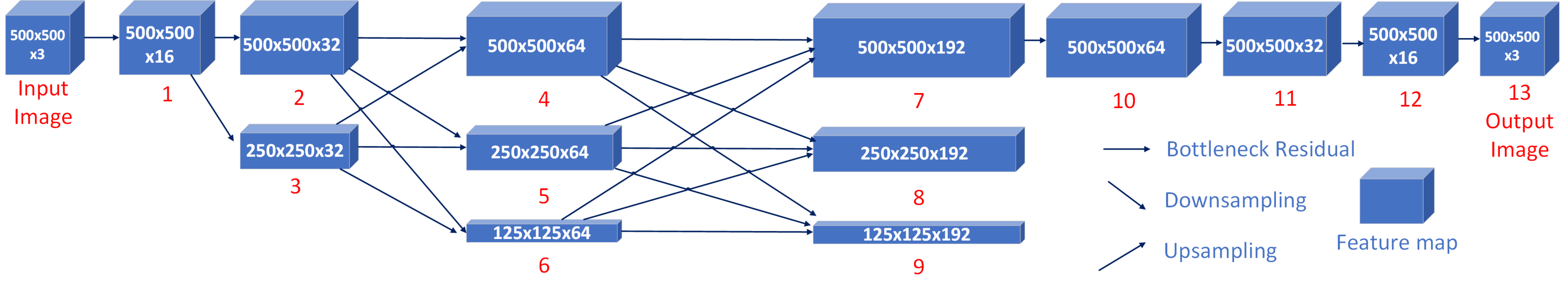}
		\subfigure[Fusion1]{
			\begin{minipage}[t]{0.2\linewidth}
				\centering
				\includegraphics[width=1in]{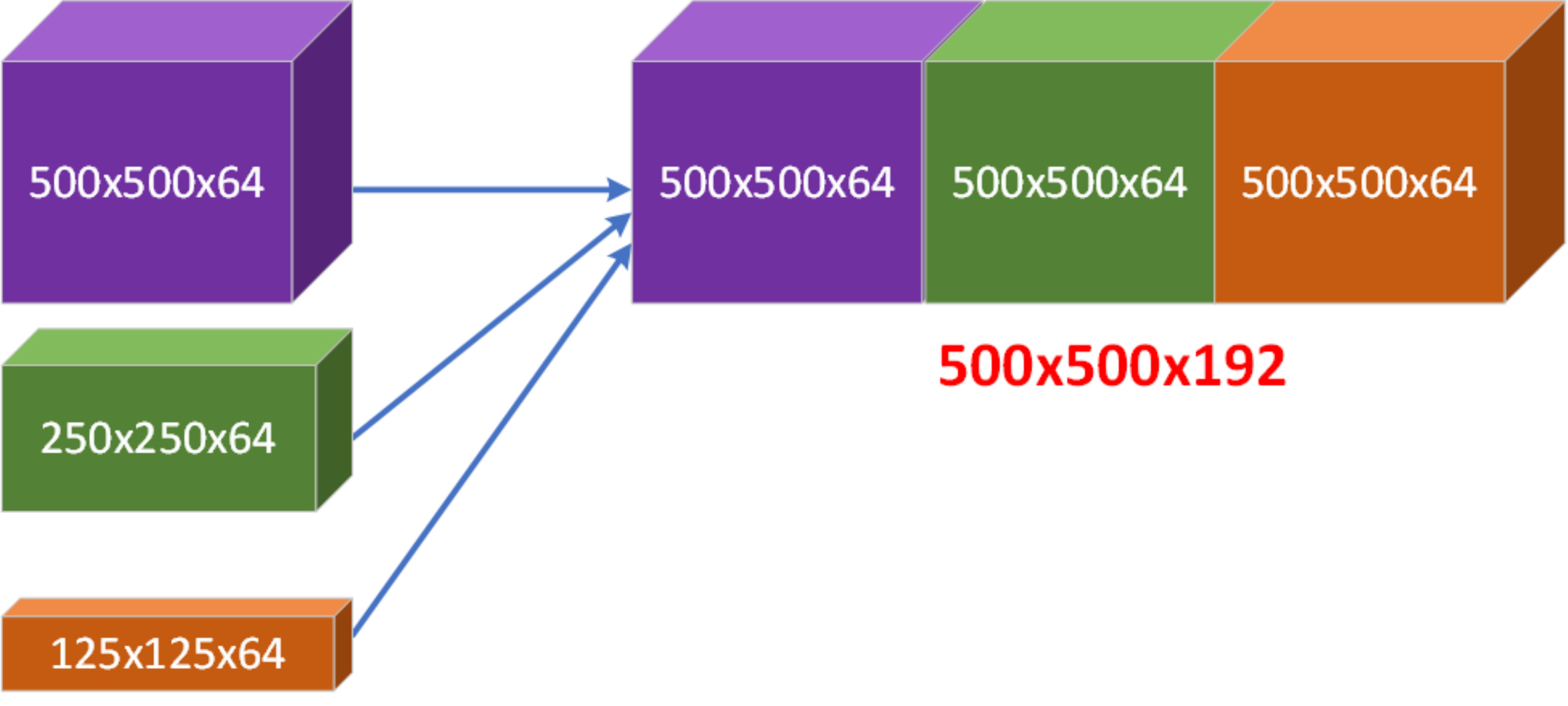}
			\end{minipage}
		}%
		\subfigure[Fusion2]{
			\begin{minipage}[t]{0.2\linewidth}
				\centering
				\includegraphics[width=1in]{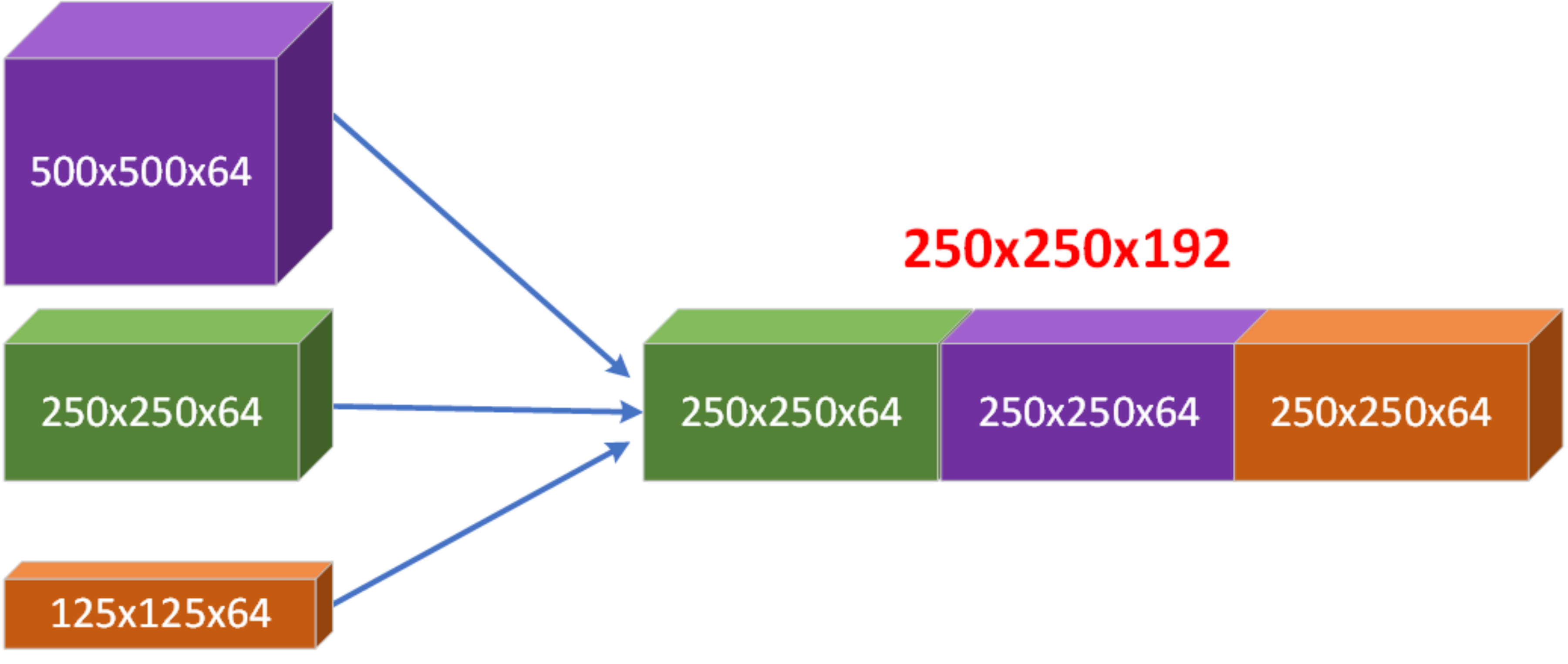}
			\end{minipage}
		}%
		\subfigure[Fusion3]{
			\begin{minipage}[t]{0.2\linewidth}
				\centering
				\includegraphics[width=1in]{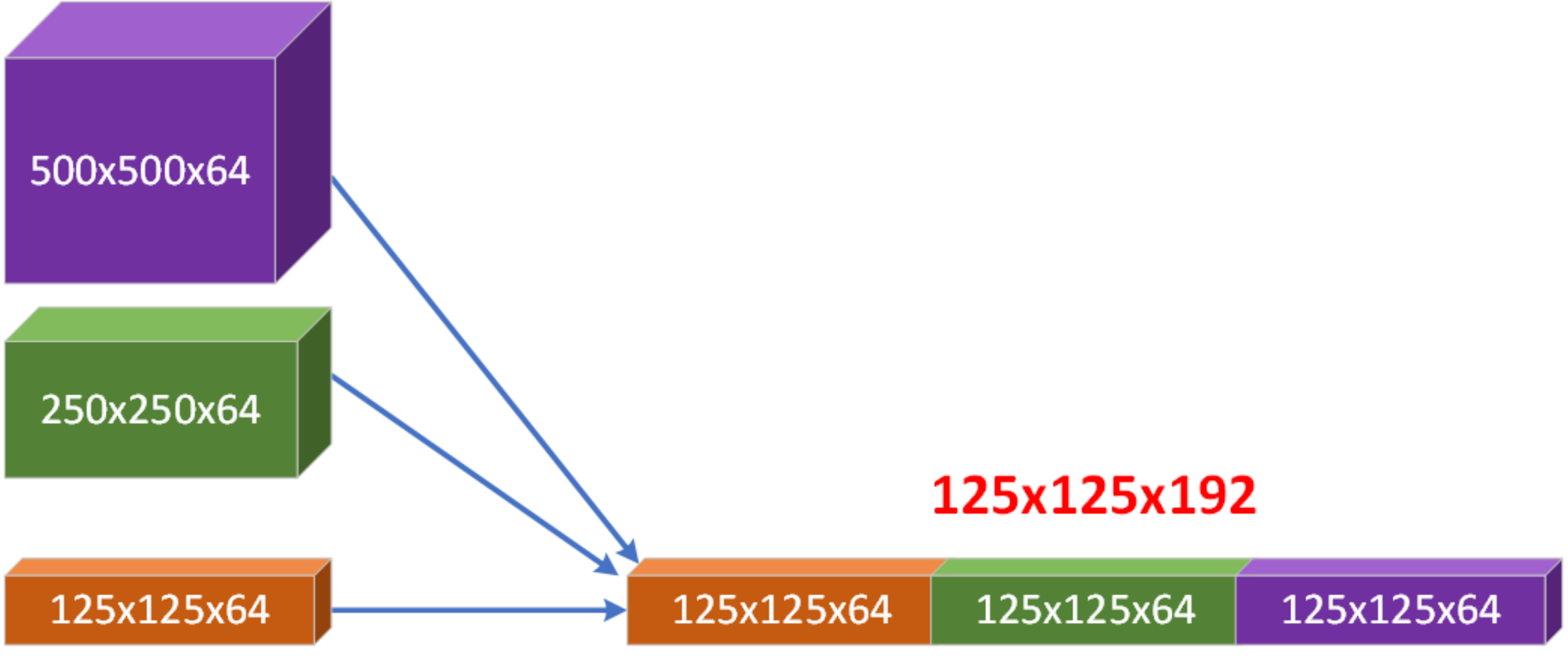}
			\end{minipage}
		}%
	
            \caption{The structure of high-resolution generation network.
            Different from \cite{sun2019deep}, when we fuse feature maps with
            different resolution, we directly concatenate these feature
            images like the inception module, for example, the feature map 4,
            is concatenated by the feature map 2 and the feature map 3. We use
            bottleneck residual to ensure that our network can be trained well
            and speedup the training while preserving good visual
            effects.}\label{HRnet}
		\vspace{-5mm}
        \end{center}
\end{figure}

To address these issues, we propose in this paper a High-Resolution Network for Photorealistic Style Transfer. 
Inspired by the network proposed by
\cite{johnson2016perceptual}, our solution has a generation network to generate the
output image, and a pre-trained network to calculate the content loss and style
loss, but the architecture of our generation network is different from the network in~\cite{johnson2016perceptual}.
To make the natural image photorealistic style transfer method successful
with a more elaborate structure and less distortion, we use the high-resolution network as the generation network. 
In addition, we use VGG19 as the
loss network to calculate the losses instead of the VGG16, because we found in our
studies that the pre-trained VGG19 is better than the pre-trained VGG16.

The main contributions of this paper are three-fold:
First, we propose a
high-resolution network as the generation network to
transfer the style with a finer structure and less distortion. Second,
we implement the 
photorealistic style transfer successfully using
traditional natural image style transfer algorithm, which provides a new choice
for photorealistic style transfer. Last, we conduct extensive experiments to
evaluate our proposal. Compared with \cite{luan2017deep}, our algorithm outperforms
existing work in terms of a more elaborate structure, less distortion and a faster rate (c.f. Section~\ref{sec:exps}). 
\footnote{All
content images and style images except Fig. \ref{fig4} and Appendix A can be
found in \cite{luan2017deep}. The experimental results of other methods are
also from \cite{luan2017deep}.}

The rest of this paper is organized as follows: Section~\ref{sec:rw} reviews the
state-of-the-art efforts on style transfer. Section~\ref{sec:method} presents
our methodology. Section~\ref{sec:exps} evaluates our proposal and makes a
comparison with existing solutions. Section~\ref{sec:conclusion} concludes the
work and highlights some potential directions.

\section{Related Work}\label{sec:rw}
\subsection{Neural Style Transfer}

\cite{gatys2016image} proposed a neural style transfer algorithm for artistic
stylization. The core idea of the algorithm is to extract features and
calculate content loss and style loss through a pre-trained network. Many followup  
research efforts such as
\cite{johnson2016perceptual,li2016combining,ulyanov2016texture,li2017diversified,chen2017stylebank,dumoulin2017learned,ghiasi2017exploring,huang2017arbitrary,li2017universal,liao2017visual}
have been devoted to further improving their stylized performance and speed.
However, these methods are unable to maintain photorealism (c.f. \ref{1.c}).
Some researchers also proposed post-processing techniques to improve these
results by matching the gradient between the input and output
photos~\cite{li2017laplacian,mechrez2017photorealistic}. 

\cite{johnson2016perceptual} and \cite{ulyanov2016texture} proposed the
fast style transfer algorithms. These two methods share a similar idea, that
is, train a generation network to generate the output image, and then put the output
image, style image and content image into a pre-trained VGG to calculate the
loss, and finally update the parameters of the generated network through
back propagation. As the number of iterations increases, the output image
becomes better. 

\begin{figure}[bt]
	\begin{center}
		\includegraphics[width=1\textwidth]{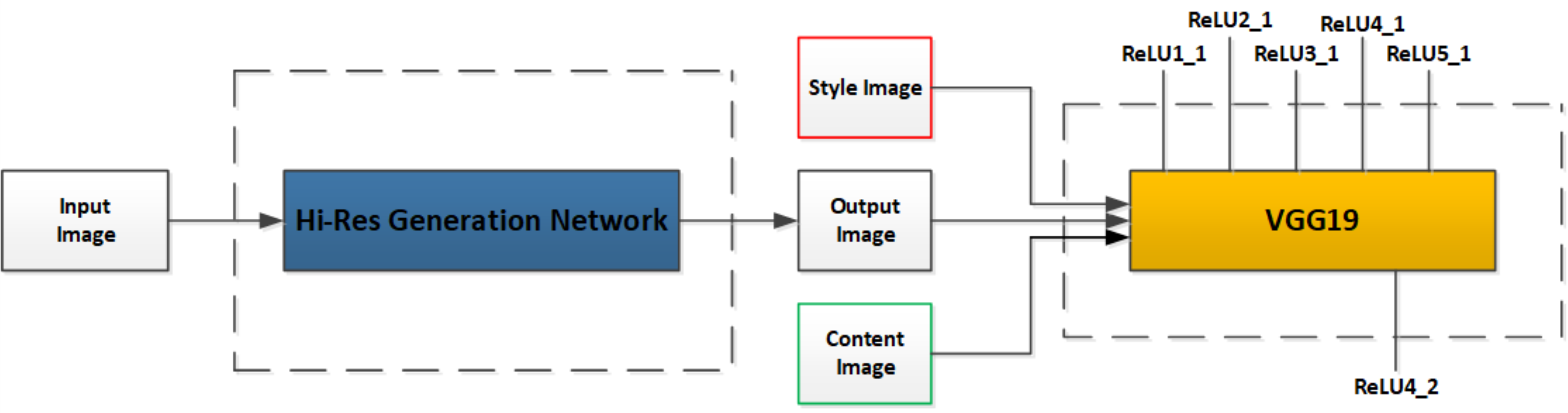}
                \caption{Overview of our network. The output image is generated
                by the high-resolution generation network, and then
                input into the pre-trained VGG to calculate the content
                loss and style loss. The parameters in the generation
                network are updated by back propagation of the total loss to make the output
                image better.} \label{overview} 
        \end{center} 
\end{figure}

\subsection{High-Resolution Network}

Our generation network is inspired by \cite{sun2019deep}, who proposed the
high-resolution network for pose estimation and refreshed the record of the
COCO pose estimation data set. Most networks have the high-to-low and
low-to-high processes. 

The high-to-low process aims to produce lower resolutions and higher channel
counts, while the low-to-high process is designed to produce high-resolution
representations and reduce the resolution of the feature maps. High-resolution
network is designed to maintain high resolution representations through the
whole process and continuously receive information from low-resolution
networks. The high-resolution network has two benefits in comparison to other
networks. 
(i) The high-resolution network connects both high and low
    resolution subnets in parallel, rather than connecting in series like most
    existing networks. (ii) Perform repeated multi-scale fusion with the help
    of low resolution representations of the same depth and similar levels to
    enhance high resolution representation. 

\subsection{Recent Work in Photorealistic Style Transfer}

Closest to our work is the methods of~\cite{luan2017deep} and
\cite{li2018closed}, but our approach is different from their methods.
\cite{luan2017deep}
proposed a two-stage optimization program to first 
style a given photo using a non-photorealistic style transfer algorithm
\cite{champandard2016semantic}, and then penalize image distortion by adding
regularization to the photo to achieve good results.
\cite{li2018closed} also proposed a two-step solutions (i.e., the stylisation
step and smoothing step) to address the efficiency issue. 
Since the proposed method uses the pixel affinity of the content, it may cause
the style transfer to exist only in part of the area. Our approach is to update
the network parameters by back propagation to make the generated image look
better. For high resolutions images, our method is more efficient, because we
only need to train a small number of times to achieve good results (usually 200
training steps for the 500$\times$500 content images).

\begin{figure}[bt]
	\centering
		\subfigure[Content]{
				\includegraphics[width=0.23\linewidth]{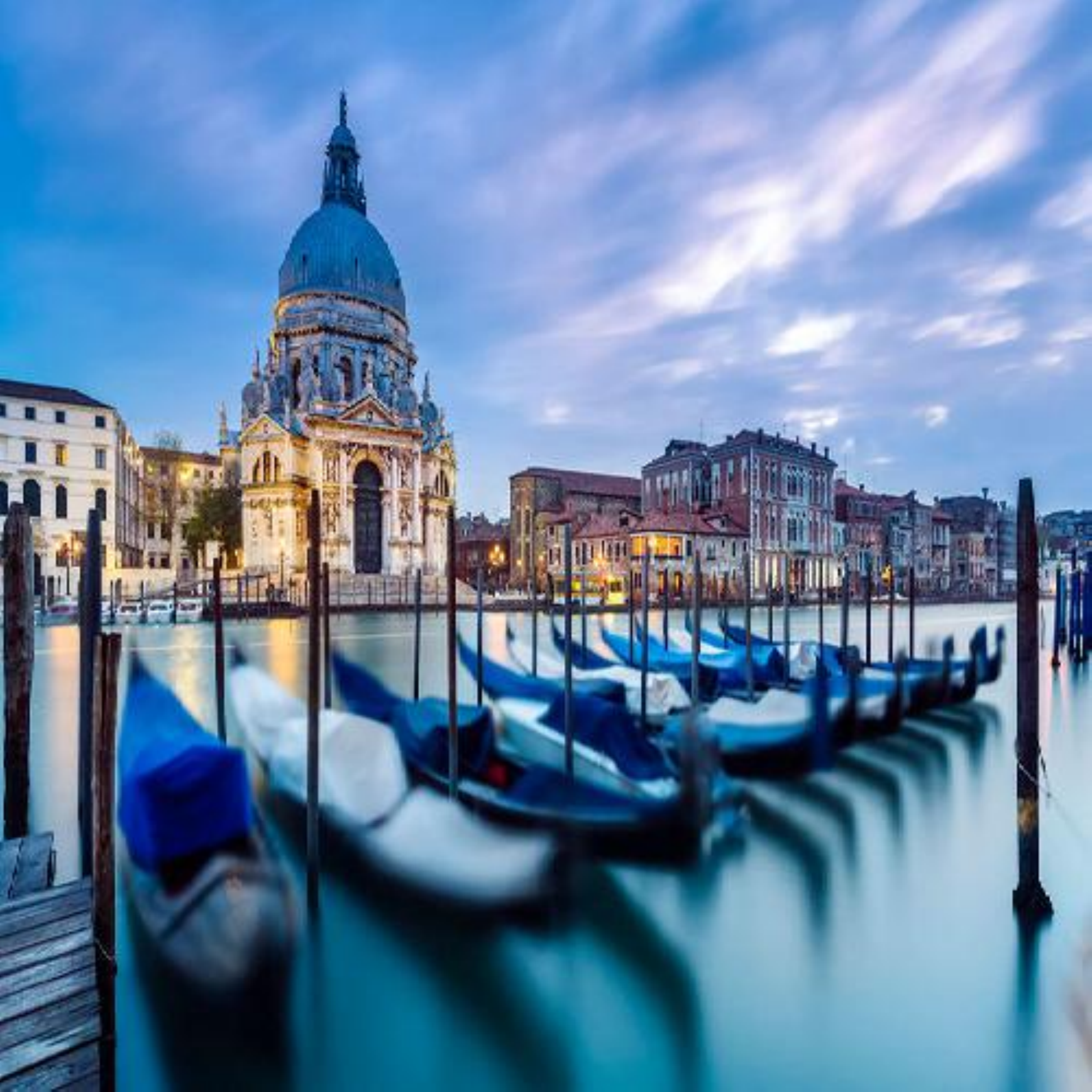}
		}%
		\subfigure[Style]{
				\includegraphics[width=0.23\linewidth]{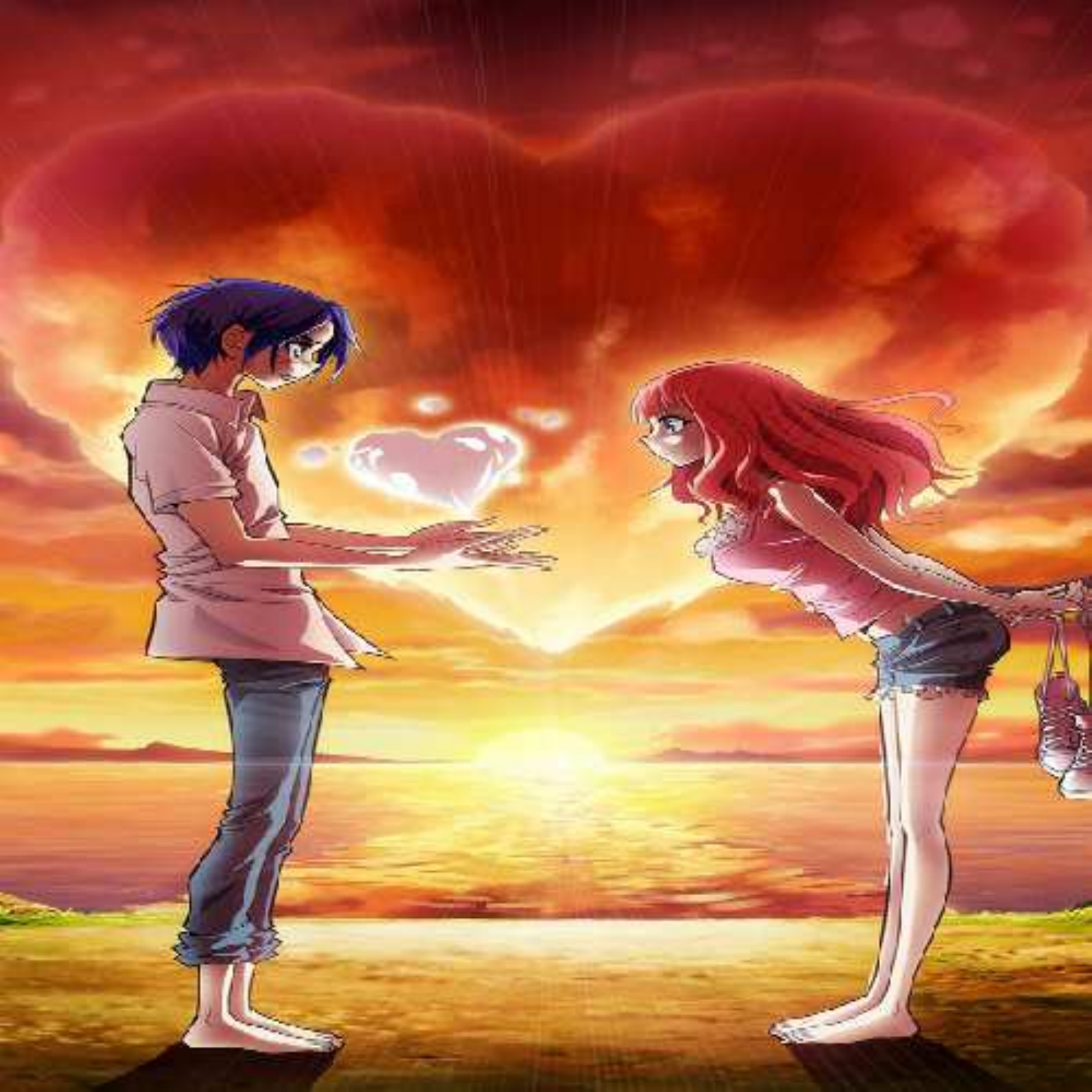}
		}%
		\subfigure[Output]{
				\includegraphics[width=0.23\linewidth]{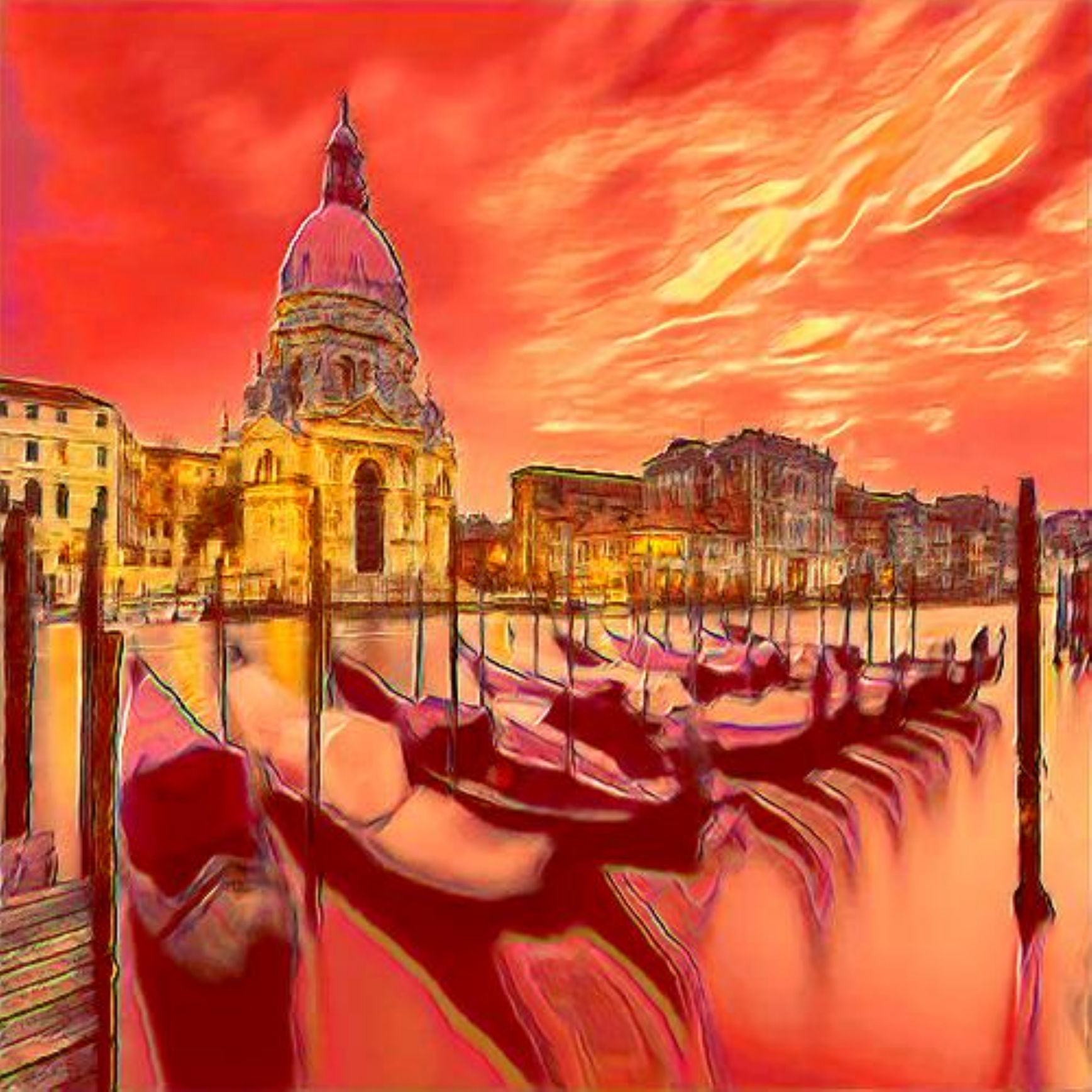}
		}%
		\subfigure[$C_{w}=0.8$]{
				\includegraphics[width=0.23\linewidth]{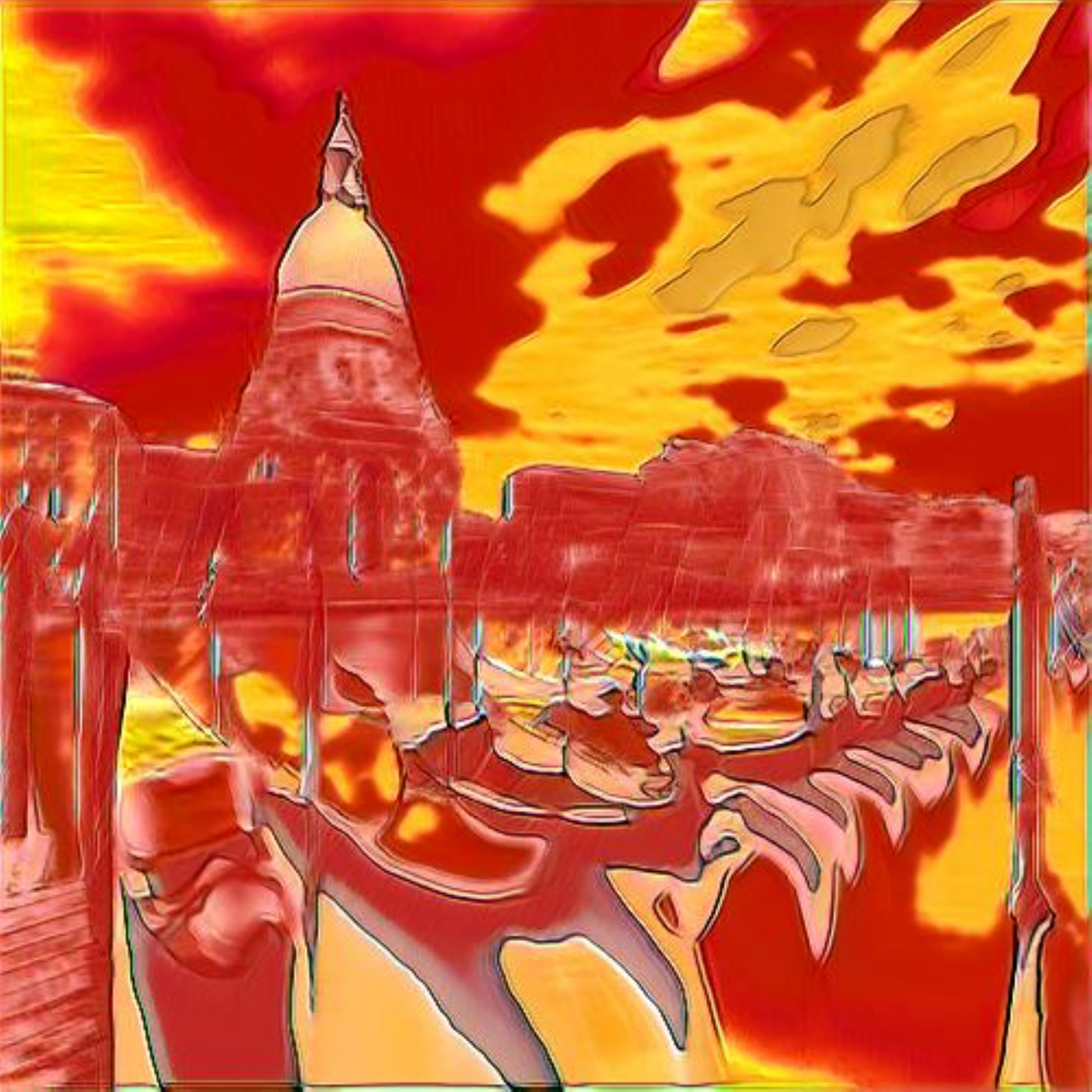}
		}

		\subfigure[$C_{w}=8$]{
				\includegraphics[width=0.23\linewidth]{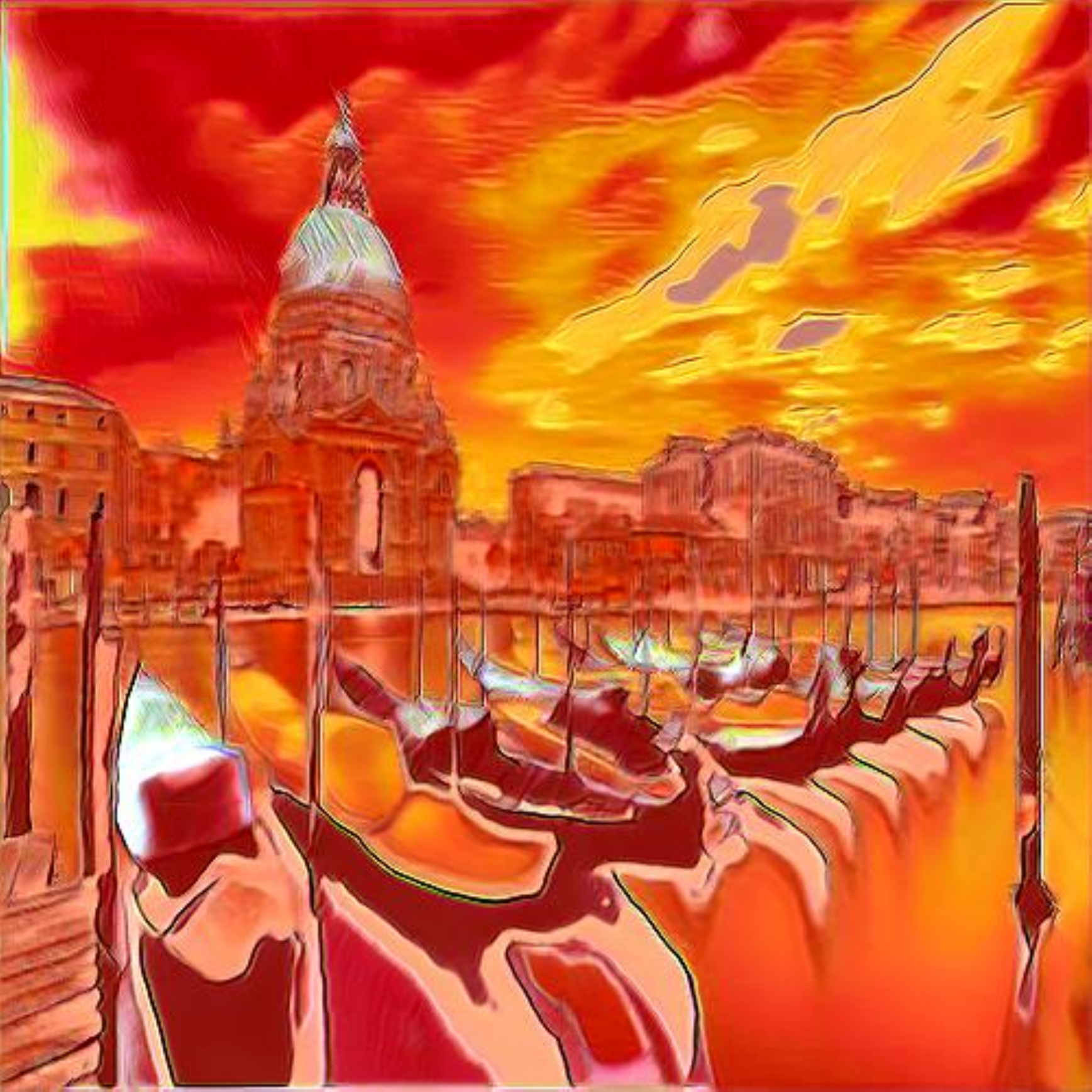}
		}%
		\subfigure[$C_{w}=80$]{
				\includegraphics[width=0.23\linewidth]{supplementary_material/output_images/different_content_weight_comparison/boat_comic_21_Wc_80-eps-converted-to}
		}%
		\subfigure[$C_{w}=800$]{
				\includegraphics[width=0.23\linewidth]{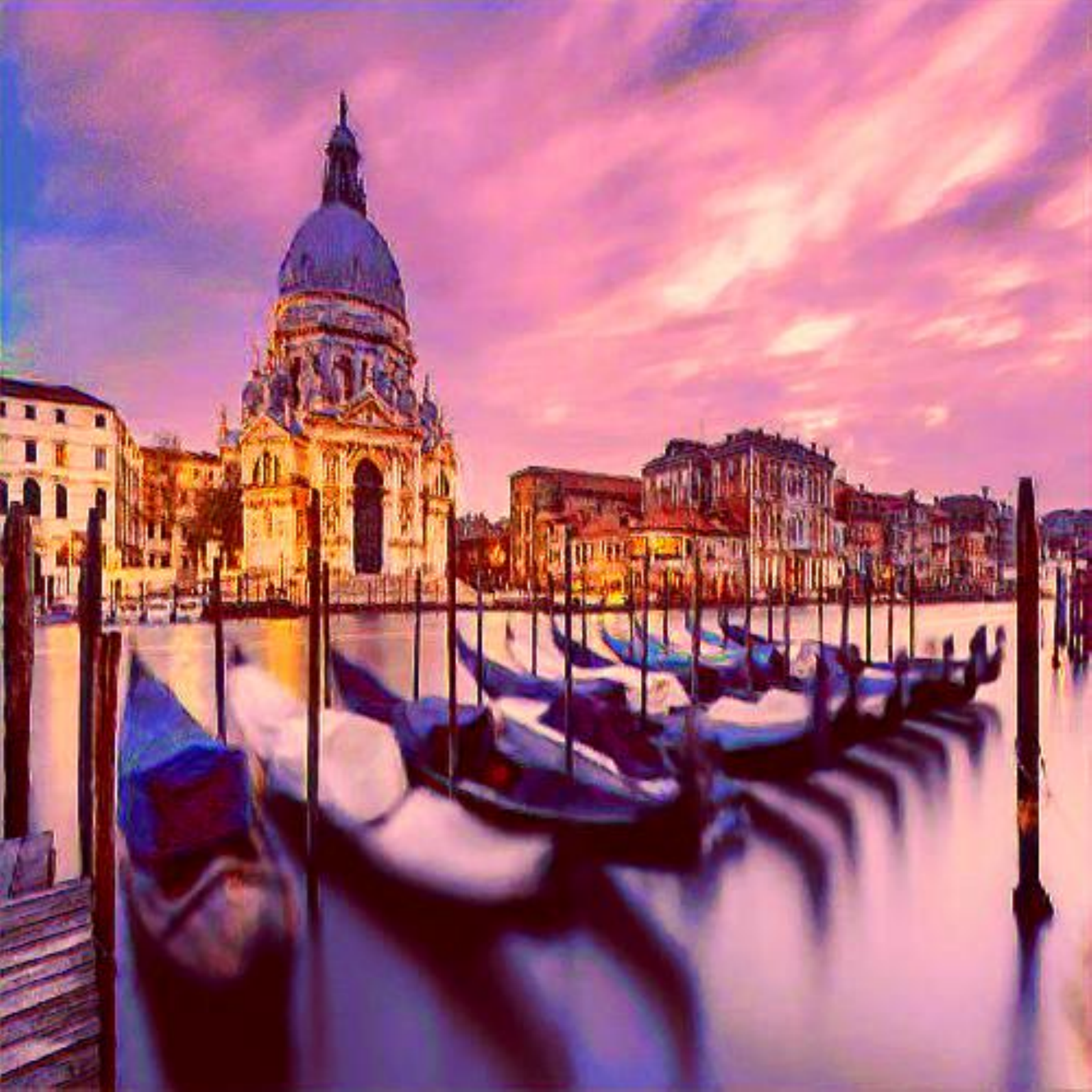}
		}%
		\subfigure[$C_{w}=8000$]{
				\includegraphics[width=0.23\linewidth]{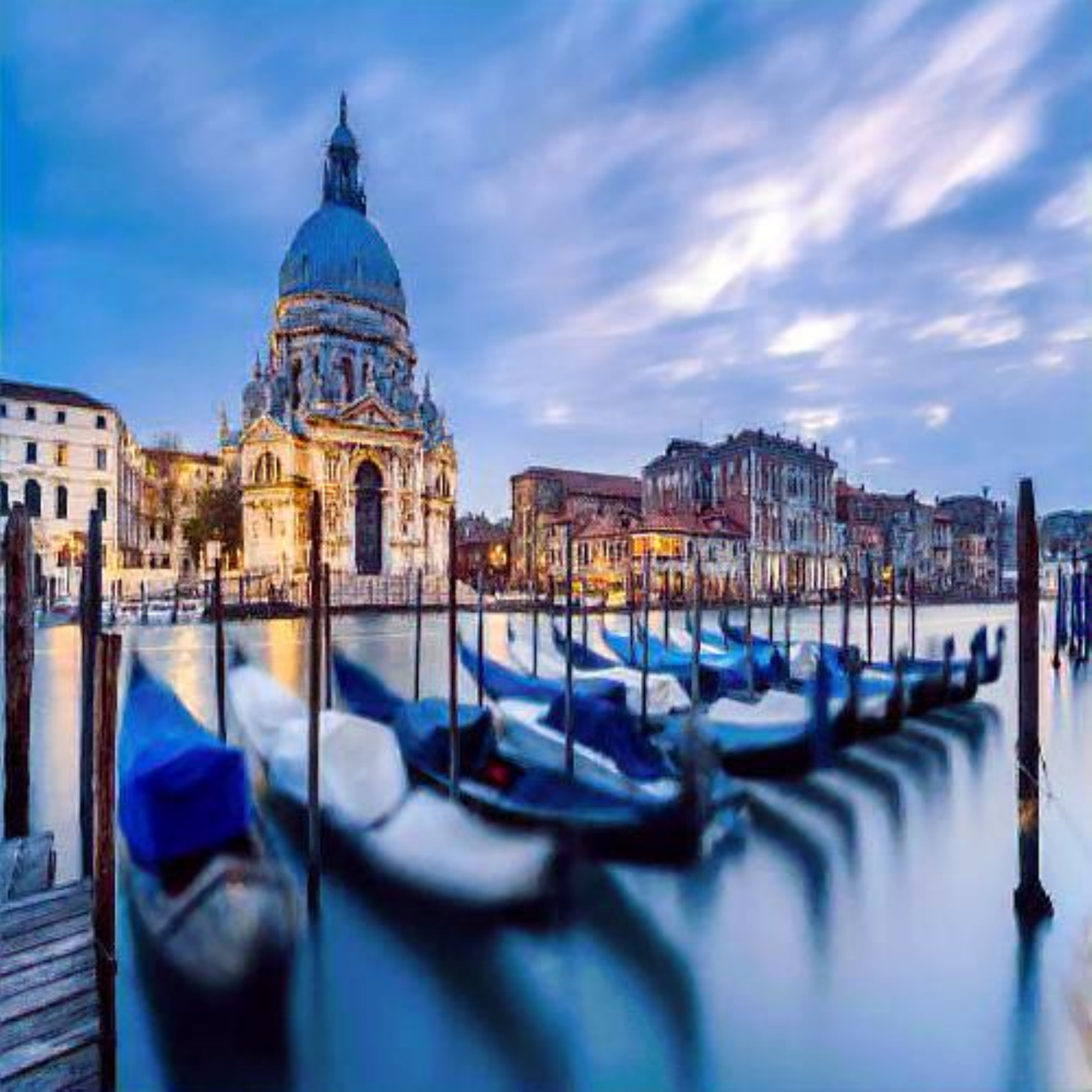}
		}%
	
            \caption{The influence of different content weights and style
            weights on the output image, $C_{w}$ is the content weight and
            $C_{s}$ is always 1. When the content weight is small, the output
            image tends to contain only a small amount of semantic information
            and precise structure of the content image. Conversely, if the
            style weight is much smaller than the content weight, the output
            image contains almost no color distribution in the style image.}
            \label{fig4} 
\end{figure}

\section{Photorealistic Style Transfer}\label{sec:method}
\subsection{Hi-Res Generation Network}

\subparagraph{High-Resolution Generation Network} Our model basically follows
the model used by \cite{johnson2016perceptual} for neural style transfer. There
is a generation network and a loss network in the model. The generation network
generates the output image, which is placed in the loss network (VGG19), and
then updates the parameters of the generated network by calculating the content
loss and the style loss.  Our image generation network roughly follows the
architectural guidelines proposed by~\cite{radford2015unsupervised}. We use
strided convolutions for in-network downsampling without any pooling
layers.
The overall structure of our model is shown in  Fig. \ref{overview}, and the
high-resolution generation network is shown in  Fig. \ref{HRnet}. For
high-resolution generation network, in the same resolution propagation process,
we used the bottleneck residual. All the convolutional layers use 3$\times$3
kernels, and the mode selected for upsampling is bilinear. Furthermore, we use all
zero padding to ensure the same resolution (we have tried other padding methods
like reflection padding, but the effect is not good.) When it comes to the
fusion between different feature maps, we concatenate these feature maps
received like the inception module. This allows high-resolution subnets to have
both high-resolution feature map information and low-resolution feature map
information.

\subparagraph{Inputs and Outputs} For photorealistic style transfer the input
image and output image are both color images of shape 3$\times$500$\times$500 by
resizing, but all style images retain their original resolution. The resolution
of the content image must be divisible by four, as shown in Fig. \ref{HRnet},
we will obtain some feature maps with a resolution of one quarter of the input
image by downsampling. The style image and the content image resolution can be
the same, but this should be based on the principle that not changes the
resolution of the style image. Once we change the resolution of the style image,
the style distribution in the style image may not be transferred to the output
image, which causes only a portion of the output image to contain the style in
the style image.

\subparagraph{Upsampling and Downsampling} All downsampling layers are
convolution layers with a convolution kernel of 3$\times$3. We use all zero
padding instead of other padding methods like reflection padding, which make our
output image have richer semantic information. Similar to
\cite{radford2015unsupervised}, we do not use any pooling layers like max-pool
or mean-pool. The pattern used by our upsampling layer is bilinear
interpolation, because in our studies, the results of using bilinear
interpolation are slightly better than using nearest neighbor interpolation or
cubic interpolation.

\subsection{Perceptual Loss Functions}

We think the content image and the output image should have the similar feature
representations computed by the loss network VGG, which means we want the
content image and the output image have the same feature map extracted by
VGG, rather than encouraging the pixels of the content image exactly match the
pixels of the output image. The total loss consists of the content loss and
the style loss. The content loss is the (squared, normalized) Euclidean
distance between feature maps:

\begin{equation}
\ell^{\phi,j}_{content}(y,\hat{y})=\frac{1}{C_{j}H_{j}W_{j}}\Arrowvert\phi_{j}(\hat{y})-\phi_{j}(y)\Arrowvert^{2}
\end{equation}

$\hat{y}$ is the output image and y is the content image. When the input image
is x, $\phi_{j}(x)$ is the activations of the $j^{th}$ layer of loss network
$\phi$. If $j$ is a convolutional layer, then $\phi_{j}(x)$ is a feature map
with the shape of $C_{j}$$\times$$H_{j}$$\times$$W_{j}$.
So the Gram matrix $G^{\phi}_{j}($x$)$ is the $C_{j}$$\times$$C_{j}$ matrix whose elements are given by:

\begin{equation}
G^{\phi}_{j}(x)_{c,\hat{c}}=\frac{1}{C_{j}H_{j}W_{j}}\sum_{h=1}^{H_{j}}\sum_{w=1}^{W_{j}}\phi_{j}(x)_{h,w,c}\phi_{j}(x)_{h,w,\hat{c}}
\end{equation}

If each point on the grid is interpreted as giving $C_{j}$ dimensional features
$\phi_{j}(x)$, then $G^{\phi}_{j}(x)$ is proportional to the non-center covariance
of the dimensional features, and each grid position can be regarded as an independent
sample. The information about which features tend to be
activated together is thus captured. We can compute the Gram matrix efficiently by reshaping
$\phi_{j}(x)$ into a matrix $\psi$ with the shape of
$C_{j}$$\times$$H_{j}W_{j}$, which means
$G^{\phi}_{j}(x)=\psi\mathbf{\psi}^\mathrm{T}/C_{j}H_{j}W_{j}$. Then, the style
loss is defined as the squared Frobenius norm of the difference between the Gram matrices
of the output and content images:

\begin{equation}
\ell^{\phi, j}_{style}(y,\hat{y})=\Arrowvert G^{\phi}_{j}(y)-G^{\phi}_{j}(\hat{y})\Arrowvert^{2}
\end{equation}

Given style image $y_{s}$ and content image $y_{c}$, the layers j and J at
which to perform feature and style reconstruction, $\lambda_{c}$ is the content
weight, $\lambda_{s}$ is the style weight and $\lambda_{TV}$ is the total
variation regularizer, an image $\hat{y}$ is generated by solving the problem:

\begin{equation}
\hat{y}=\mathop{\arg\min_{y}}\lambda_{c}\ell^{\phi,j}_{content}(y,y_{c})+\lambda_{s}\ell^{\Phi,J}_{style}(y,y_{s})+\lambda_{TV}\ell_{TV}(y)
\end{equation}

\subsection{Implementation Details}

This section describes the implementation details of our approach. We resize
all the images to 500$\times$500 and then use the pre-trained VGG19 provided by
PyTorch as the feature extractor to calculate content loss and style loss. We
choose conv4\textunderscore2 as the content representation,
conv1\textunderscore1(weight:0.1), conv2\textunderscore1(weight:0.2),
conv3\textunderscore1(weight:0.4), conv4\textunderscore1(weight:0.8) and
conv5\textunderscore1(weight:1.6) as the style representation. We do not use
dropout, but for some situations (like the resolution of content image varies
greatly with the resolution of style image), weight decay is a good way to get
good results. The total style loss is equal to the sum of each layer's style
loss multiplied by its weight. The combination of different content pictures
and style pictures has different style weight and content weight, but in
general, we achieve good results with content weights equal to [50, 100], and
style weights equal to [1, 10]. For all residual bottleneck layers, we first
use 3$\times$3 convolution to make the number of channels of the feature map a
quarter of the input feature map, and finally use 1$\times$1 convolution to
restore the number of channels. We use Adam \cite{kingma2014adam} with a
learning rate of 1$\times 10^{-3}$. The effect of content weight and style
weight is illustrated in Fig. \ref{fig4}.

\section{Experiments}\label{sec:exps}

\subsection{Experimental Setup} We compare the proposed algorithm with two
types of stylized algorithms: realism and artistry. Photo-realistic style
algorithms evaluated include \cite{reinhard2001color,pitie2005n,luan2017deep}.
\cite{reinhard2001color} and \cite{pitie2005n} represents a classic technique
based on color statistics matching, and \cite{luan2017deep} is based on the
neural style transfer algorithm \cite{johnson2016perceptual}. On the other hand,
the set of artistic style algorithms evaluated includes \cite{gatys2016image}
and \cite{li2016combining}. They all use deep networks to realize the fusion of
style images and content images. In the experiments, we extracted the contour
information of the image and compare the effects (e.g., finer structure and
less distortion) of our solution with above mentioned solutions. We also
compared the time of different solutions using in the training. 

We also conducted two empirical studies to compare the visual effects. We
invited students from different grades and different majors to choose the most
successful images of style transfer, the pictures with richer semantic
information and the best visual images. In order to ensure the accuracy and
credibility of the results, the ratio of male to female participants is about
1:1. Their majors include art, science, engineering, management, etc. In
addition, to make the subjects in the study representative and diverse, nearly
one-third subjects are students from universities in other cities.

\begin{figure}[htp]
	\begin{center}
		\subfigure[Content]{
			\begin{minipage}[t]{0.32\linewidth}
				\centering
				\includegraphics[width=1.6in]{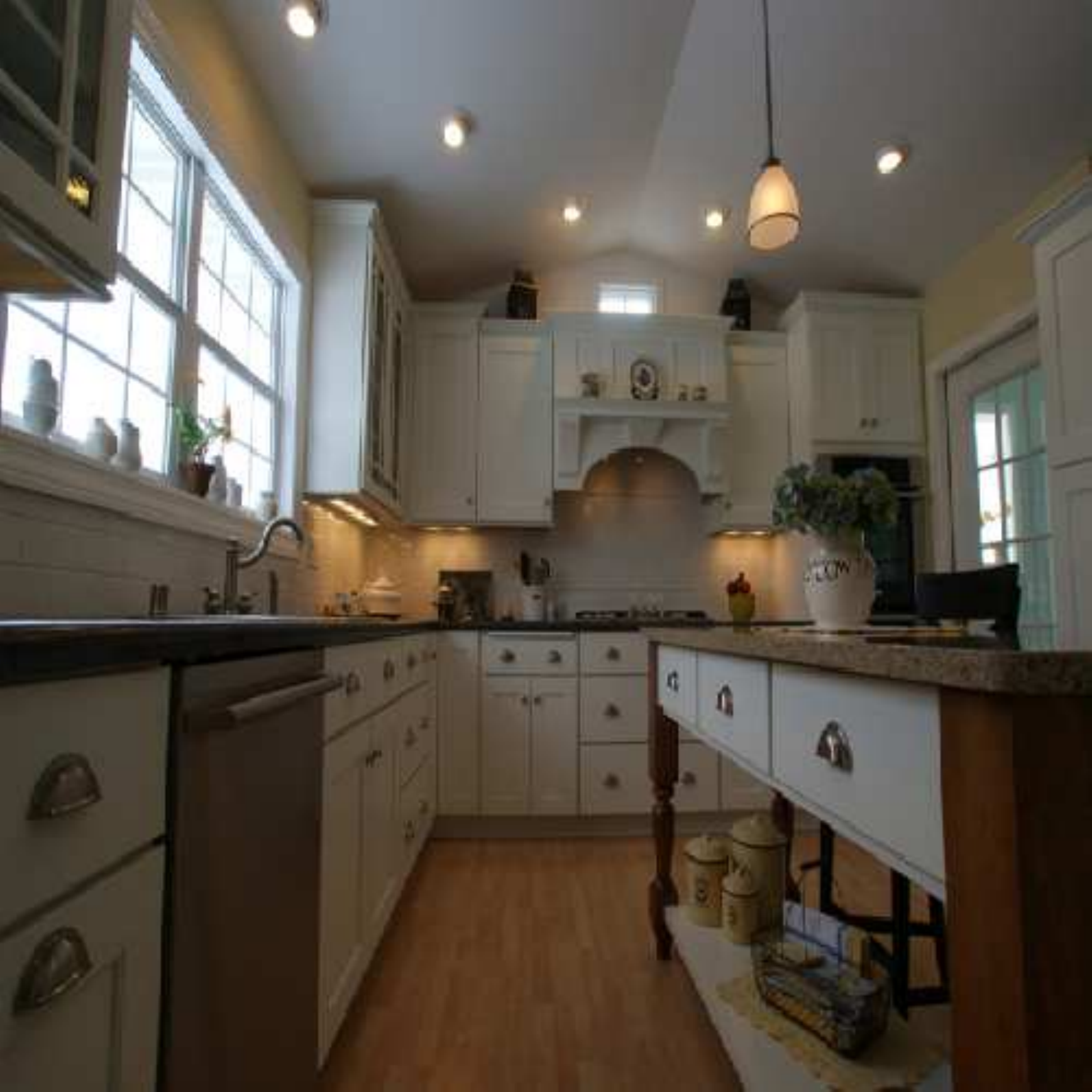}
			\end{minipage}
		}%
		\subfigure[Style]{
			\begin{minipage}[t]{0.32\linewidth}
				\centering
				\includegraphics[width=1.6in]{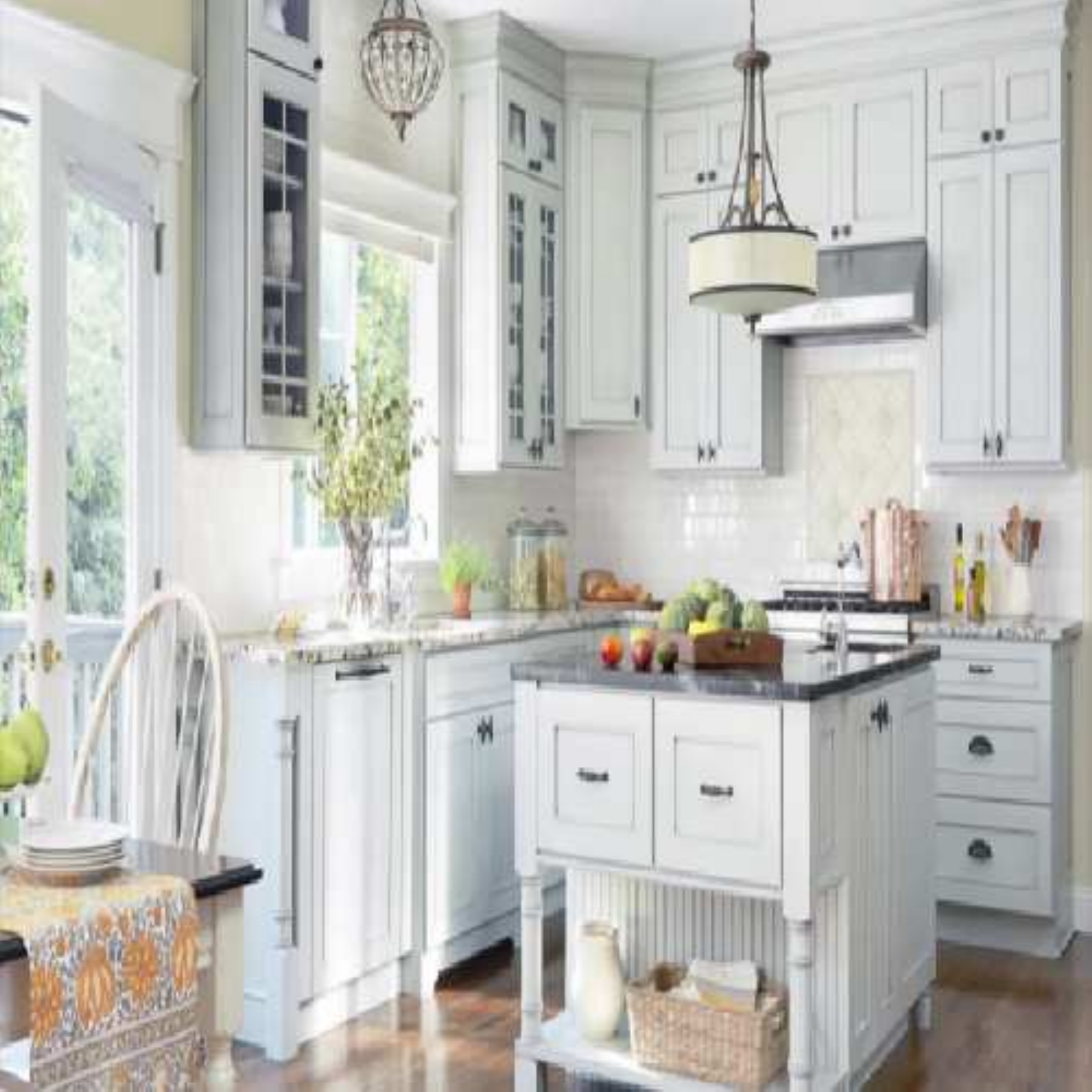}
			\end{minipage}
		}%
		\subfigure[Neural Style]{
			\begin{minipage}[t]{0.32\linewidth}
				\centering
				\includegraphics[width=1.6in]{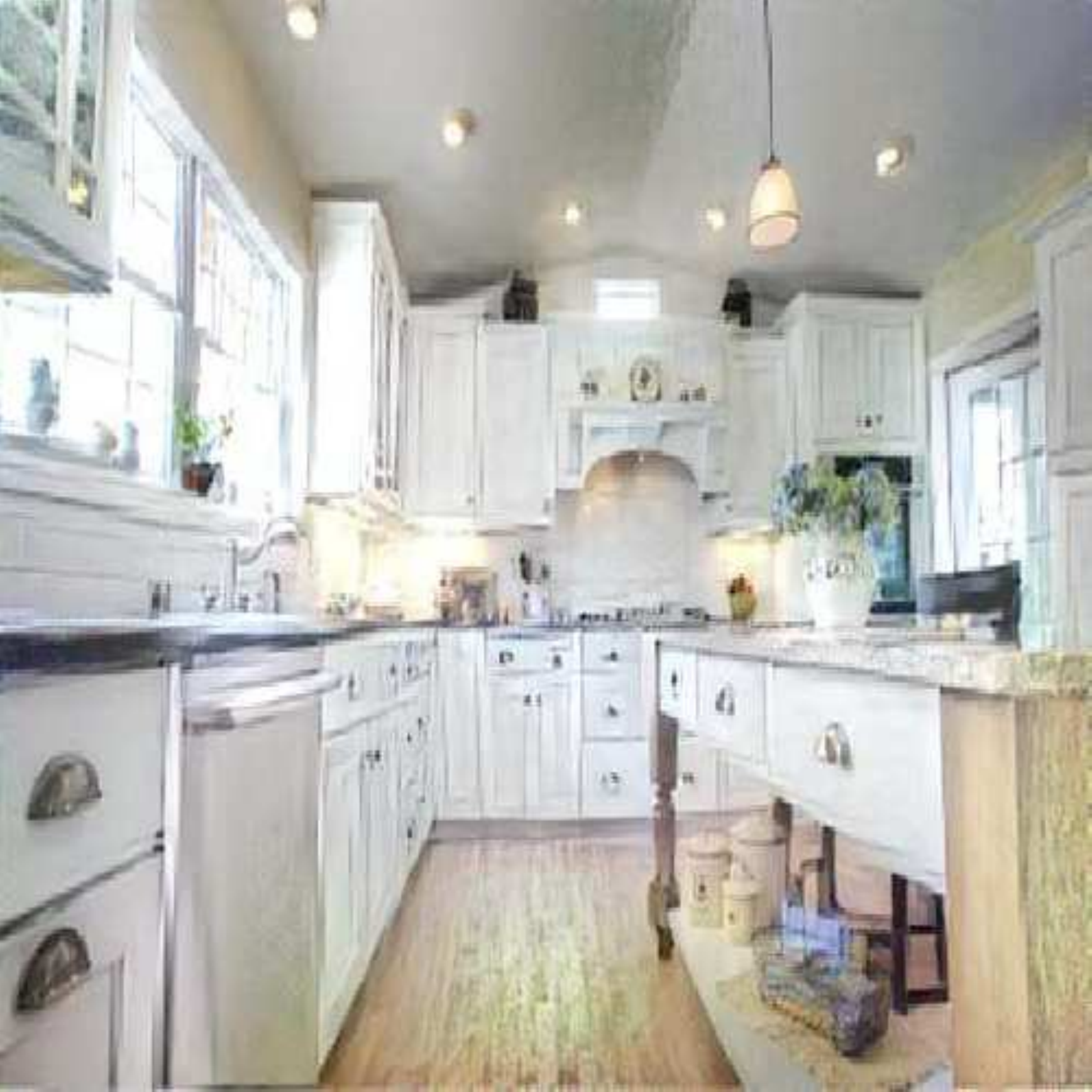}
				\label{5c}
			\end{minipage}
		}%
		
		\subfigure[CNNMRF]{
			\begin{minipage}[t]{0.32\linewidth}
				\centering
				\includegraphics[width=1.6in]{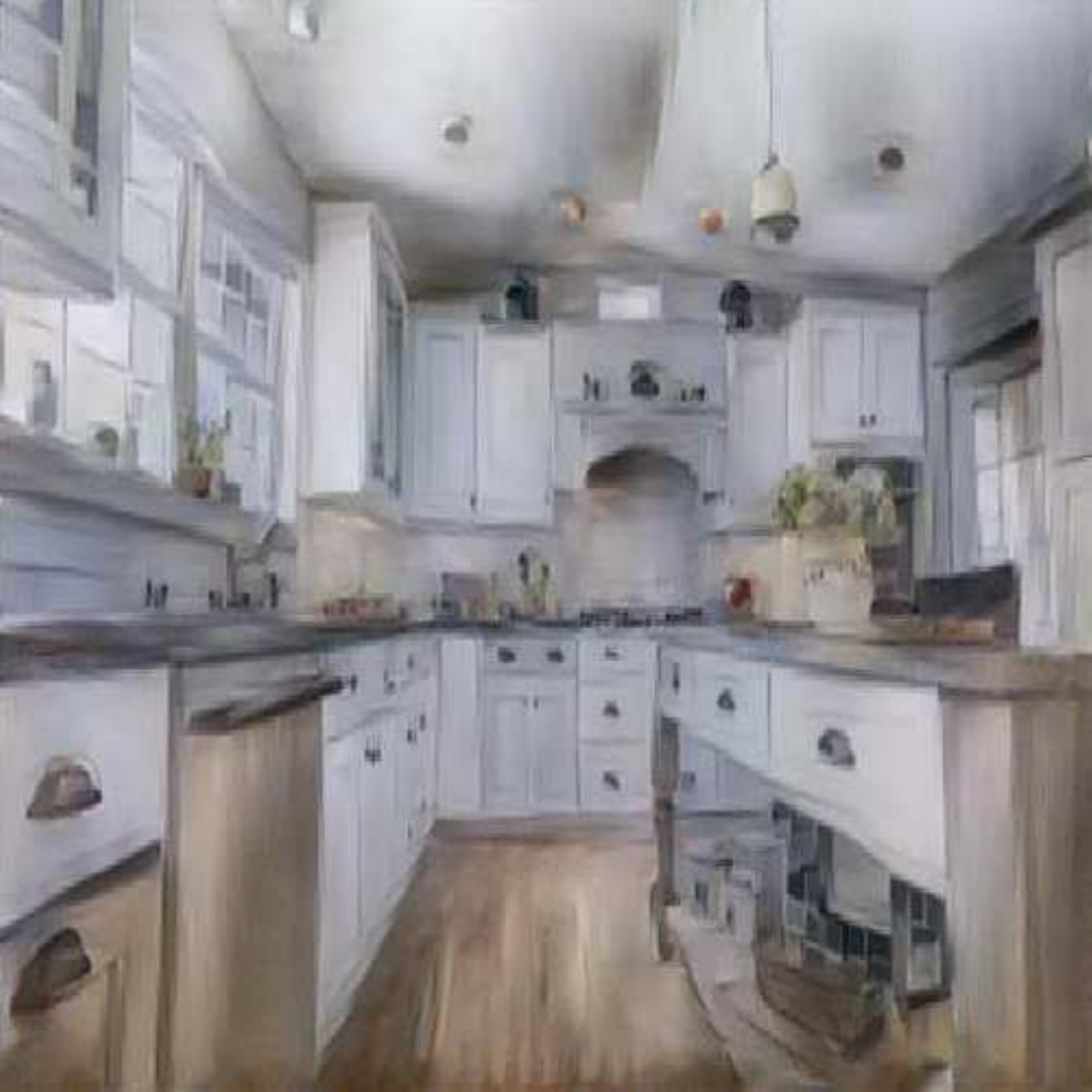}
			\end{minipage}
		}%
		\subfigure[\cite{luan2017deep}]{
			\begin{minipage}[t]{0.32\linewidth}
				\centering
				\includegraphics[width=1.6in]{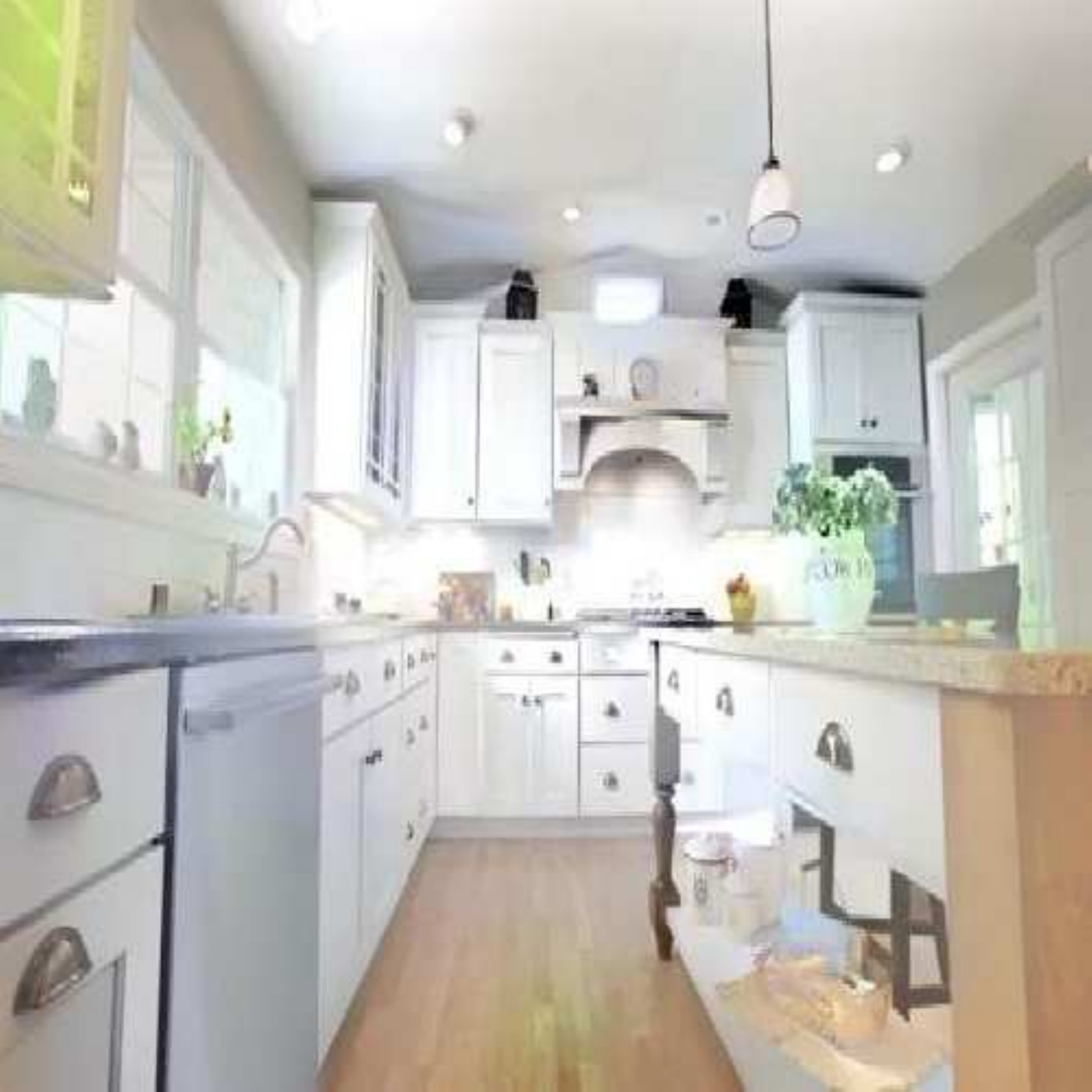}
			\end{minipage}
			\label{5e}
		}%
		\subfigure[Ours]{
			\begin{minipage}[t]{0.32\linewidth}
				\centering
				\includegraphics[width=1.6in]{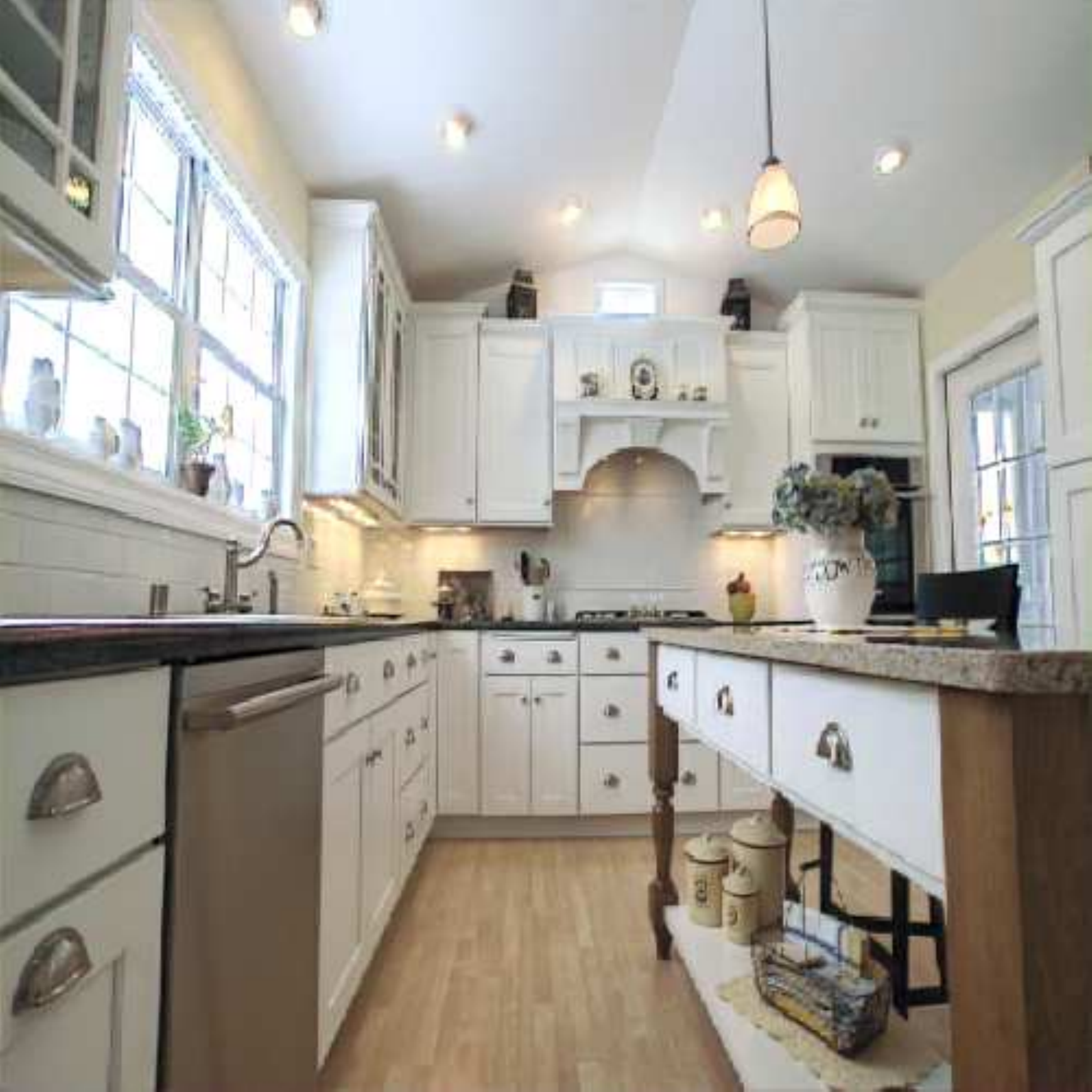}
			\end{minipage}
		}%
		
		\subfigure[Content]{
			\begin{minipage}[t]{0.32\linewidth}
				\centering
				\includegraphics[width=1.6in]{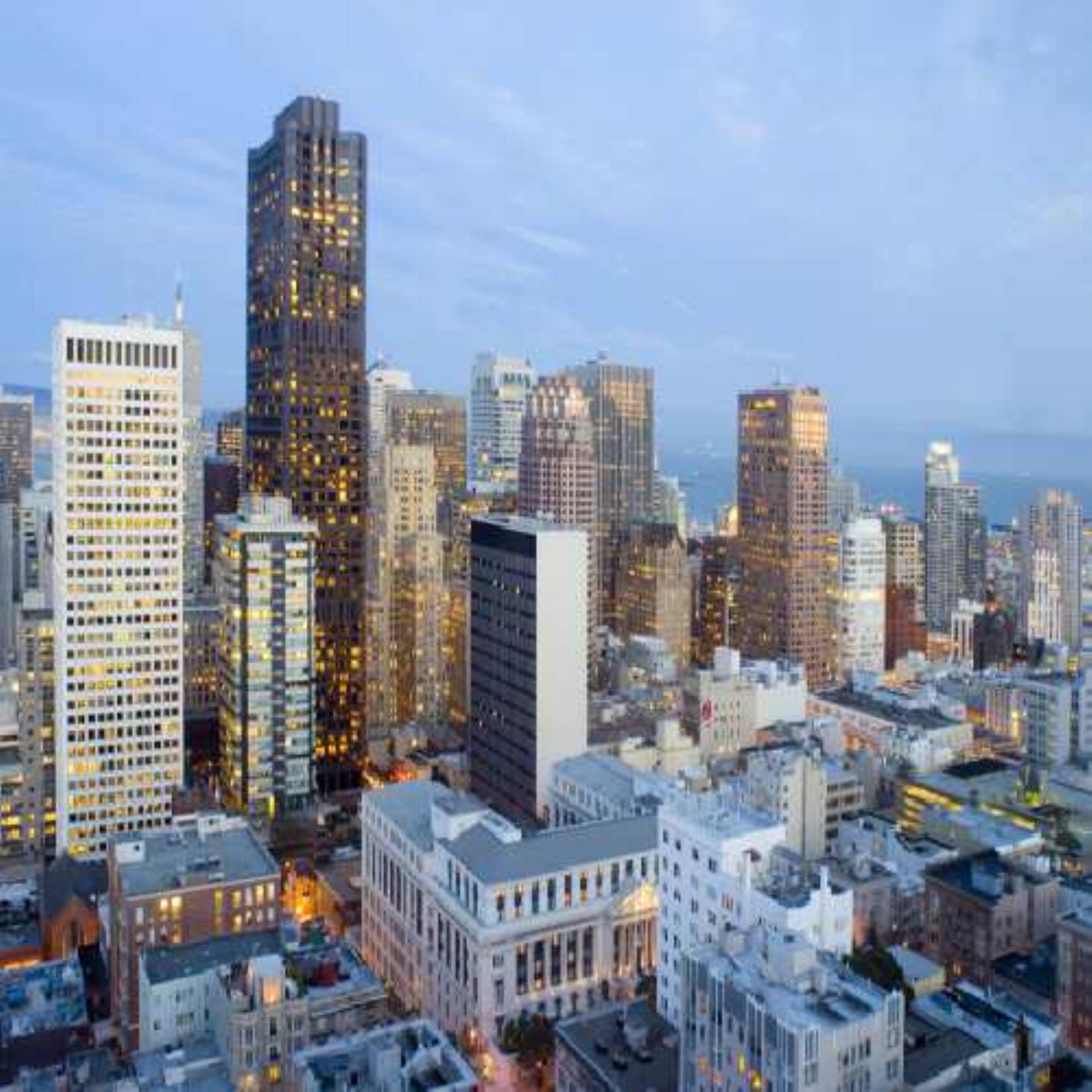}
			\end{minipage}
		}%
		\subfigure[Style]{
			\begin{minipage}[t]{0.32\linewidth}
				\centering
				\includegraphics[width=1.6in]{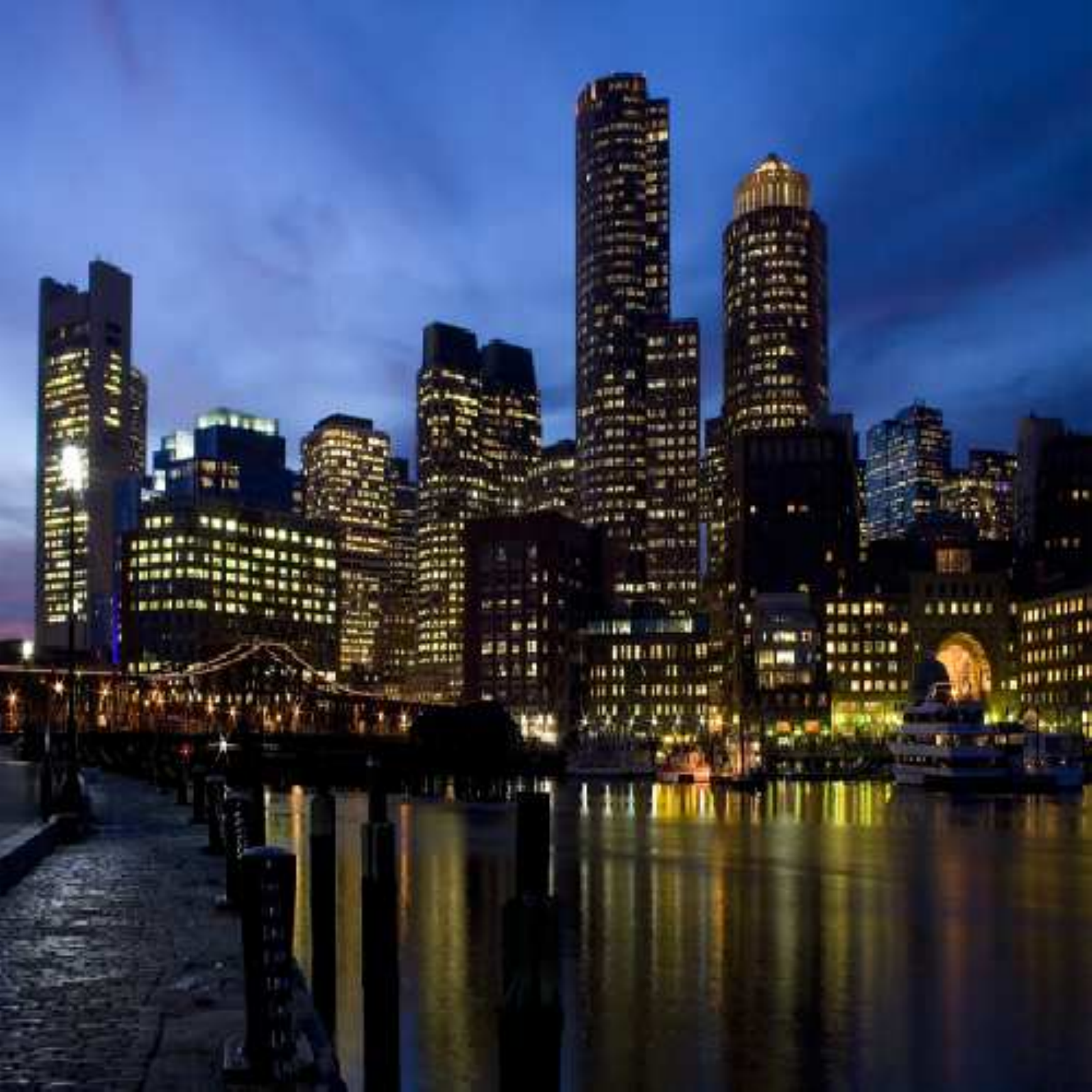}
			\end{minipage}
		}%
		\subfigure[Neural Style]{
			\begin{minipage}[t]{0.32\linewidth}
				\centering
				\includegraphics[width=1.6in]{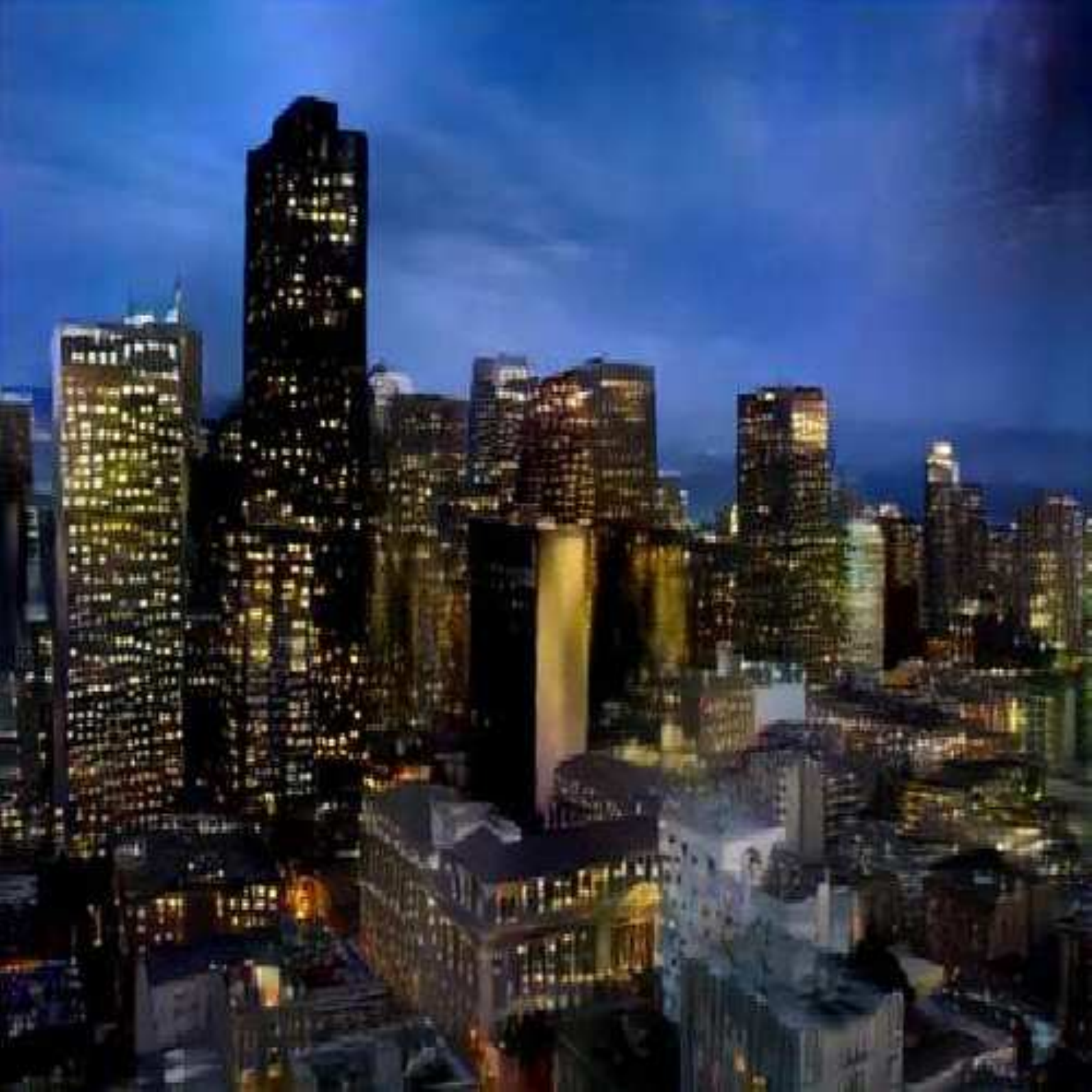}
				\label{5i}
			\end{minipage}
		}%
		
		\subfigure[CNNMRF]{
			\begin{minipage}[t]{0.32\linewidth}
				\centering
				\includegraphics[width=1.6in]{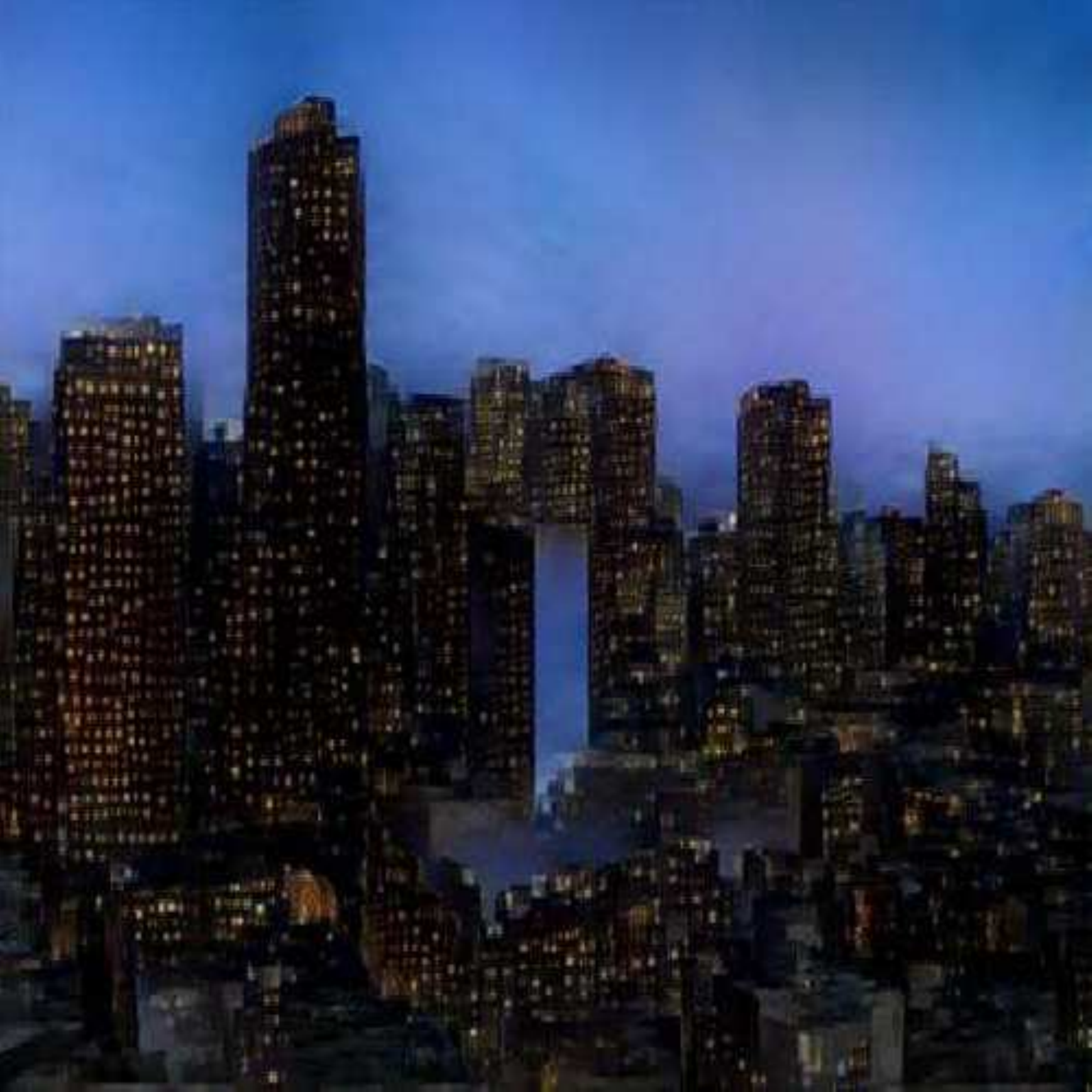}
			\end{minipage}
		}%
		\subfigure[\cite{luan2017deep}]{
			\begin{minipage}[t]{0.32\linewidth}
				\centering
				\includegraphics[width=1.6in]{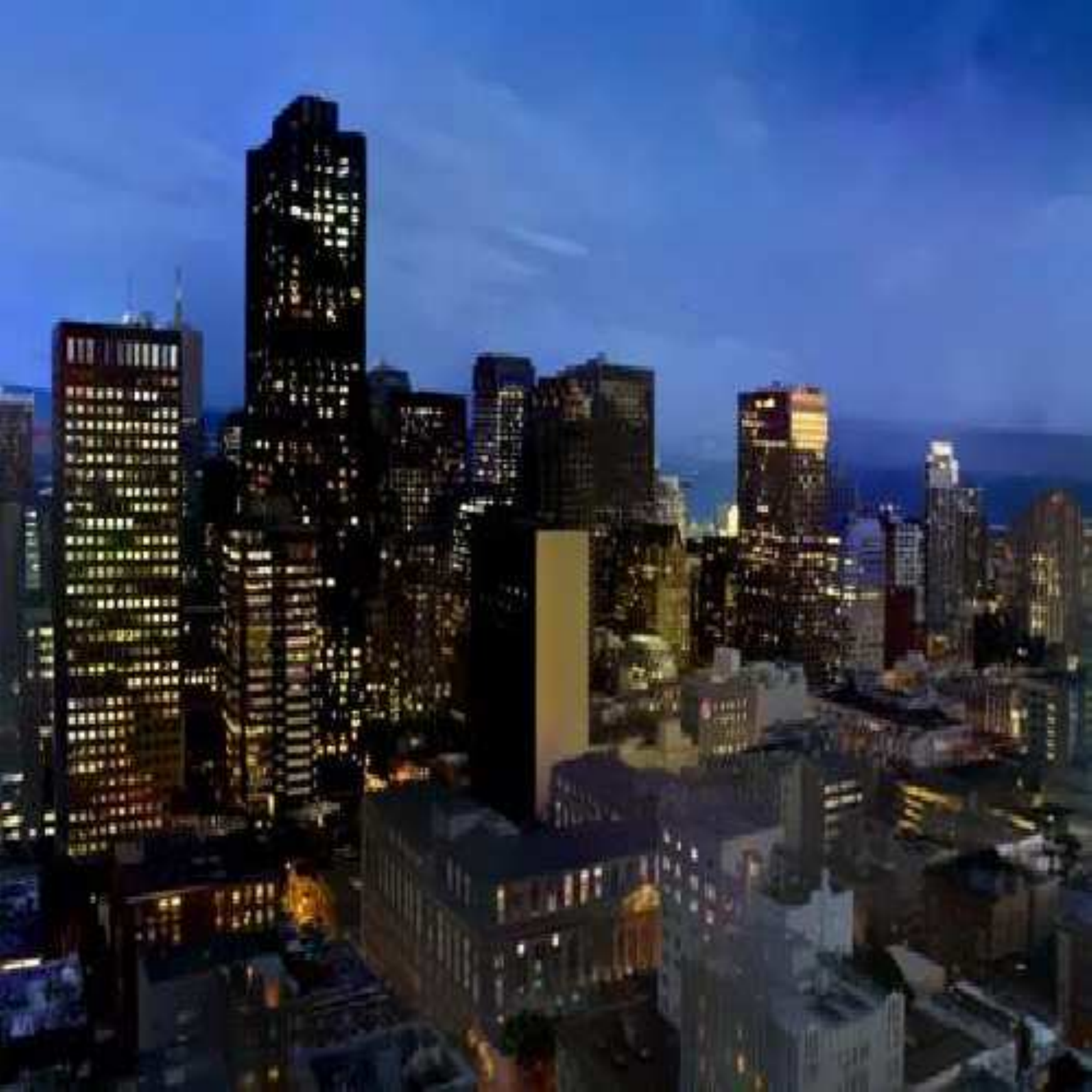}
			\end{minipage}
		}%
		\subfigure[Ours]{
			\begin{minipage}[t]{0.32\linewidth}
				\centering
				\includegraphics[width=1.6in]{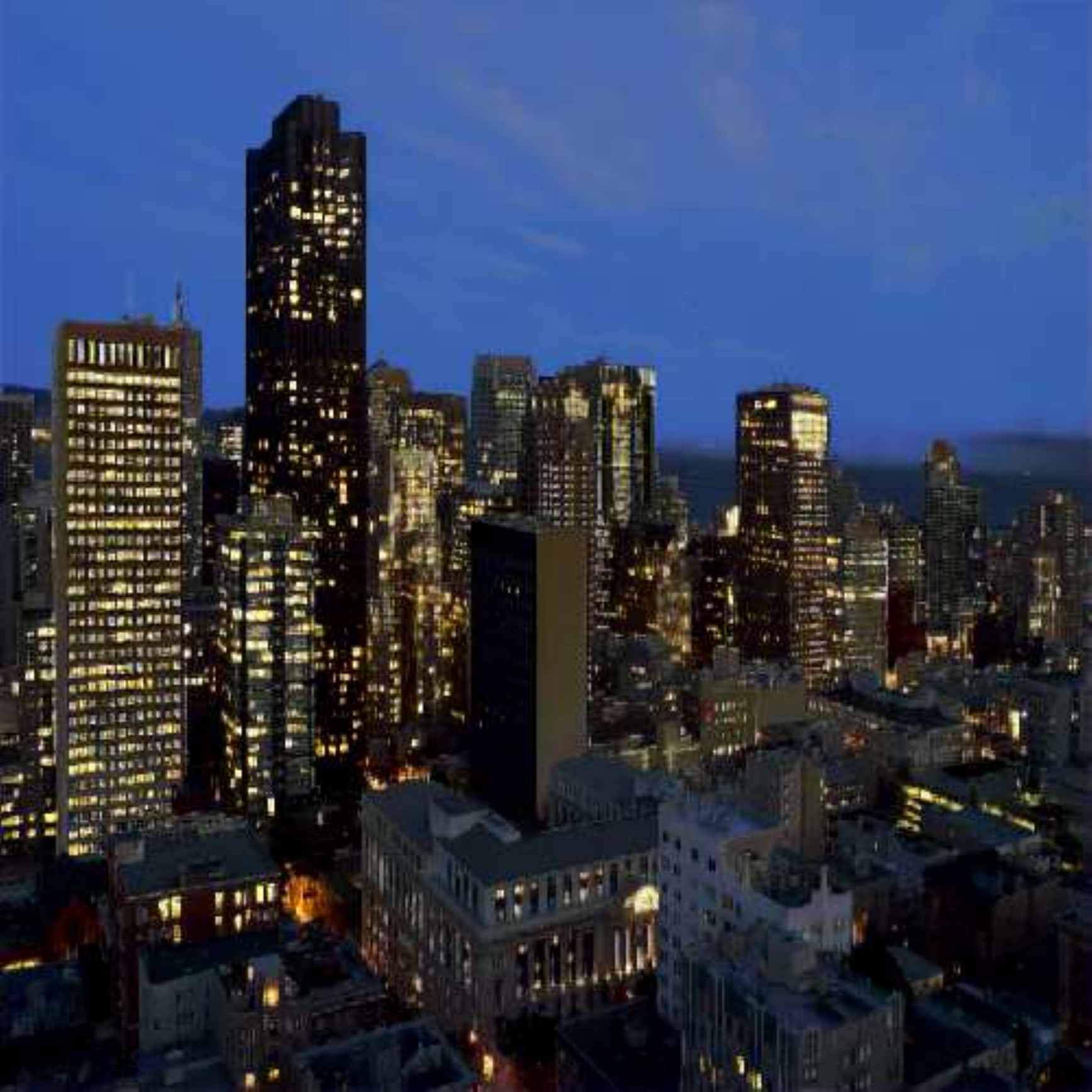}
			\end{minipage}
		}%
                
                \caption{Comparison of our method against Neural Style, CNNMRF
                and \cite{luan2017deep}. Both neural transfer and CNNMRF
                produce image distortion in the output image. The neural
                transfer also completely ignores the semantic information of
                content image. Compared to Luan, our method has a more
                elaborate structure and a more realistic color distribution.}
                \label{fig5} 
            \end{center} 
        \end{figure}

\begin{figure}[htp]
	\begin{center}
		\subfigure[Content]{
			\begin{minipage}[t]{0.32\linewidth}
				\centering
				\includegraphics[width=1.6in]{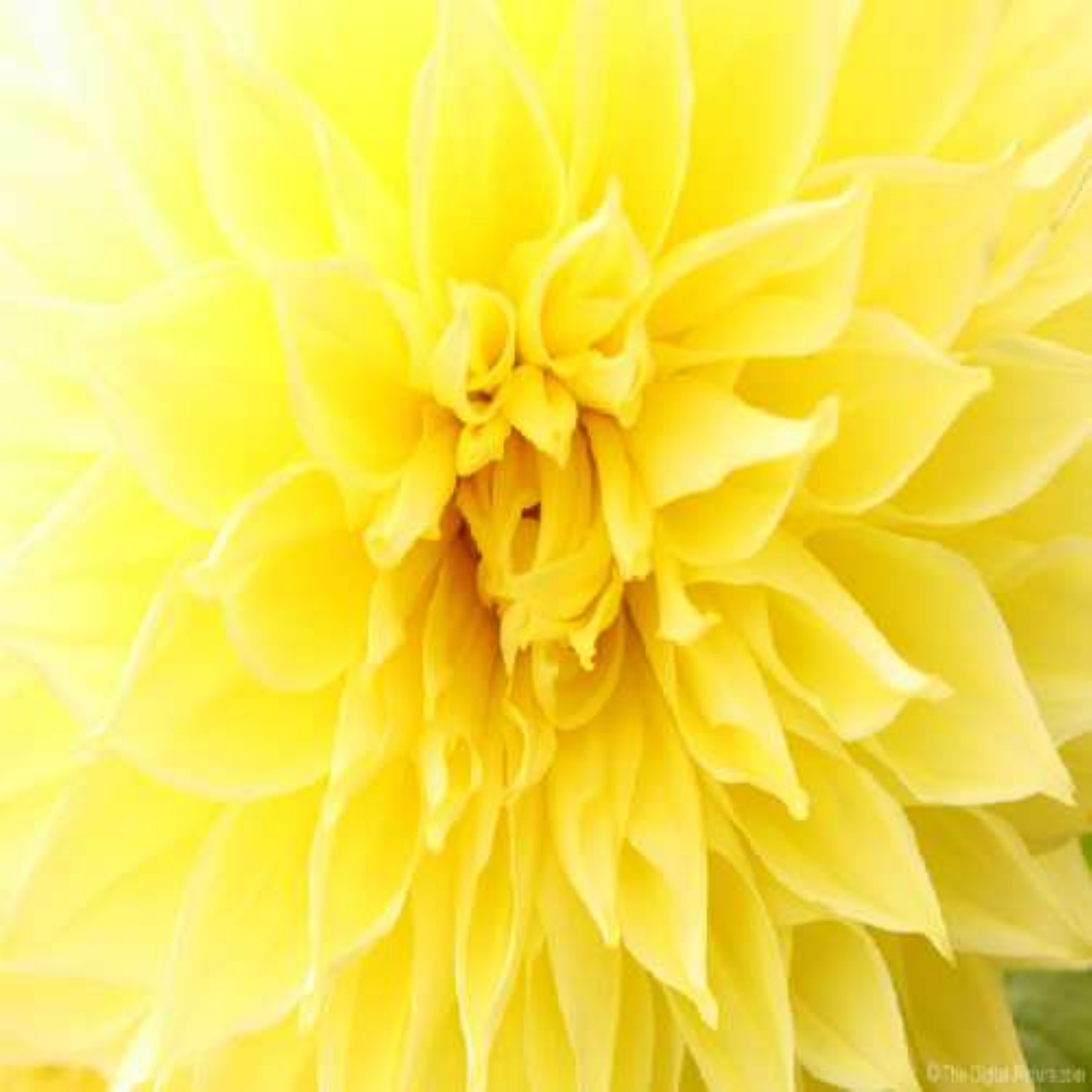}
			\end{minipage}
		}%
		\subfigure[Style]{
			\begin{minipage}[t]{0.32\linewidth}
				\centering
				\includegraphics[width=1.6in]{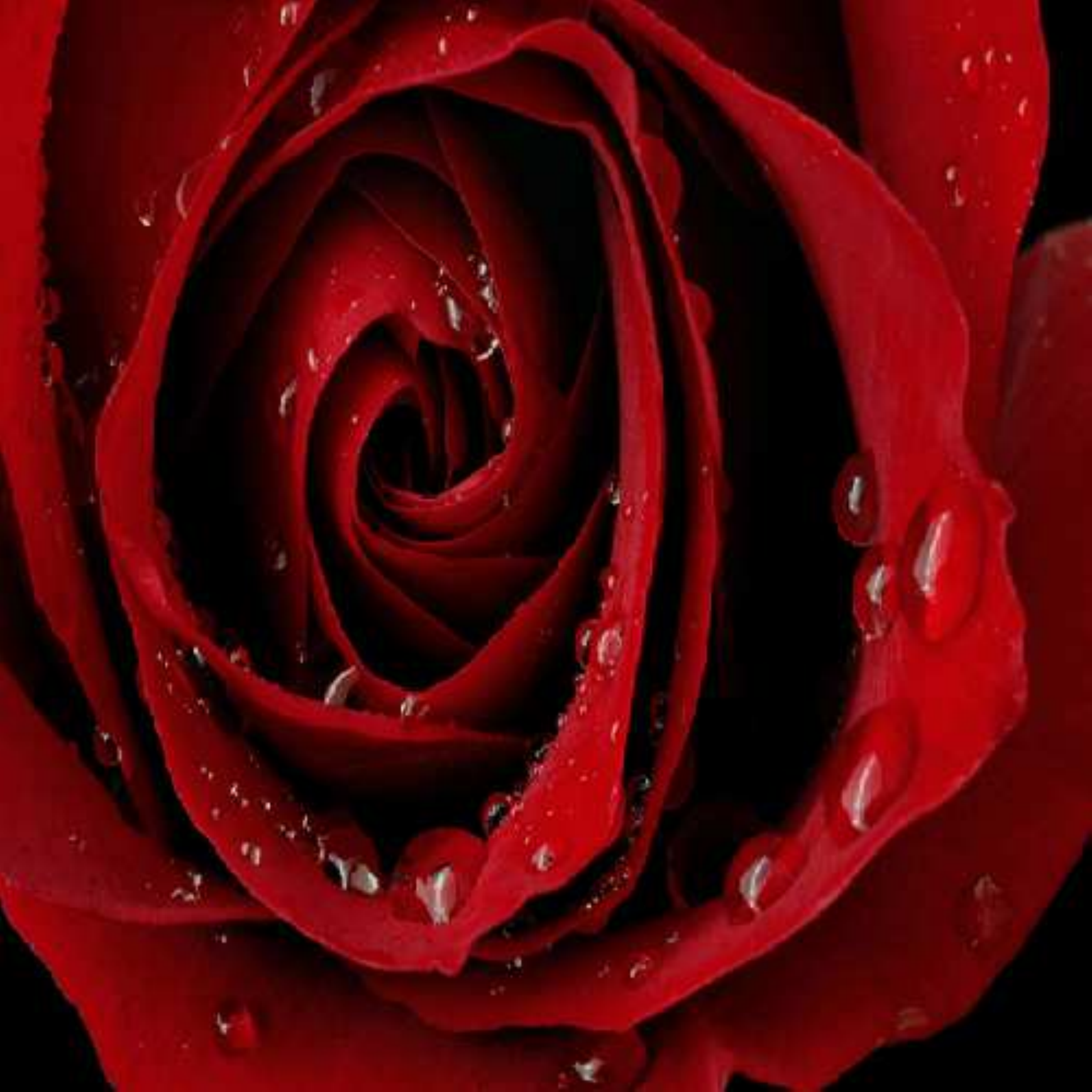}
			\end{minipage}
		}%
		\subfigure[\cite{reinhard2001color}]{
			\begin{minipage}[t]{0.32\linewidth}
				\centering
				\includegraphics[width=1.6in]{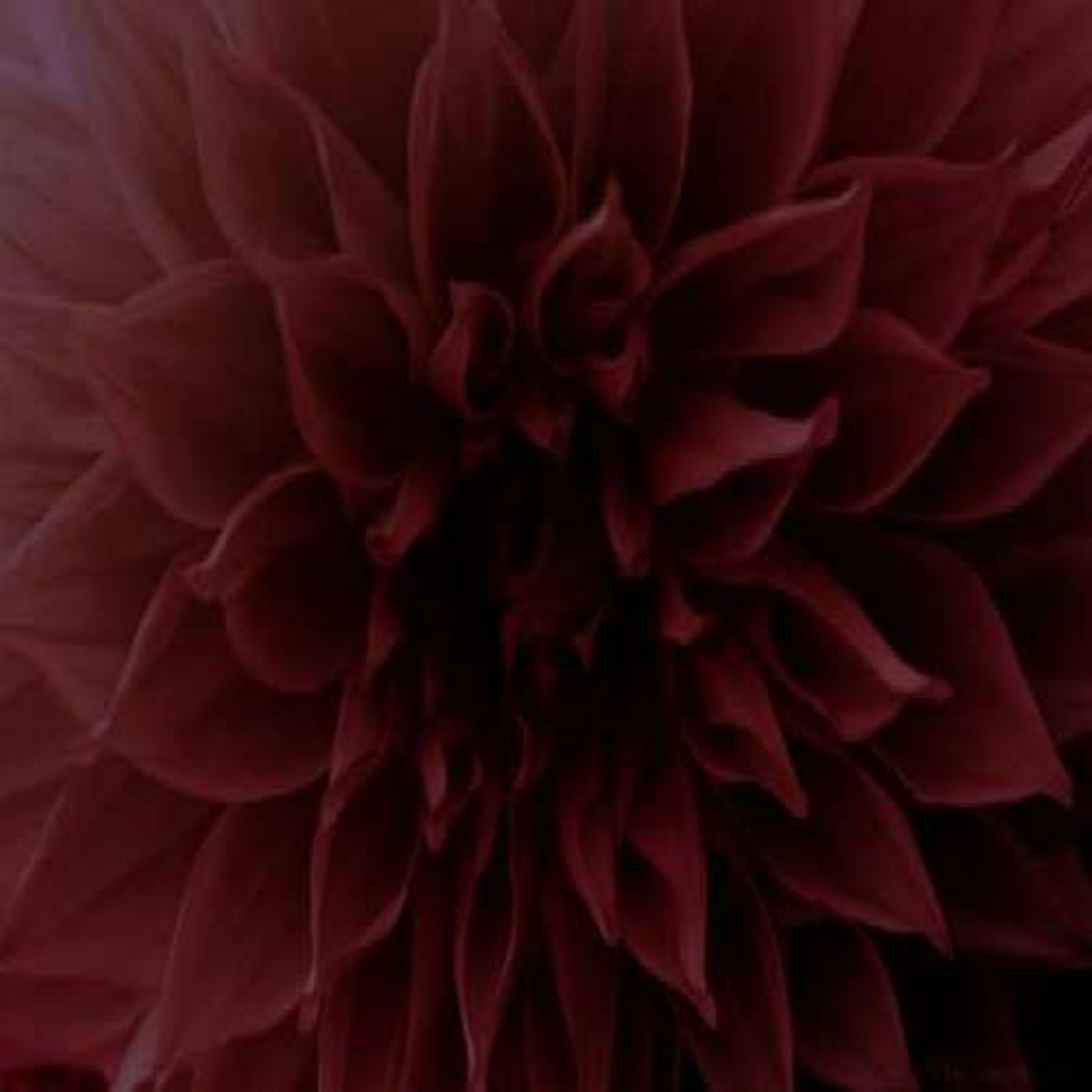}
			\end{minipage}
		}%
		
		\subfigure[\cite{pitie2005n}]{
			\begin{minipage}[t]{0.32\linewidth}
				\centering
				\includegraphics[width=1.6in]{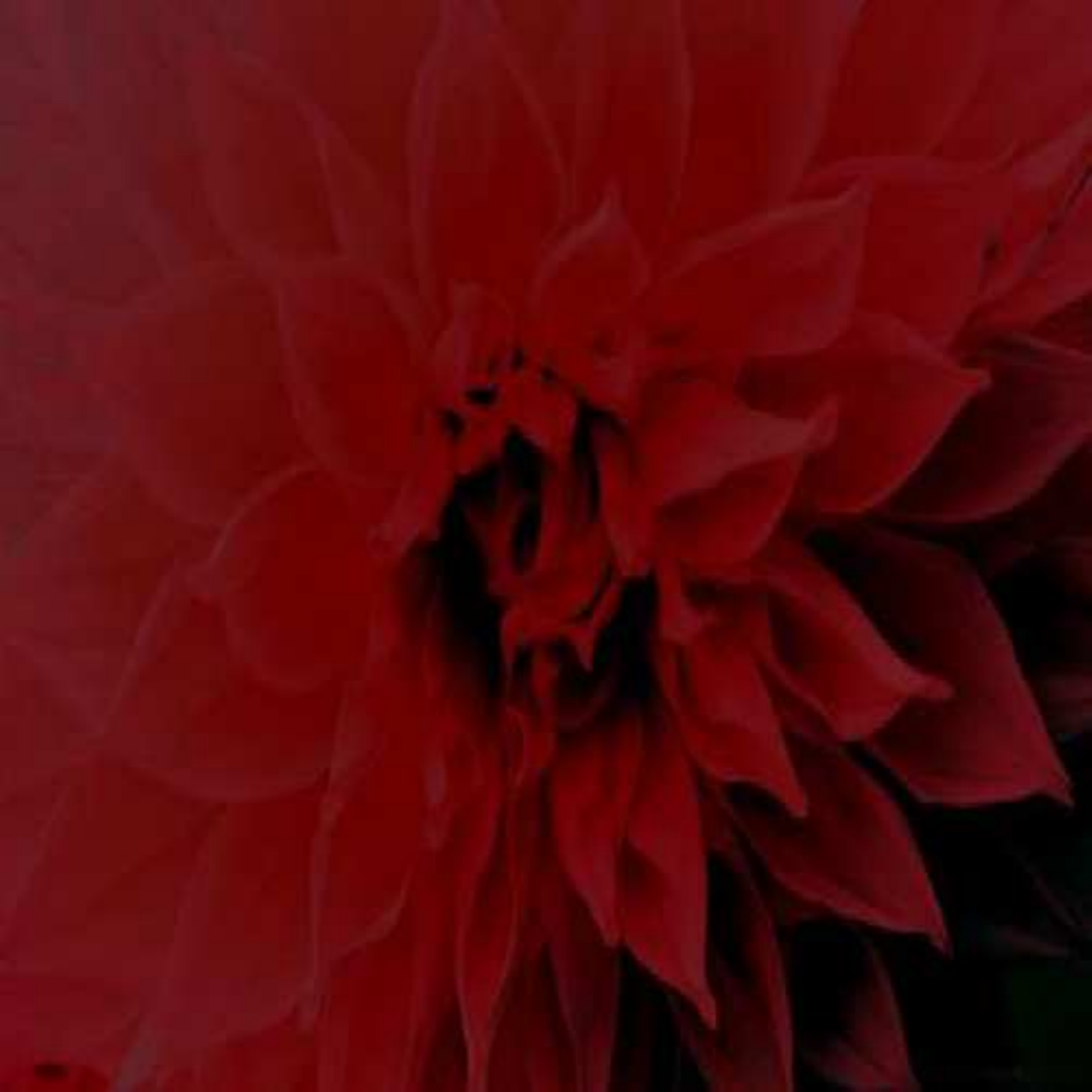}
			\end{minipage}
		}%
		\subfigure[\cite{luan2017deep}]{
			\begin{minipage}[t]{0.32\linewidth}
				\centering
				\includegraphics[width=1.6in]{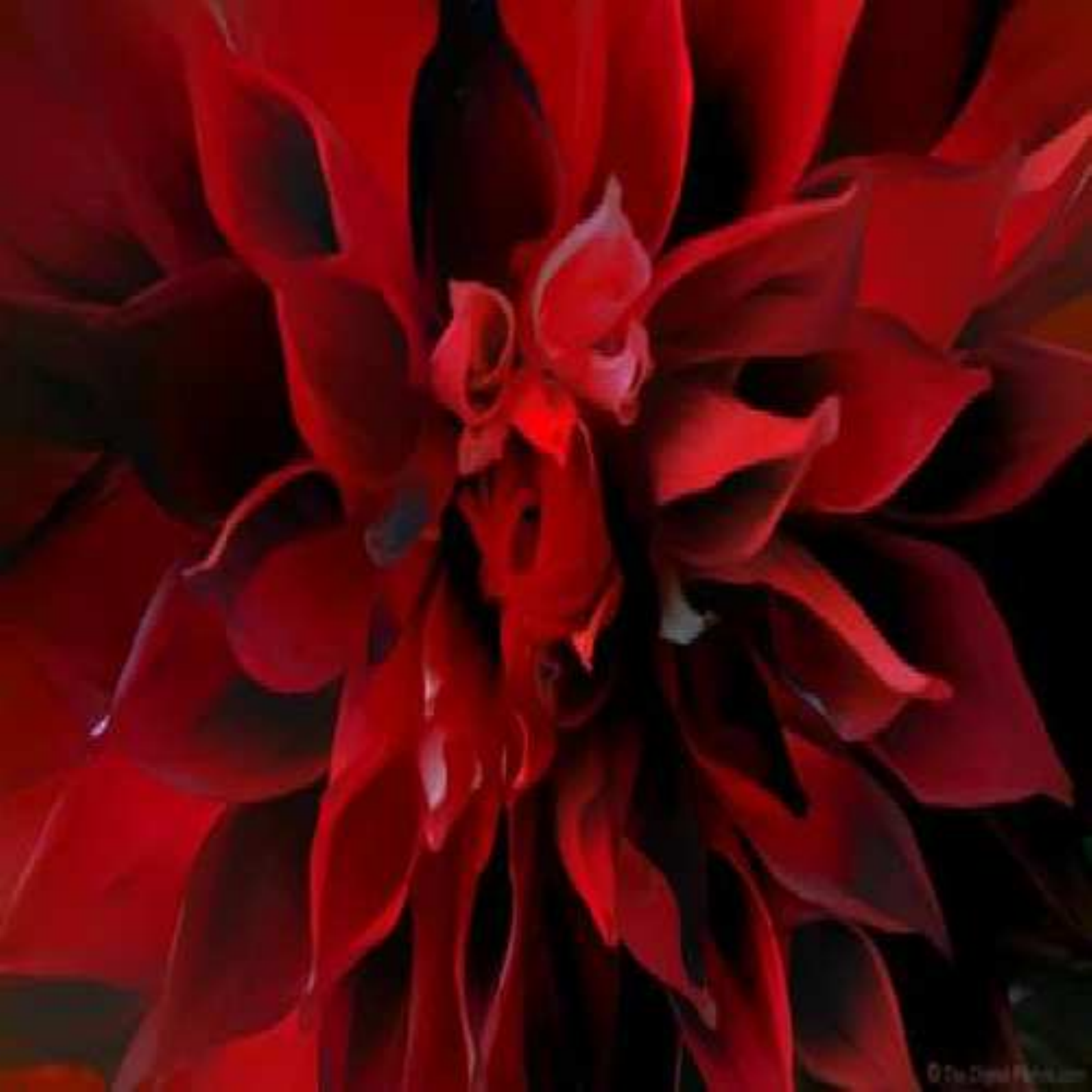}
				\label{6e}
			\end{minipage}
		}%
		\subfigure[Ours]{
			\begin{minipage}[t]{0.32\linewidth}
				\centering
				\includegraphics[width=1.6in]{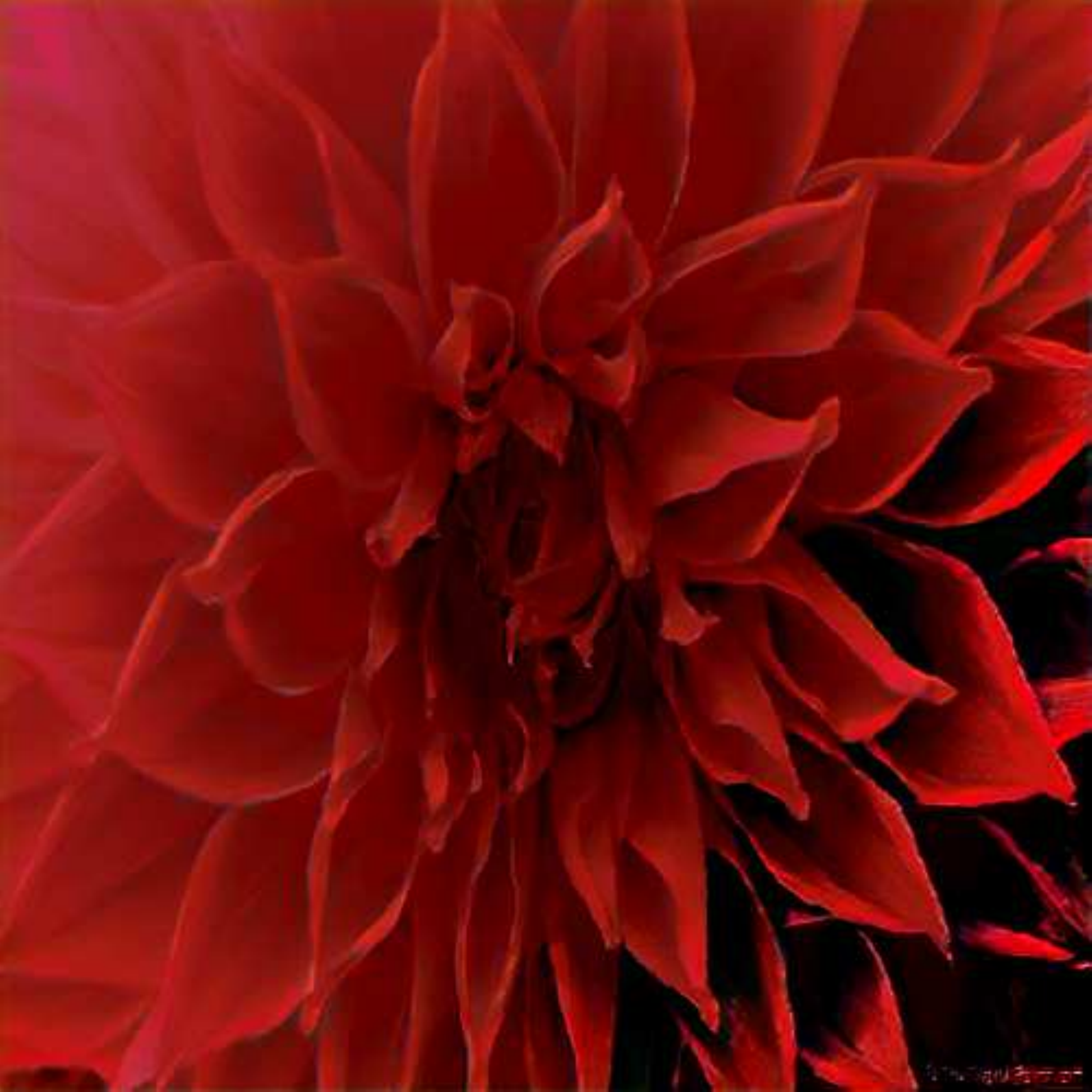}
			\end{minipage}
		}%
		
		\subfigure[Content]{
			\begin{minipage}[t]{0.32\linewidth}
				\centering
				\includegraphics[width=1.6in]{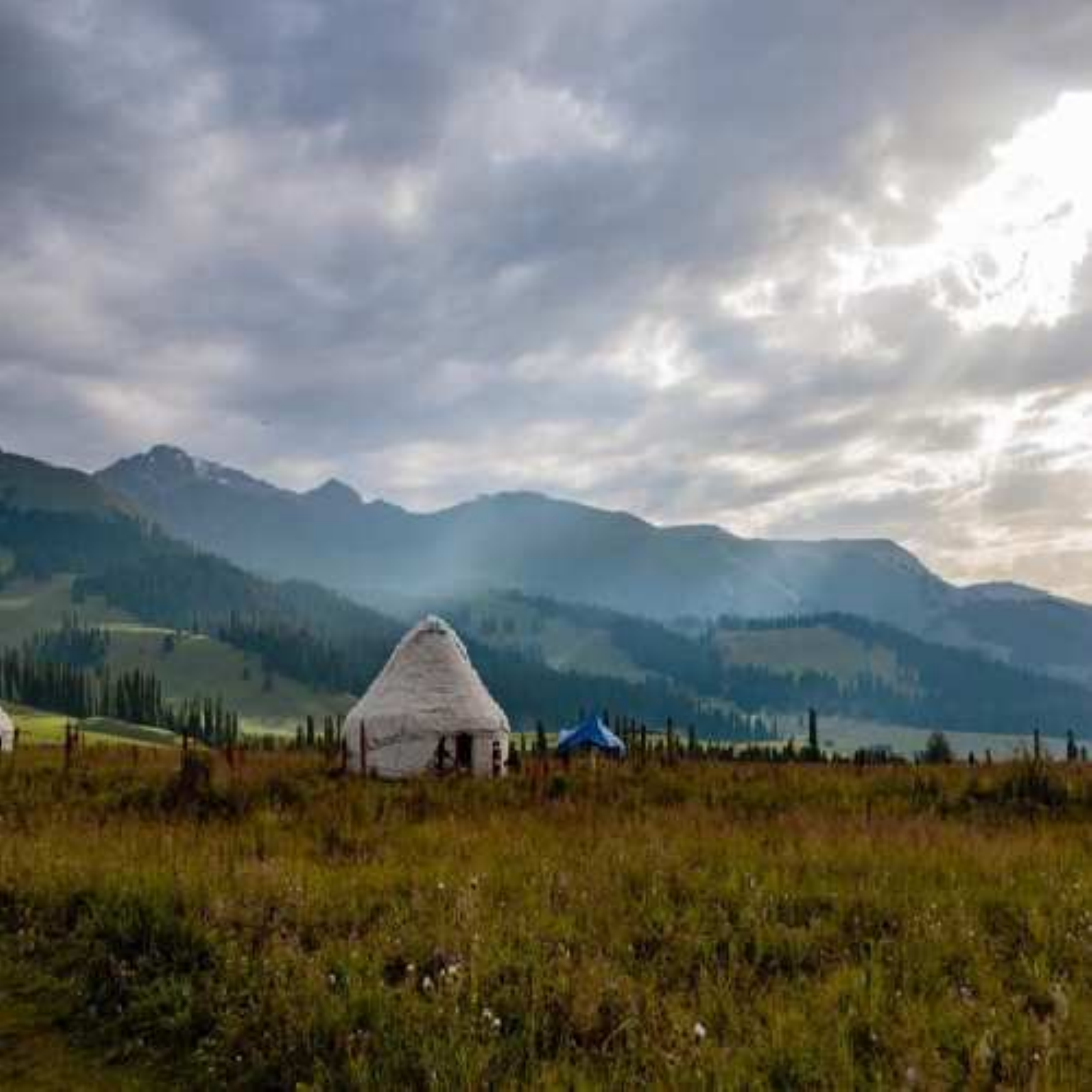}
			\end{minipage}
		}%
		\subfigure[Style]{
			\begin{minipage}[t]{0.32\linewidth}
				\centering
				\includegraphics[width=1.6in]{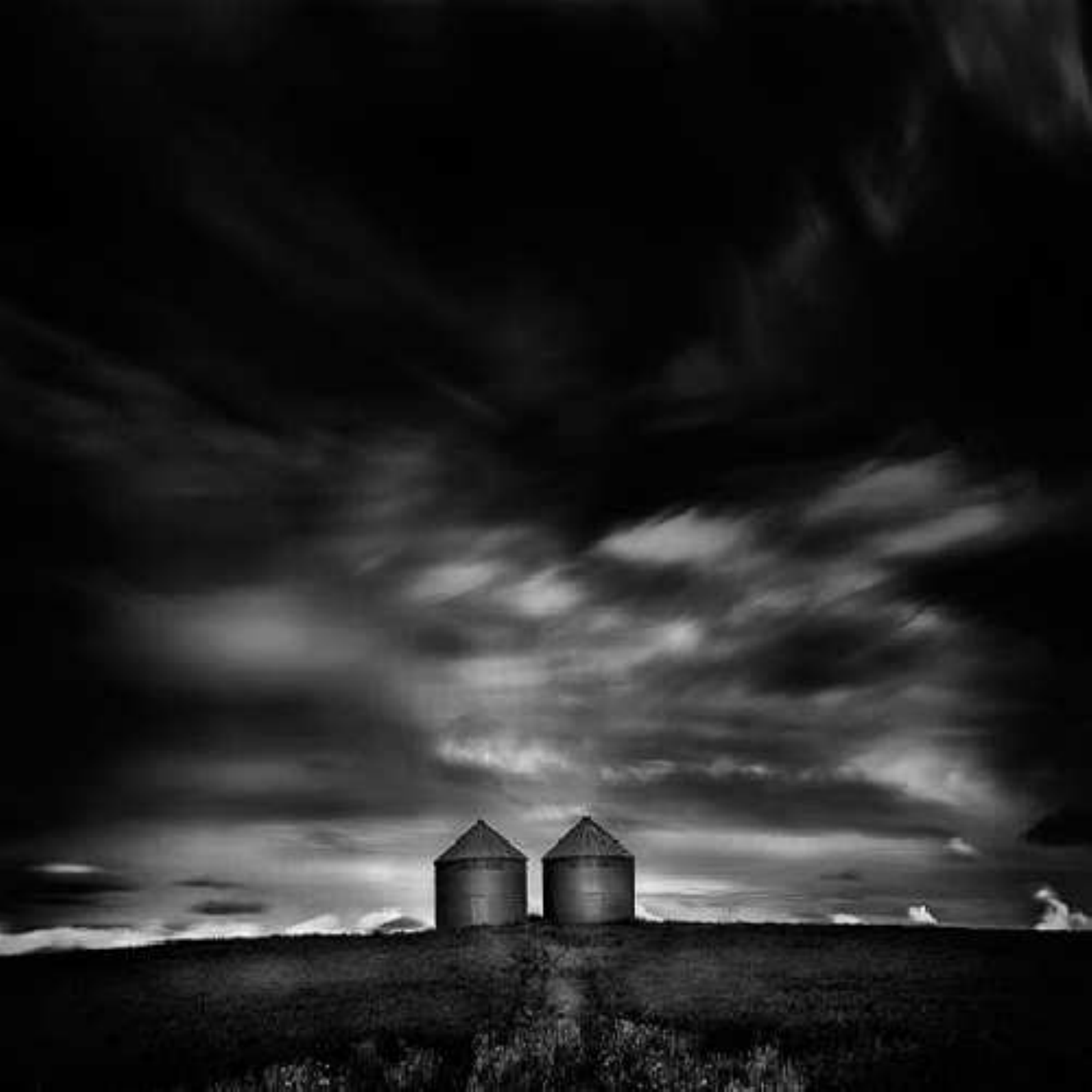}
			\end{minipage}
		}%
		\subfigure[\cite{reinhard2001color}]{
			\begin{minipage}[t]{0.32\linewidth}
				\centering
				\includegraphics[width=1.6in]{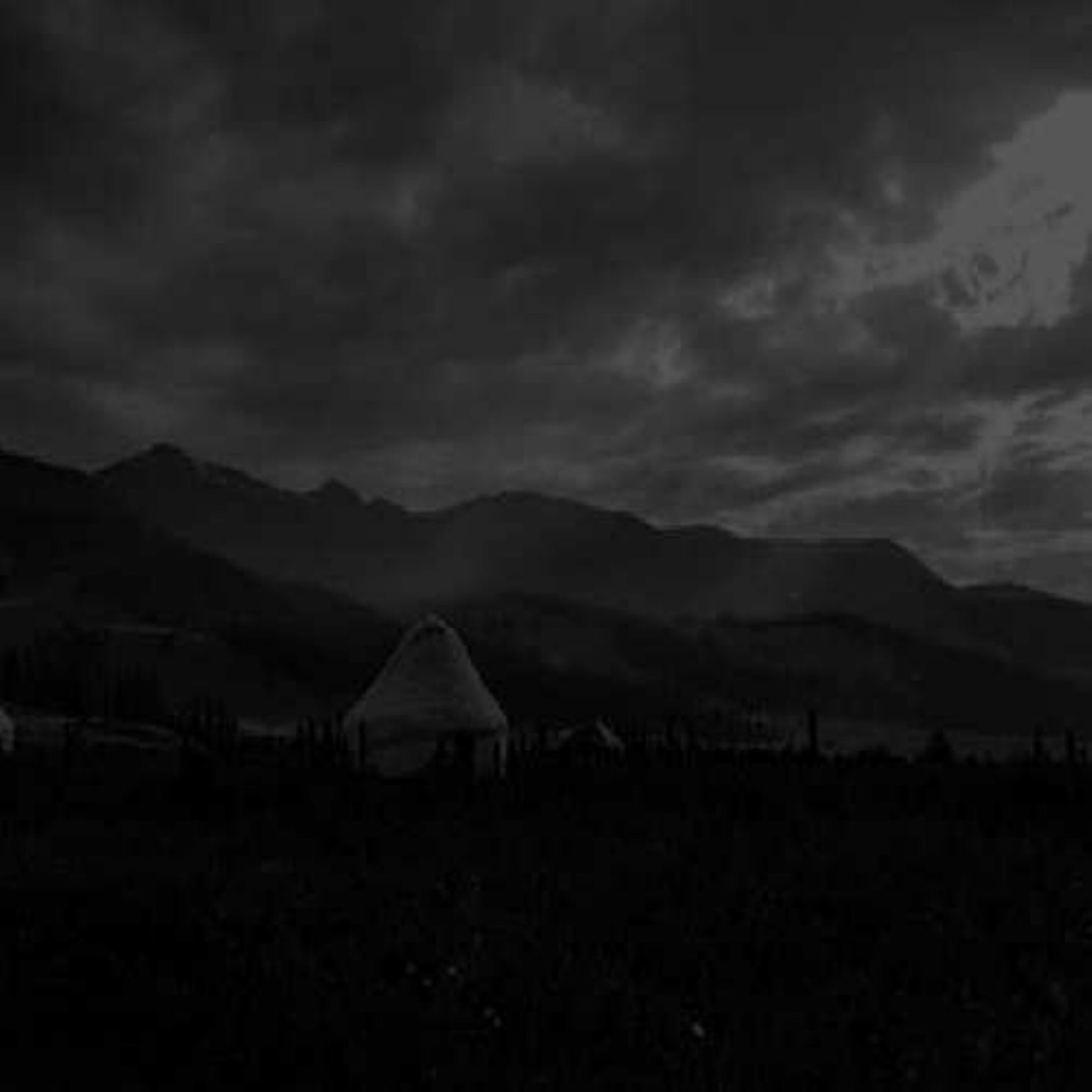}
			\end{minipage}
		}%
		
		\subfigure[\cite{pitie2005n}]{
			\begin{minipage}[t]{0.32\linewidth}
				\centering
				\includegraphics[width=1.6in]{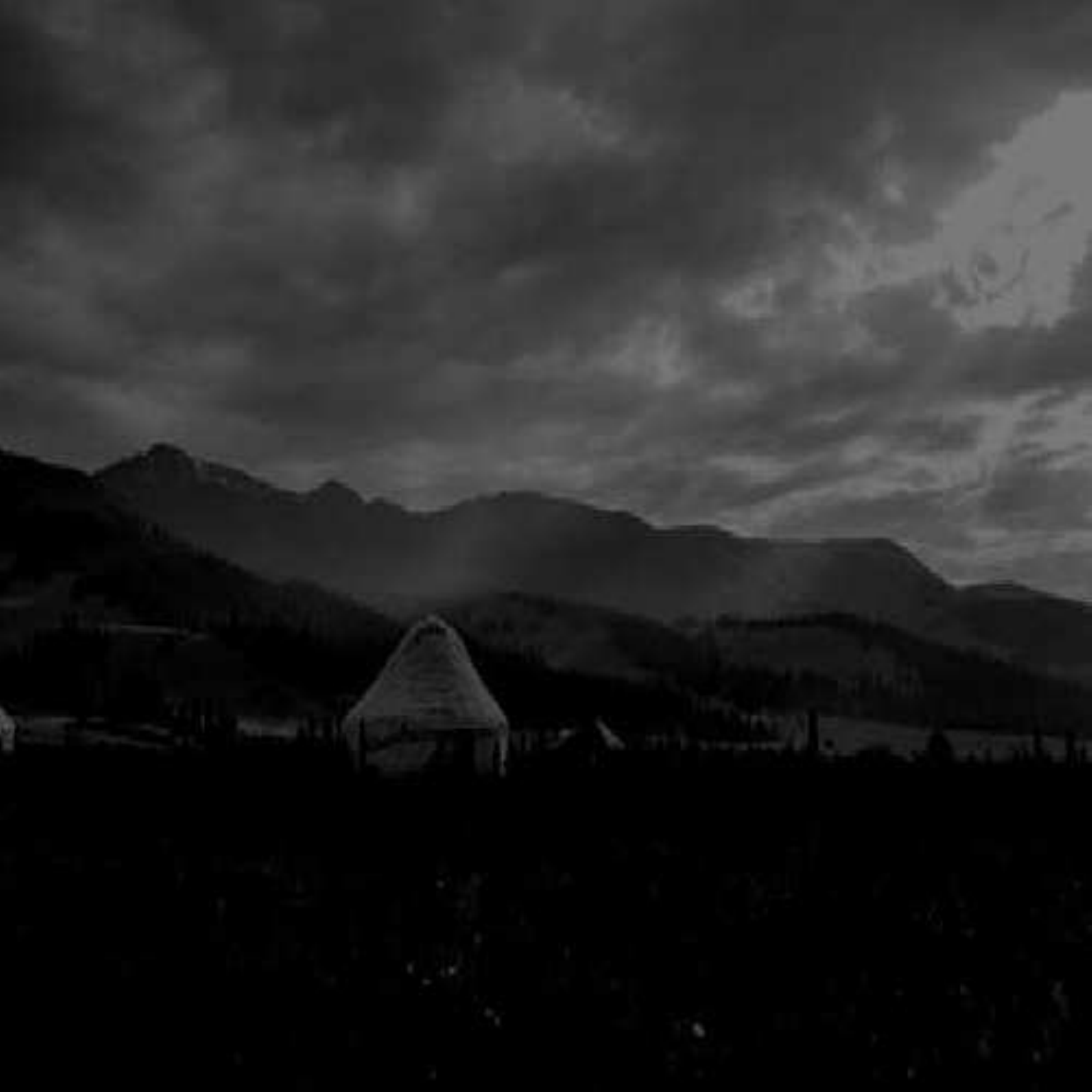}
			\end{minipage}
		}%
		\subfigure[\cite{luan2017deep}]{
			\begin{minipage}[t]{0.32\linewidth}
				\centering
				\includegraphics[width=1.6in]{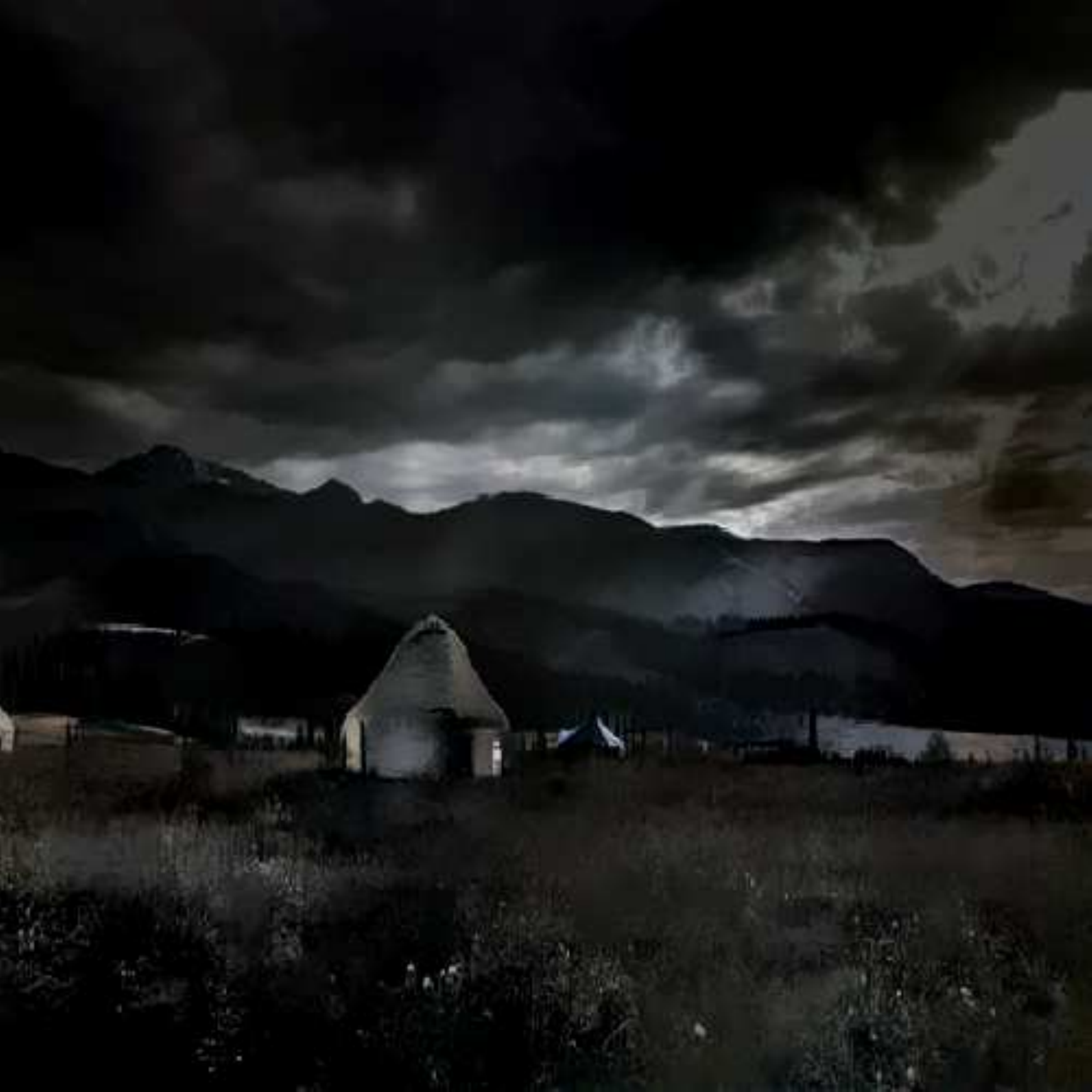}
			\end{minipage}
		}%
		\subfigure[Ours]{
			\begin{minipage}[t]{0.32\linewidth}
				\centering
				\includegraphics[width=1.6in]{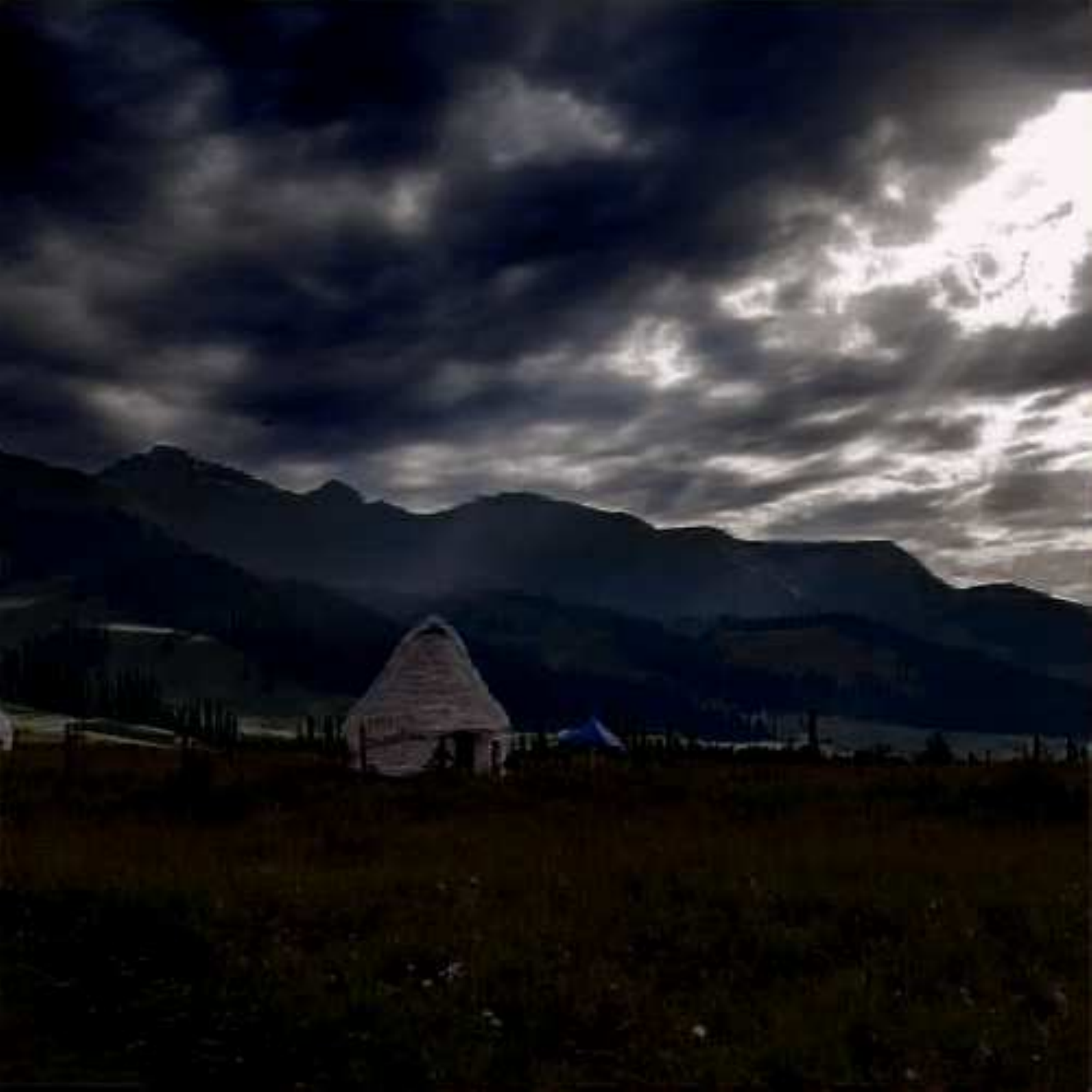}
			\end{minipage}
		}%
		
                \caption{Our approach provides greater flexibility in
                delivering spatially varying color variations and produces
                better results than \cite{reinhard2001color} and
                \cite{pitie2005n}. Compared to \cite{luan2017deep}, our output
                image has a more uniform color distribution and better image
                detail (e.g., the white tent).} \label{fig6} 
            \end{center}
        \end{figure}

\subsection{Results}

The experimental results show that our output images have a contour closer to
the content image, indicating that our results have more semantic information.
The speed comparison shows that our algorithm runs faster than
\cite{luan2017deep} and the empirical study shows that more people regard our
results as more faithful. Detailed Comparison are illustrated as follows:

\subparagraph{Effect Comparison} We compare our approach to
\cite{gatys2016image}(Neural style transfer) and \cite{li2016combining}
(CNNMRF). As shown in Fig.~\ref{fig5}, both of these techniques produce
results similar to the distortion of the painting, which are carried out in the
context of the shift in photographic style. In several cases, the neural style
algorithm is also affected by the spillover effect, CNNMRF often generates
partial style transfers, ignoring important parts of the style image.  In
contrast, our images are generated by a high-resolution generation network that
prevents these artifacts from happening, furthermore, our results look more
like real-life photos. Compared to Luan's proposed method, our output image has
a more uniform color distribution, which makes our images have more realistic
semantic information (e.g. Closet color in the upper left corner in Fig.
\ref{5e}.)

We also compare our approach to global style delivery methods that do not
distort the image~\cite{reinhard2001color} and \cite{pitie2005n}.  As shown in
Fig. \ref{fig6}, both their techniques apply a global color map to match the
color statistics between the input image and the style image, which limits the
loyalty of the results when transmitting a color transform that requires
spatial variation. \cite{luan2017deep} achieves a good stylized effect at
first glance, however, careful observation revealed that the resulting
photograph contained significant artifacts, e.g. the irregular shape of petals
and white tent. Several semantically similar regions are stylized
inconsistently. In contrast, our approach is more capable of retaining semantic
information in content images while successfully transferring styles.

\begin{table}[bt] 
\begin{center}
	
       \caption{Run-time comparison. We calculated the running time of these
       algorithms at different resolutions (in seconds) with a GTX 1060 GPU,
       cuda 9.0 and cuDNN \cite{chetlur2014cudnn}.} 

       \begin{tabular}{cccc}
		\toprule
		\cite{gatys2016image} & \cite{johnson2016perceptual} & \cite{luan2017deep} & Ours\\
		\midrule 
		\textbf{13.14}(128$\times$128) & 83.74(128$\times$128) & 199.01(256$\times$128) & 59.72(128$\times$128)\\
		\textbf{39.68}(256$\times$256) & 148.99(256$\times$256) & 466.3(512$\times$256)  & 110.91(256$\times$256)\\
		\textbf{126.84}(512$\times$512) & 434.18(512$\times$512) & 952.05(768$\times$384) & 380.66(512$\times$512)\\
		\bottomrule 
		\label{table1}
	\end{tabular} 
	\vspace{-5mm}
\end{center}
\end{table}

\subparagraph{Speed Comparison} As can be seen from Table \ref{table1}, our
method is much faster than \cite{johnson2016perceptual} and
\cite{luan2017deep}. The numbers in parentheses represent the resolution of the
style image and the content image.  Since \cite{luan2017deep} method is
special, we did not reproduce their work, but estimated the computing power of
different GPUs used. The estimation and comparison of computing power between
different GPUs comes from the official website of NVIDIA. Compared with the
method proposed by \cite{gatys2016image}, although our method requires more
time to train once, the number of trainings required by their method to achieve
good results is much greater than the number of trainings required by our
method, and their approach is prone to distortion as shown in Fig. \ref{5c}
and \ref{5i}.

\begin{table}[bt] 
	\begin{center}
		\caption{User preference score comparison with \cite{reinhard2001color} and \cite{pitie2005n}:} 
		\begin{tabular}{cccc}
			\toprule
			& \cite{reinhard2001color} & \cite{pitie2005n} & \textbf{Ours}\\
			\midrule 
			More style information  & \textbf{46.83\%} & 13.88\% & 39.29\%\\
			More semantic information & 20.42\% & 26.31\% & \textbf{53.27\%}\\
			Better visual effect  & 17.27\% & 19.92\% & \textbf{62.81\%}\\
			\bottomrule 
			\label{table2}
		\end{tabular} 
		\vspace{-5mm}
	\end{center}
\end{table}

\begin{table}[bt] 
	\begin{center}
	
                \caption{User preference score comparison with
                \cite{gatys2016image}(Neural Style),
                \cite{li2016combining}(CNNMRF) and \cite{luan2017deep}:} 
	
               \begin{tabular}{ccccc}
			\toprule
			 & Neural Style & CNNMRF & \cite{luan2017deep} & \textbf{Ours}\\
			\midrule 
			More style information  & 23.65\% & 4.30\% & 34.41\% & \textbf{38.04\%}\\
			More semantic information  & 11.83\% & 8.61\% & 25.80\% & \textbf{53.76\%}\\
			Better visual effect  & 17.20\% & 3.23\% & 29.03\% & \textbf{50.54\%}\\
			\bottomrule 
			\label{table3}
		\end{tabular} 
		\vspace{-5mm}
	\end{center}
\end{table}

\subparagraph{Visual Comparison} Three evaluation indicators, including more
style information, rich semantic information and better visual effects are used
in our empirical studies. The result of first study is shown in Table
\ref{table2}. Compared with these classic realistic stylized methods
\cite{reinhard2001color} and \cite{pitie2005n}, although our method scores
slightly lower in style transfer, the scores of semantic information and visual
effects are greater than the sum of other algorithms. The second study results
are shown in Table \ref{table3}. We compare the results with Neural Style,
CNNMRF and \cite{luan2017deep}, and it turns out that our stylized images have
more style information, more semantic information and better visual effect than
existing methods. These two studies show that our method can produce better
stylized images than existing methods. In particular, our method is
slightly inferior to \cite{reinhard2001color} in terms of more style
information, but our methods are better than these methods in terms of more
semantic information and better visual effects. 

\section{Conclusion}\label{sec:conclusion}

We designed a high-resolution generation network as the model generation network.
By connecting multiple subnets in parallel and repeating multi-scale fusion,
the neural style transfer algorithm using the high-resolution network achieves
good results in photorealistic style transfer  
%
%
with a finer structure and less distortion of images. We
conducted extensive experiments and empirical studies to evaluate the proposal.
The experimental results show that compared with existing methods, the
stylized output image by our algorithm has better visual effects and faster
generation speed.


Although our algorithm can achieve better results more quickly than existing
methods, it is still unable to transfer photorealistic style in real-time. We
plan to use the high-resolution generation network trained on the big data set
as the loss network instead of the generation network. Due to the
characteristics of the high-resolution network, it may be able to extract the
feature information of the image well compared to the pre-trained VGG network.
In addition, although our method can perform a good image style transfer on the
whole picture, we cannot transfer the style of a specific thing in the image.
In the future, we plan to use the instance segmentation to realize the instance
image style transfer.

\acks{We thank Leon Gatys, Fujun Luan, Yijun Li, Ke Sun, Justin Johnson for
their great work as well as the reviewers for their valuable discussions. This
work was supported in part by the Key Research and Development Program of
Hainan Province under grant No. ZDYF2017010, the National Natural Science
Foundation of China under grant No. 61562019, 61379047, 60903052, and grants
from State Key Laboratory of Marine Resource Utilization in South China Sea and
Key Laboratory of Big Data and Smart Services of Hainan Province.}

\bibliography{acml19}

\appendix
\section{More Results}\label{apd:first}
\begin{figure}[bt]
	\begin{center}
		\subfigure[Content]{
			\begin{minipage}[t]{0.32\linewidth}
				\centering
				\includegraphics[width=1.8in,height=1.3in]{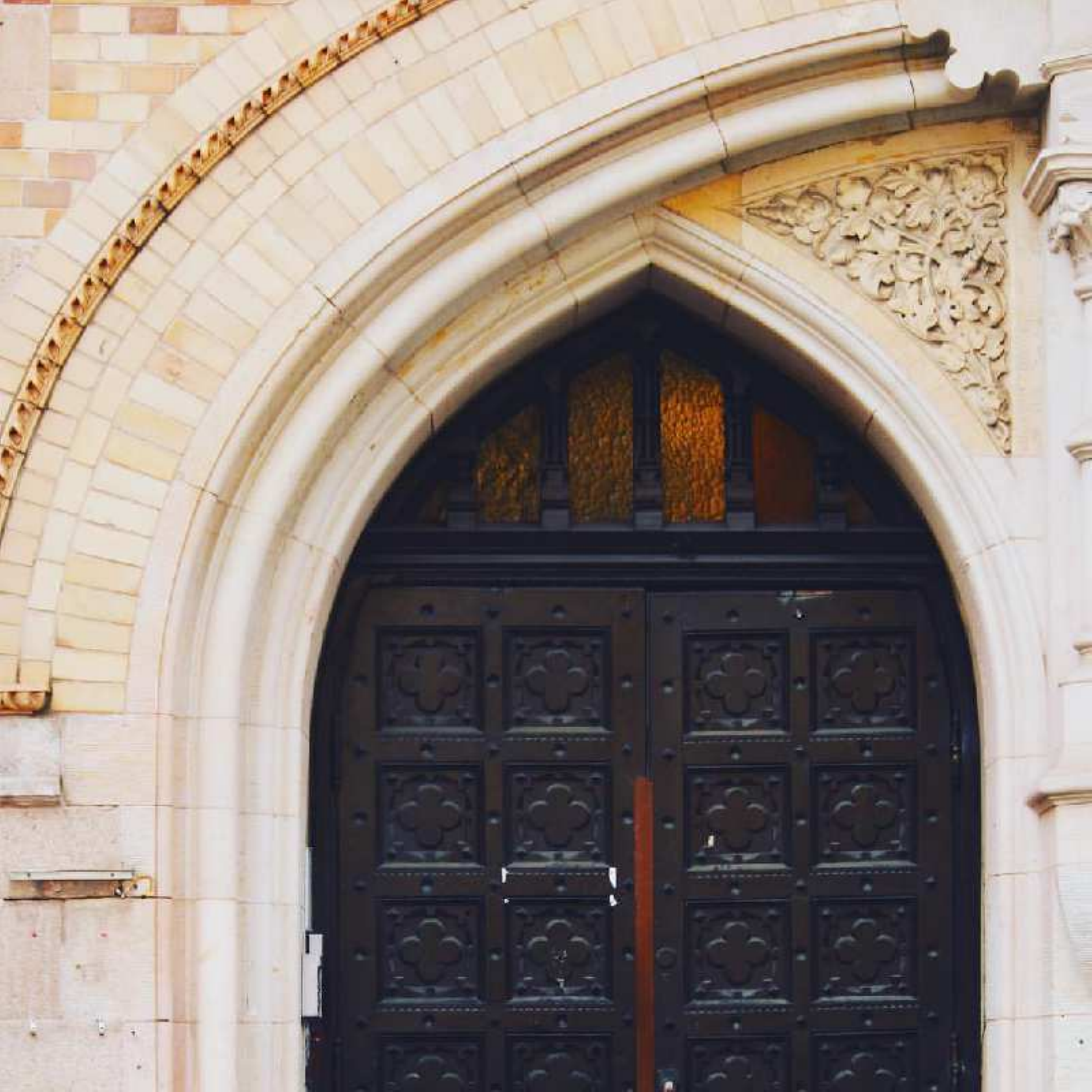}
			\end{minipage}
		}%
		\subfigure[Style]{
			\begin{minipage}[t]{0.32\linewidth}
				\centering
				\includegraphics[width=1.8in,height=1.3in]{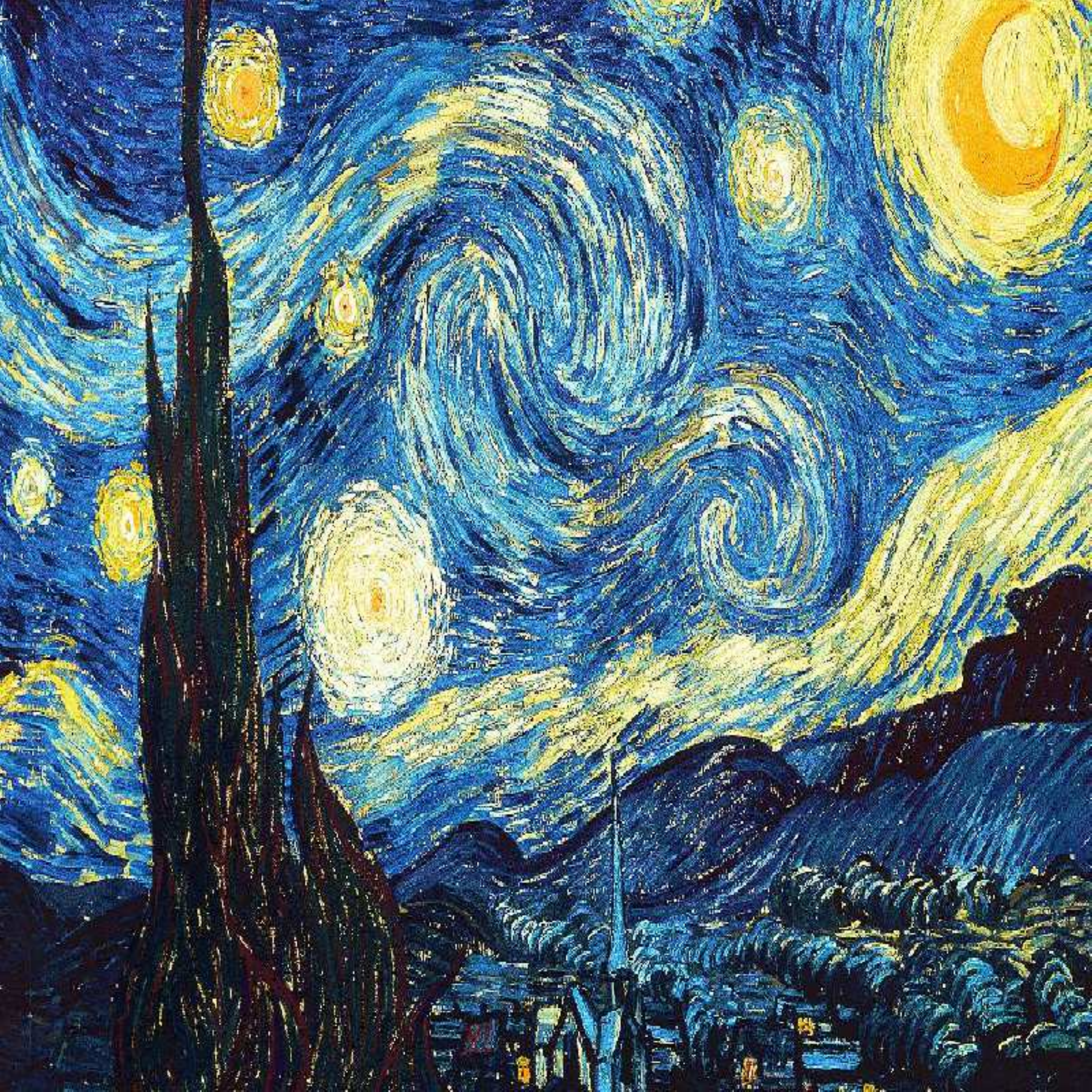}
			\end{minipage}
		}%
		\subfigure[Output]{
			\begin{minipage}[t]{0.32\linewidth}
				\centering
				\includegraphics[width=1.8in,height=1.3in]{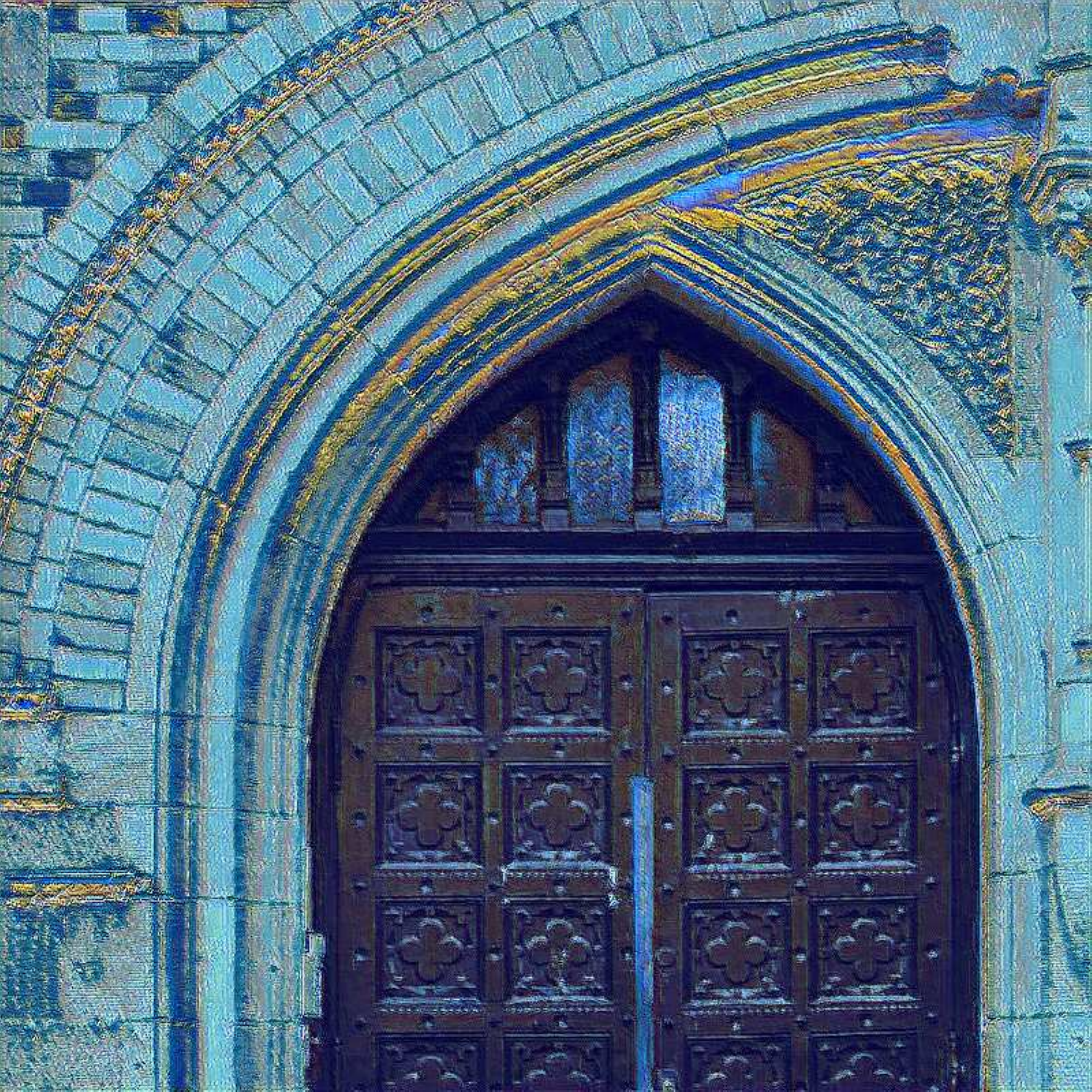}
			\end{minipage}
		}%
		
		\subfigure[Content]{
			\begin{minipage}[t]{0.32\linewidth}
				\centering
				\includegraphics[width=1.8in,height=1.3in]{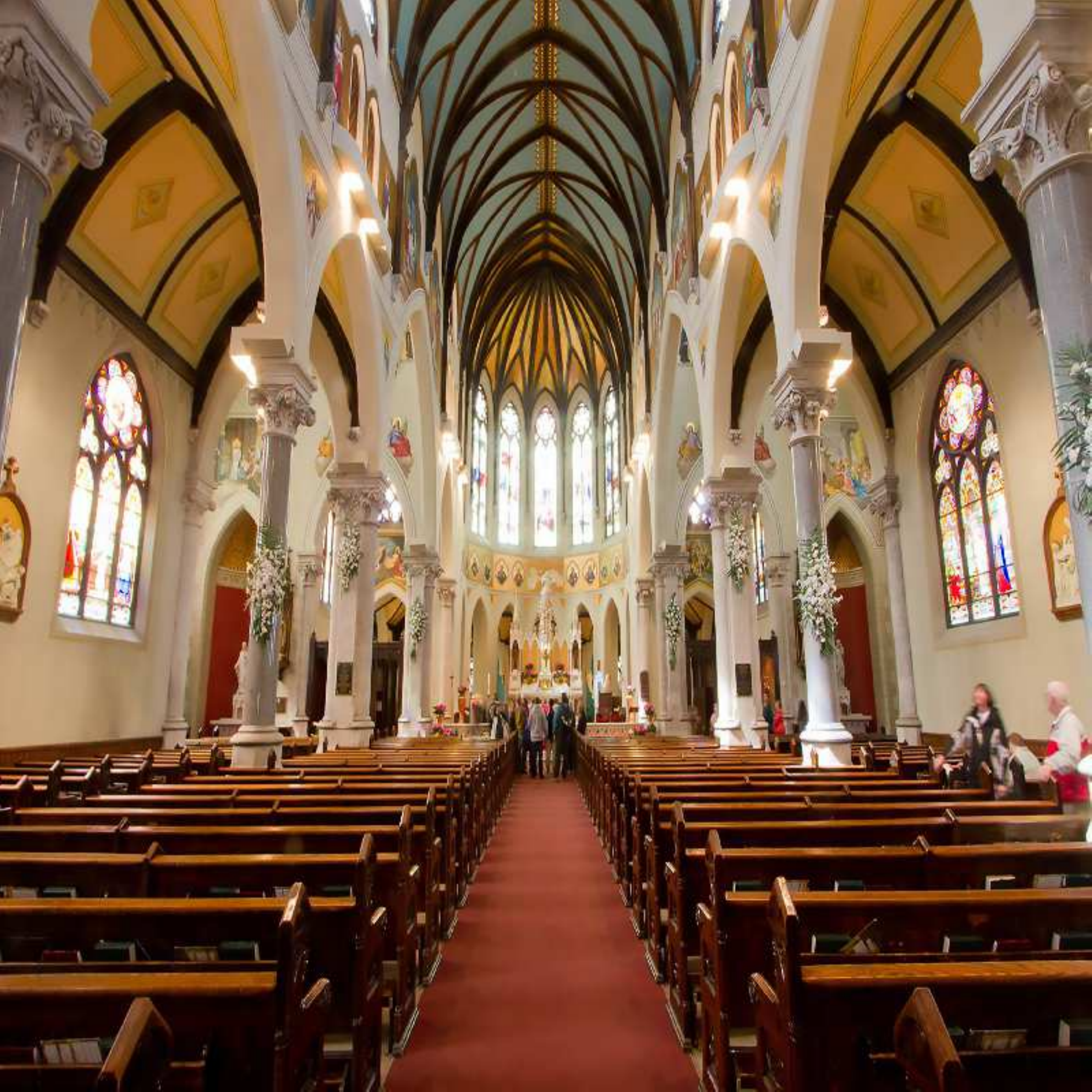}
			\end{minipage}
		}%
		\subfigure[Style]{
			\begin{minipage}[t]{0.32\linewidth}
				\centering
				\includegraphics[width=1.8in,height=1.3in]{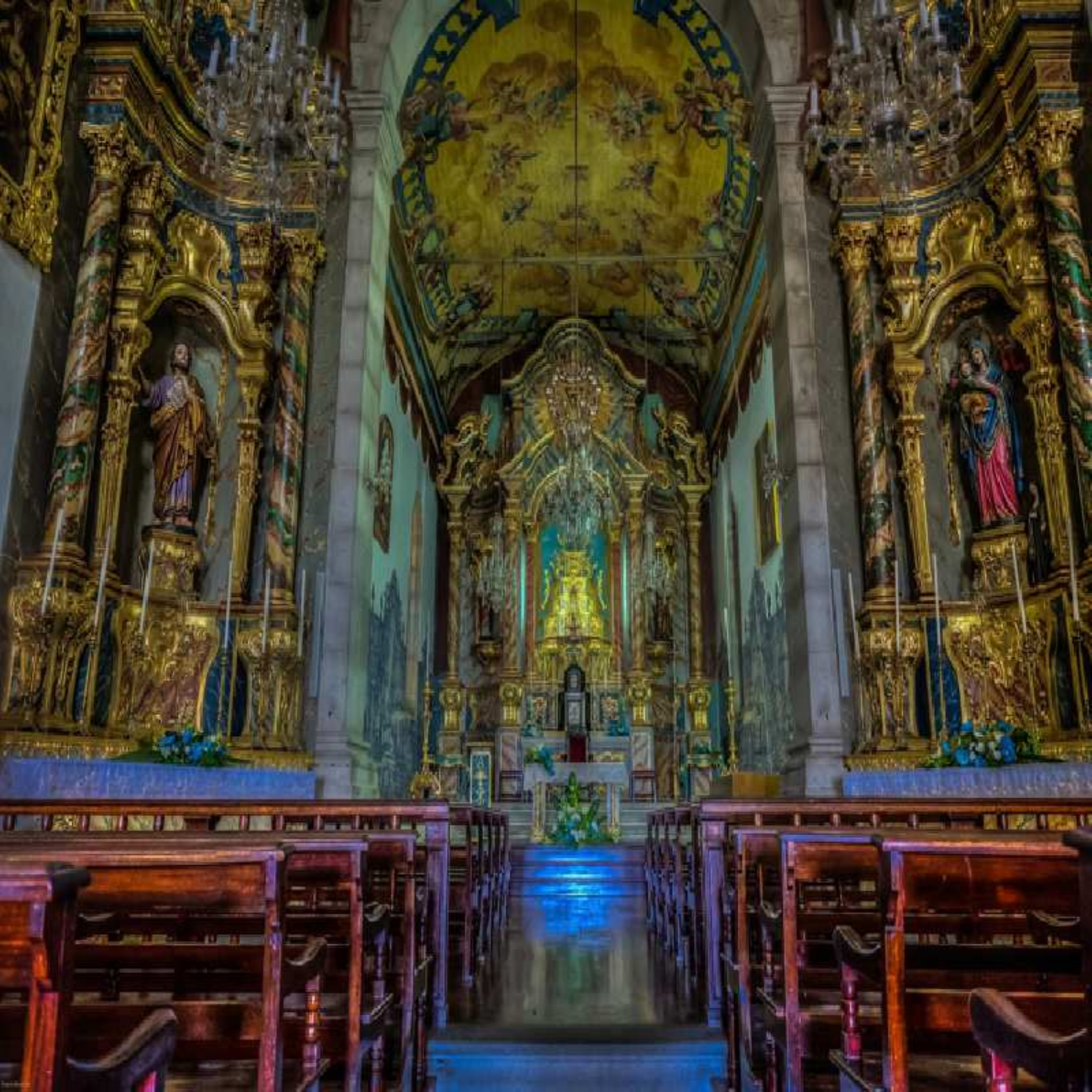}
			\end{minipage}
		}%
		\subfigure[Output]{
			\begin{minipage}[t]{0.32\linewidth}
				\centering
				\includegraphics[width=1.8in,height=1.3in]{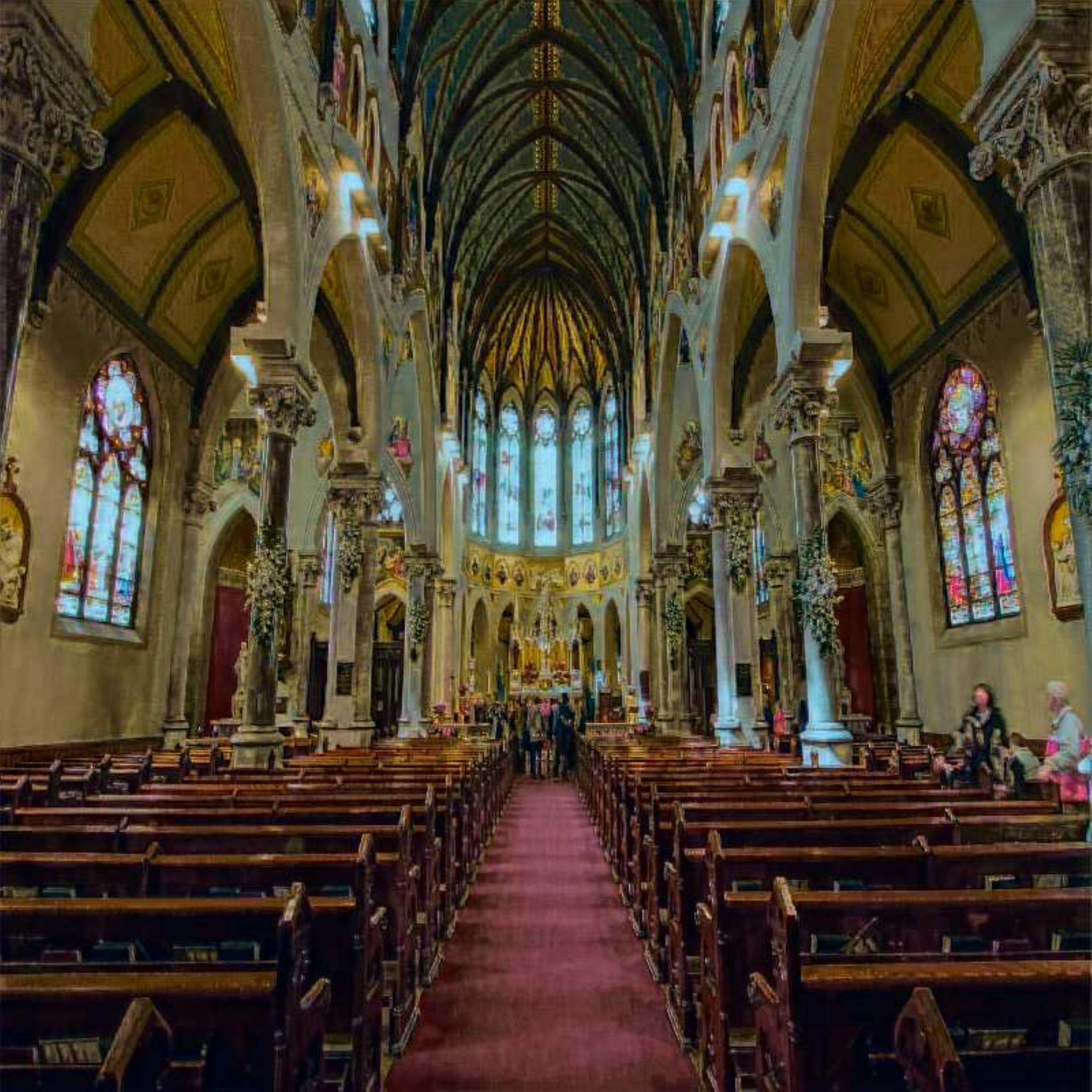}
			\end{minipage}
		}%
	
		\subfigure[Content]{
			\begin{minipage}[t]{0.32\linewidth}
				\centering
				\includegraphics[width=1.8in,height=1.3in]{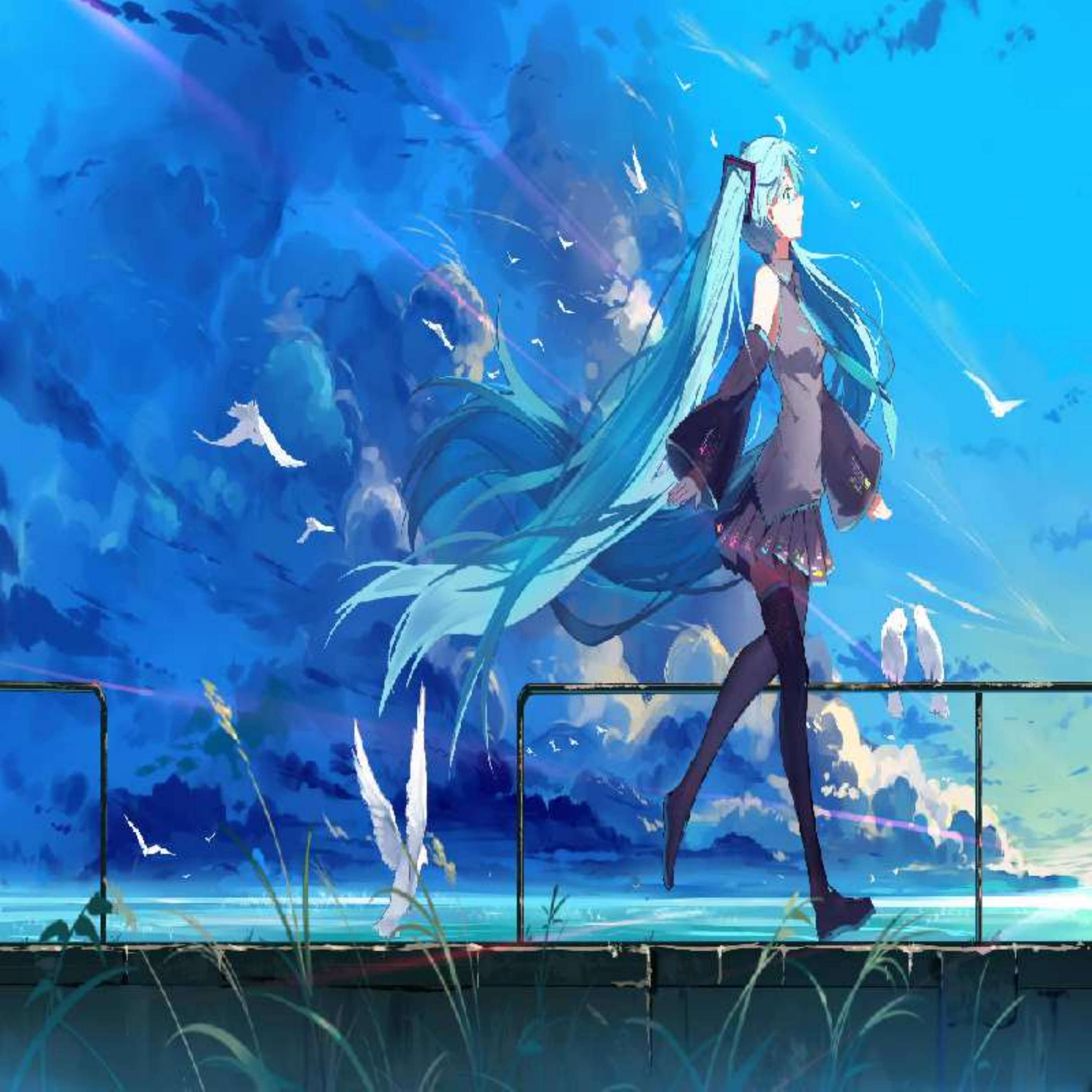}
			\end{minipage}
		}%
		\subfigure[Style]{
			\begin{minipage}[t]{0.32\linewidth}
				\centering
				\includegraphics[width=1.8in,height=1.3in]{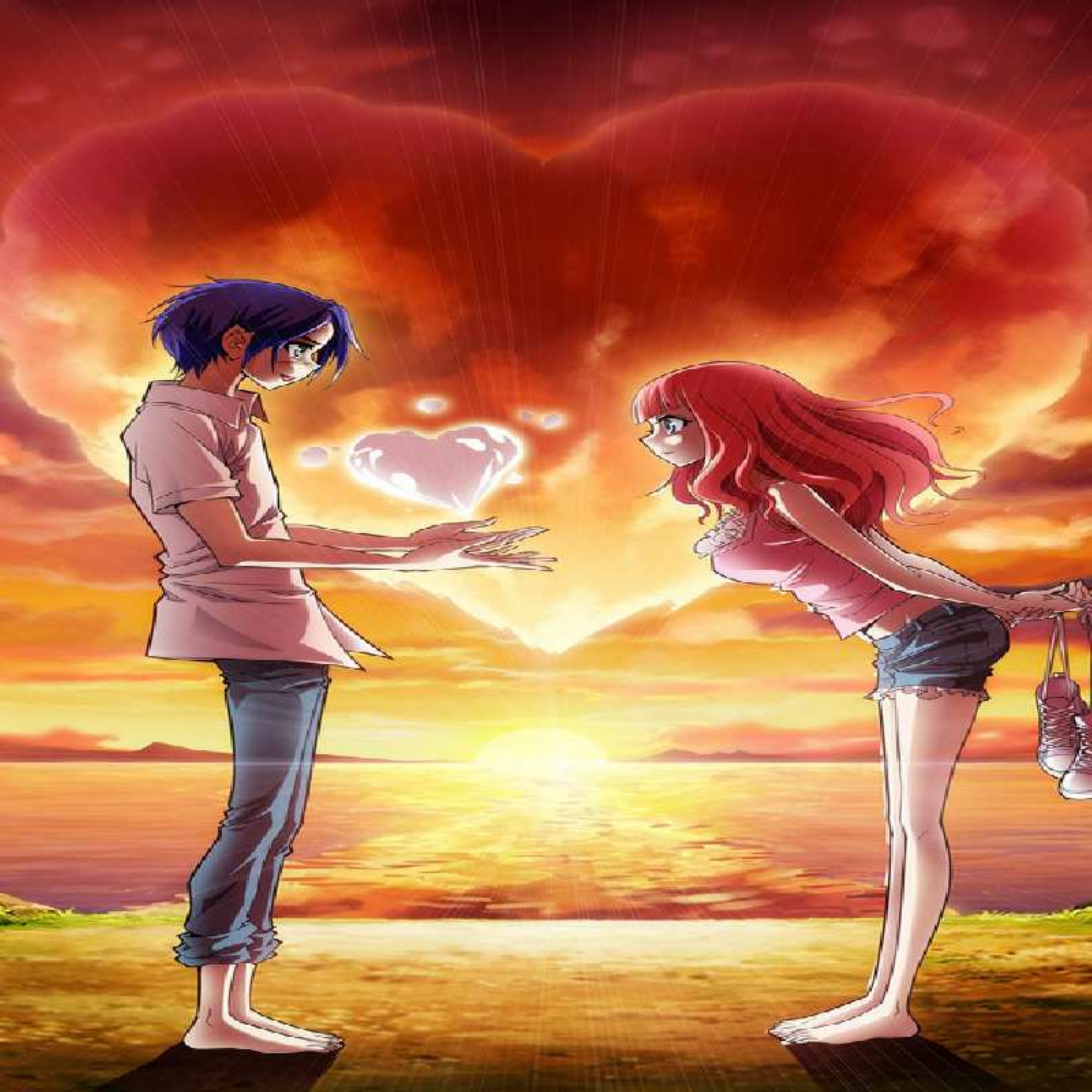}
			\end{minipage}
		}%
		\subfigure[Output]{
			\begin{minipage}[t]{0.32\linewidth}
				\centering
				\includegraphics[width=1.8in,height=1.3in]{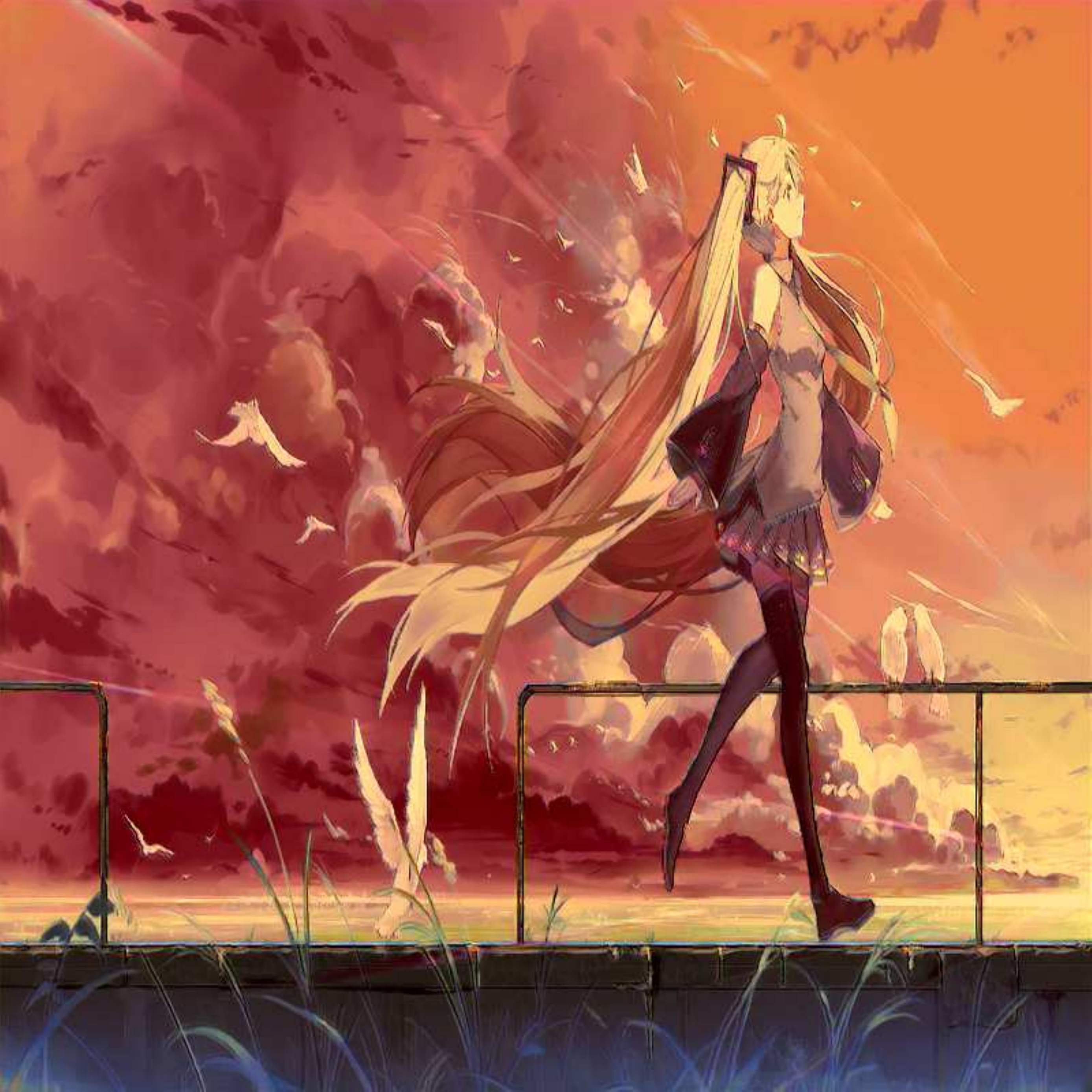}
			\end{minipage}
		}%
	
		\subfigure[Content]{
			\begin{minipage}[t]{0.32\linewidth}
				\centering
				\includegraphics[width=1.8in,height=1.3in]{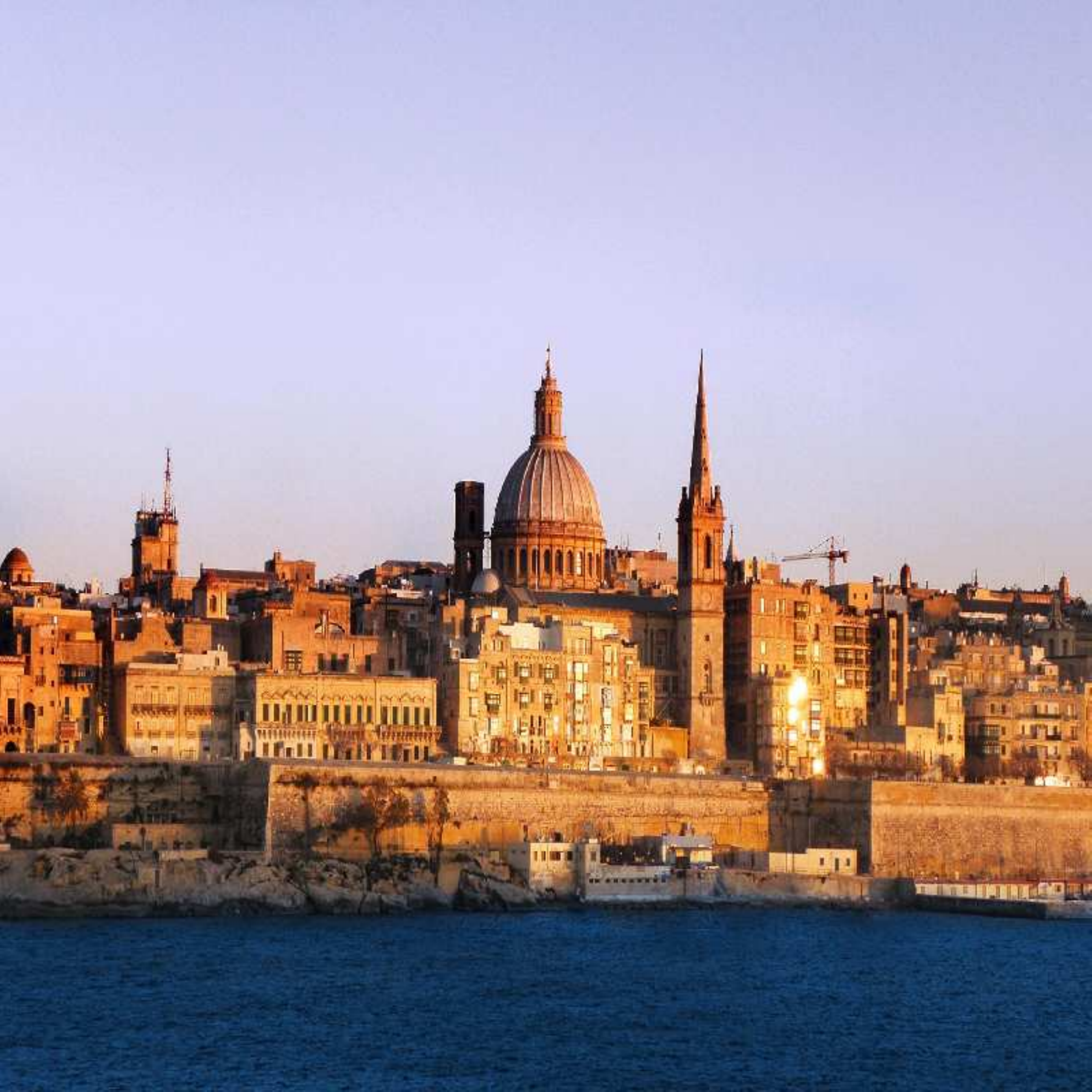}
			\end{minipage}
		}%
		\subfigure[Style]{
			\begin{minipage}[t]{0.32\linewidth}
				\centering
				\includegraphics[width=1.8in,height=1.3in]{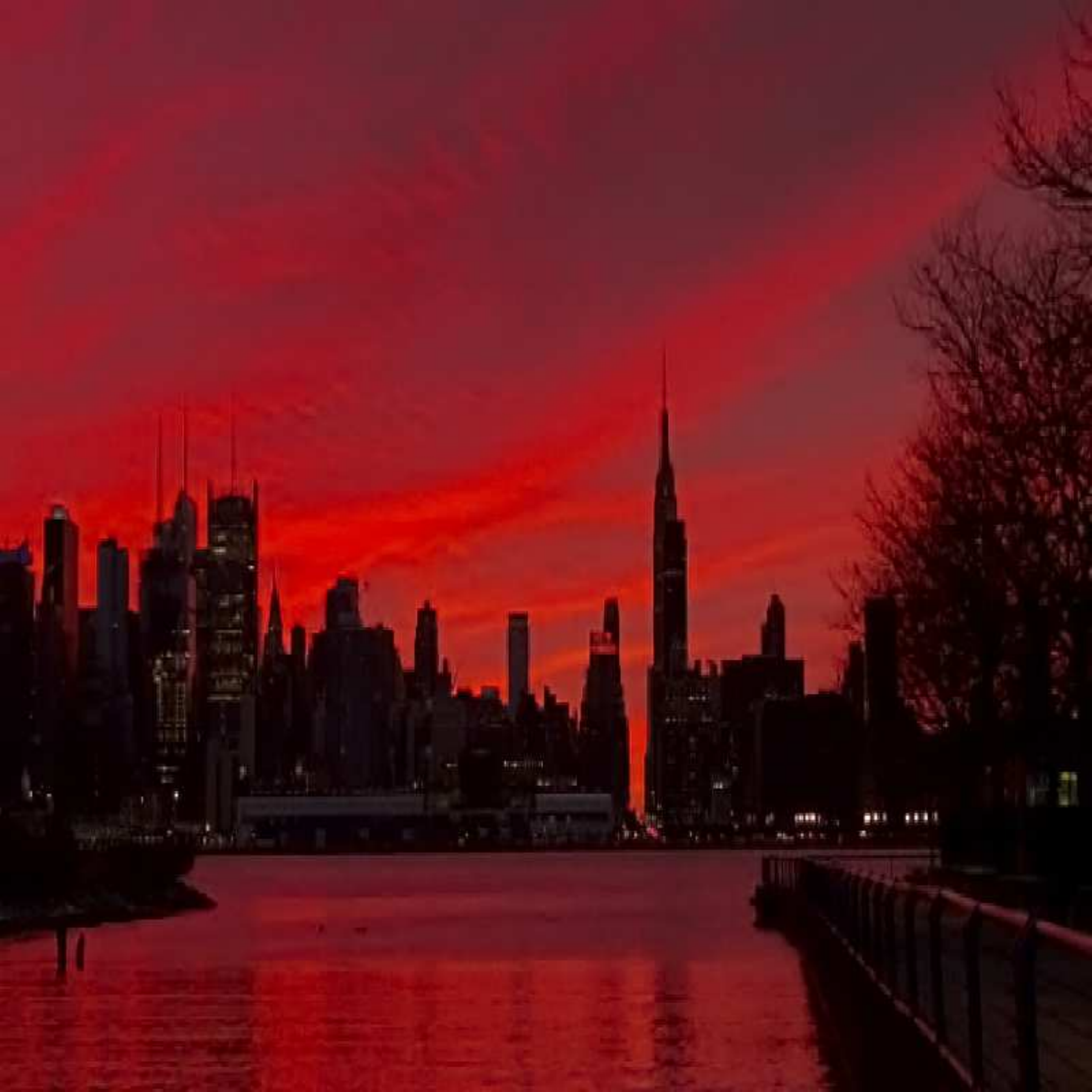}
			\end{minipage}
		}%
		\subfigure[Output]{
			\begin{minipage}[t]{0.32\linewidth}
				\centering
				\includegraphics[width=1.8in,height=1.3in]{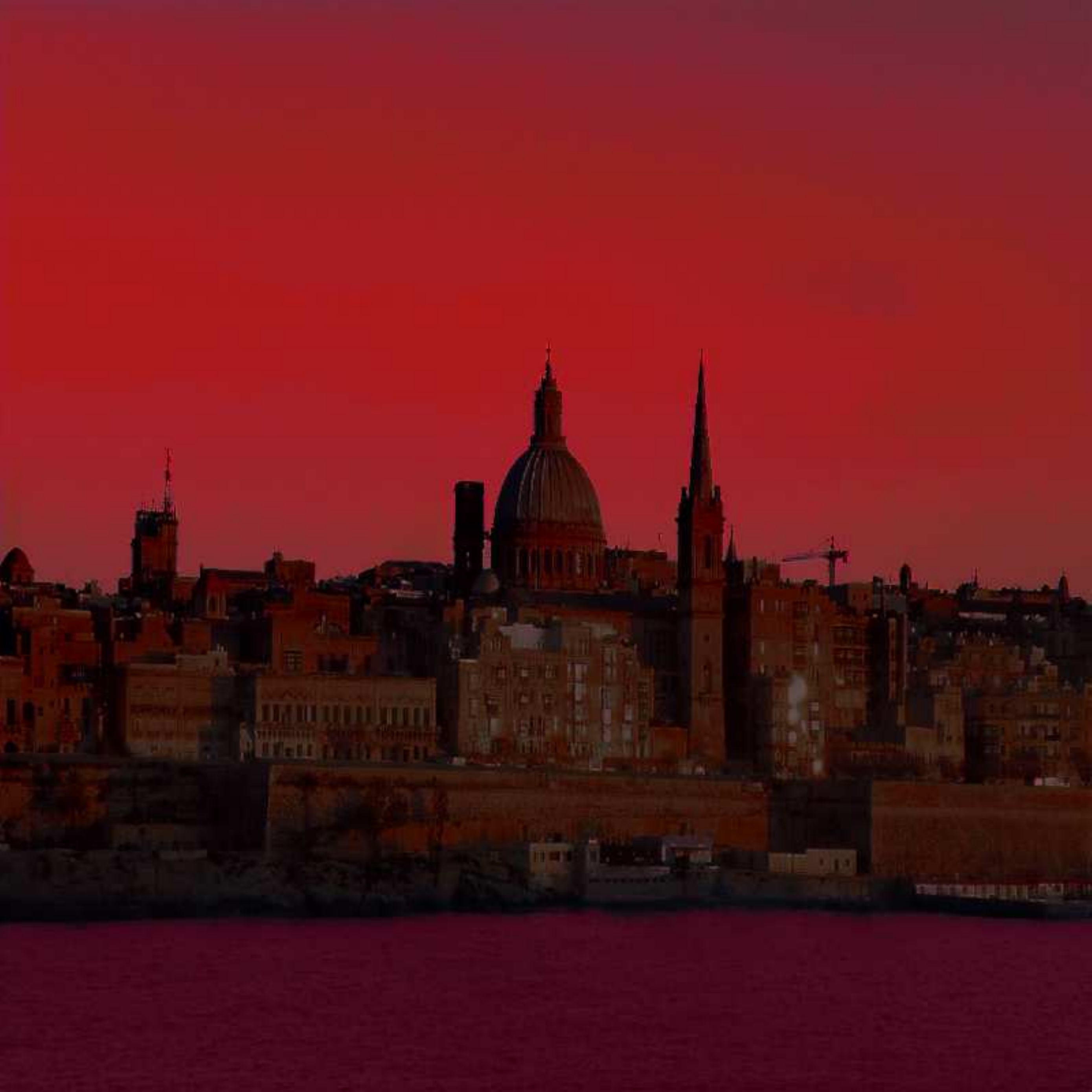}
			\end{minipage}
		}%
	
		\subfigure[Content]	{
			\begin{minipage}[t]{0.32\linewidth}
				\centering
				\includegraphics[width=1.8in,height=1.3in]{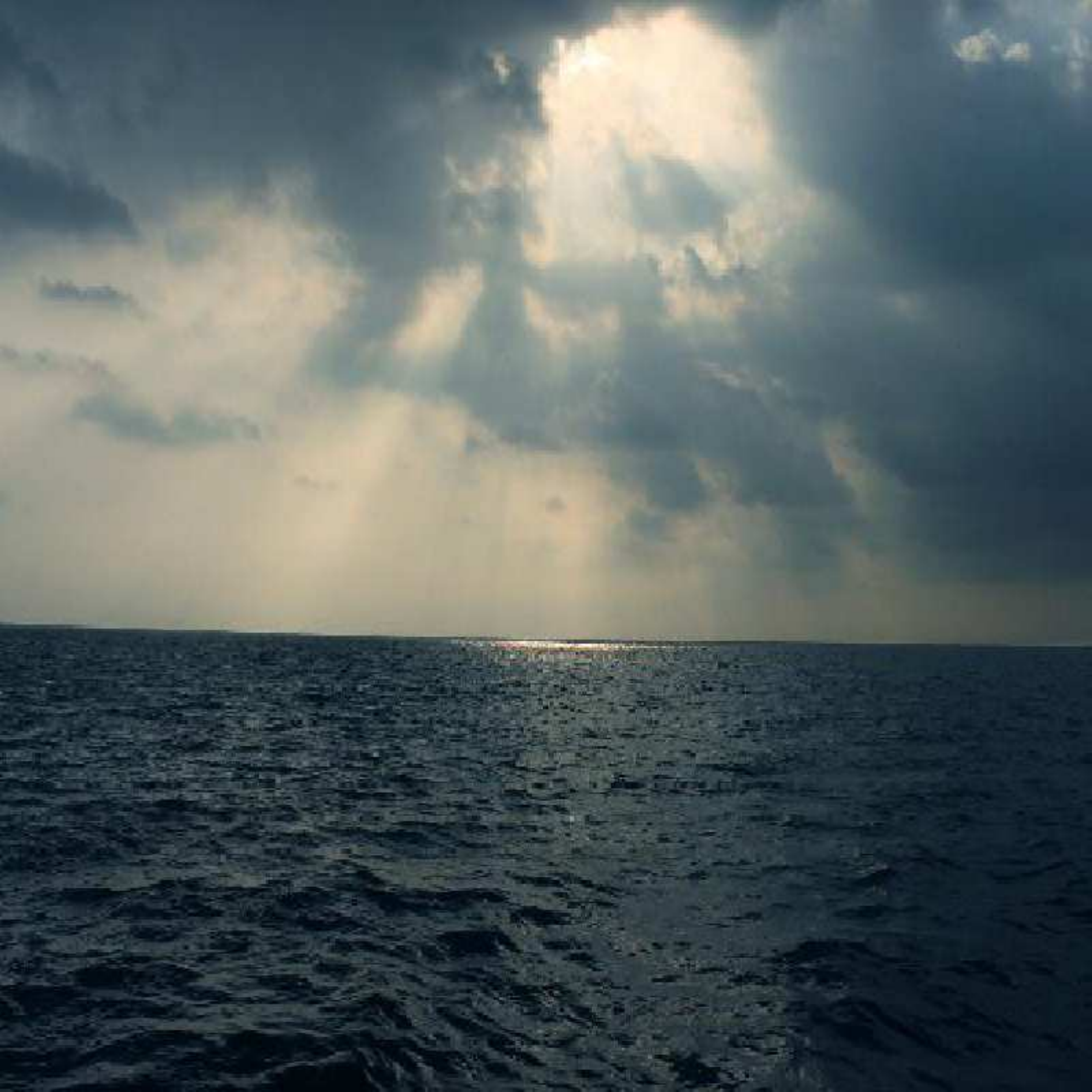}
			\end{minipage}
		}%
		\subfigure[Style]{
			\begin{minipage}[t]{0.32\linewidth}
				\centering
				\includegraphics[width=1.8in,height=1.3in]{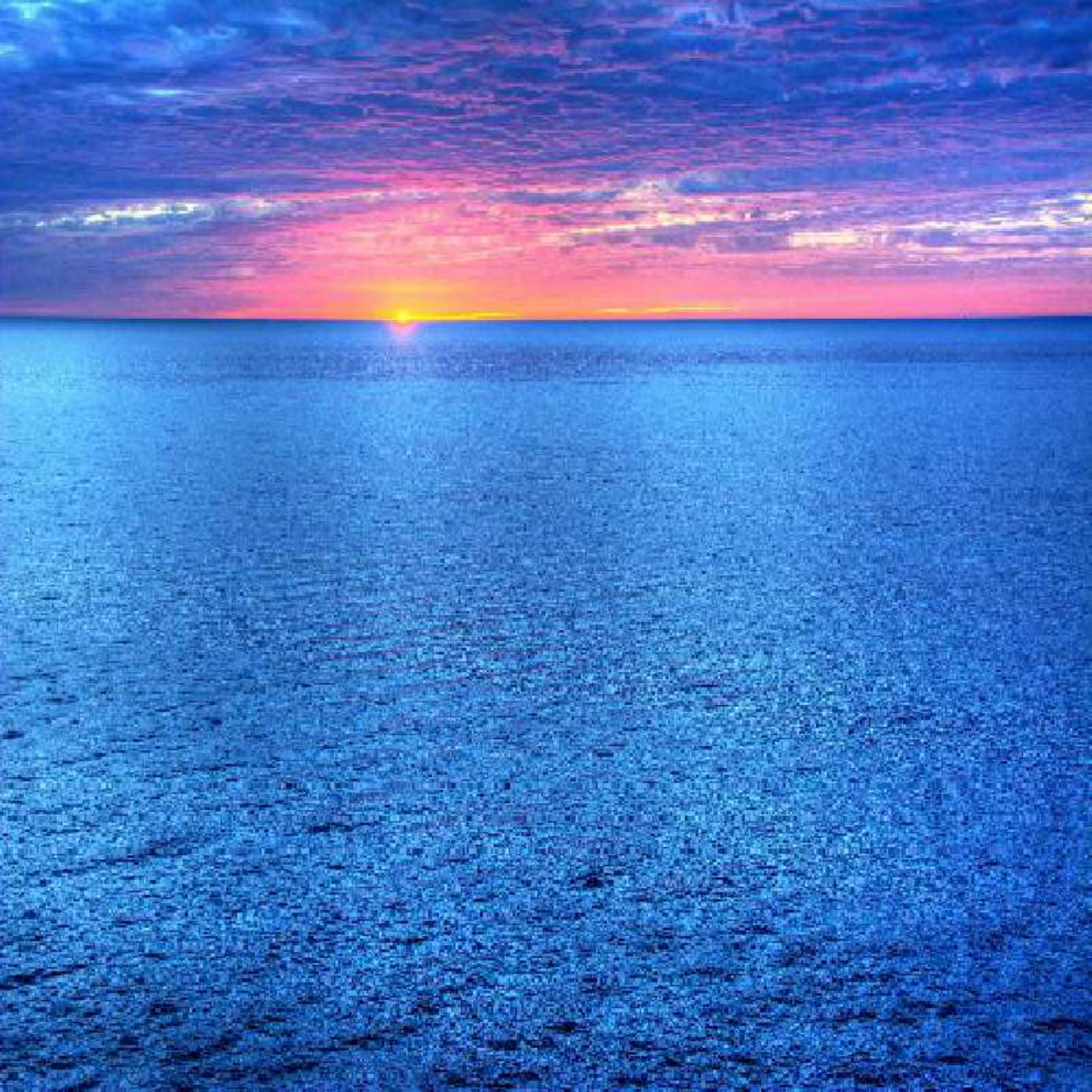}
			\end{minipage}
		}%
		\subfigure[Output]{
			\begin{minipage}[t]{0.32\linewidth}
				\centering
				\includegraphics[width=1.8in,height=1.3in]{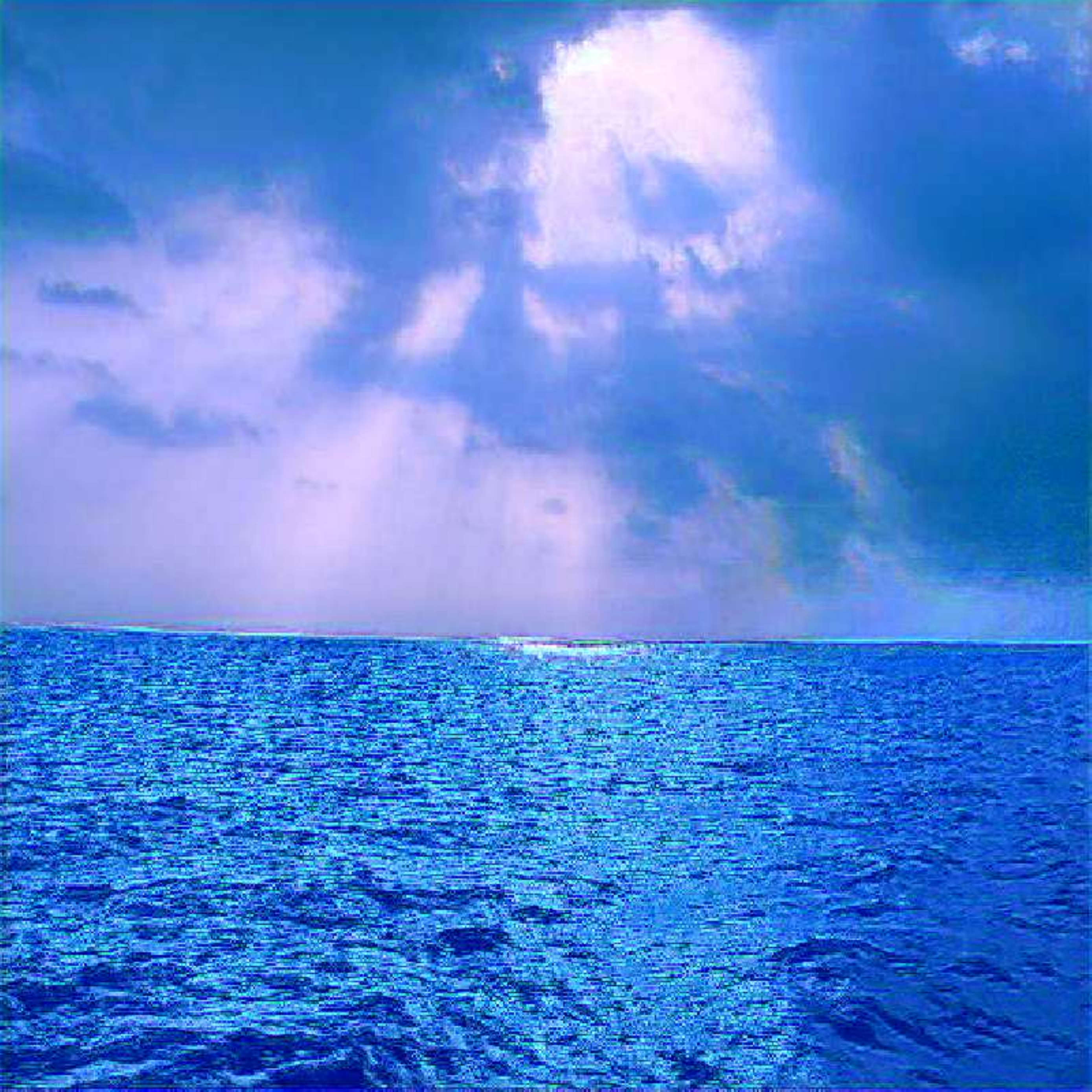}
			\end{minipage}
		}%
        \caption{More results}        
        \label{fig:moreresult}
	\end{center}
\end{figure}

\begin{figure}[htbp]
	\centering
	\begin{minipage}[t]{1\textwidth}
		\centering
		\includegraphics[width=6in,height=0.8in]{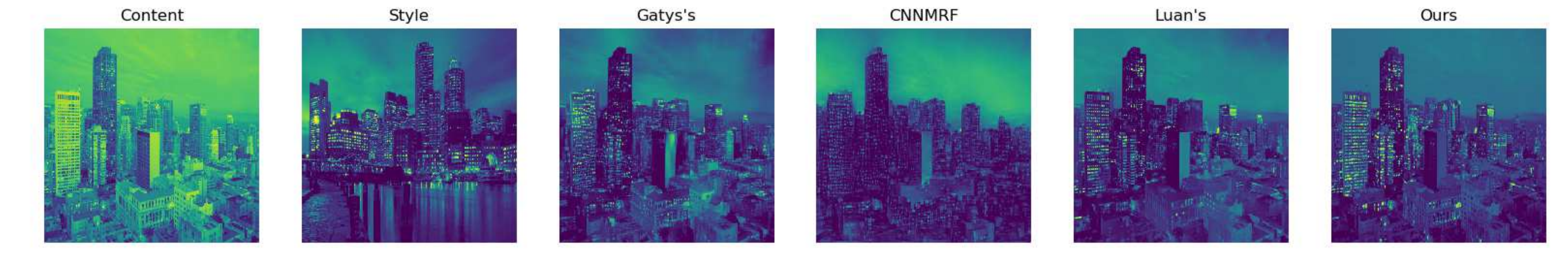}
	\end{minipage}
	\quad
	\begin{minipage}[t]{1\textwidth}
		\centering
		\includegraphics[width=6in,height=0.8in]{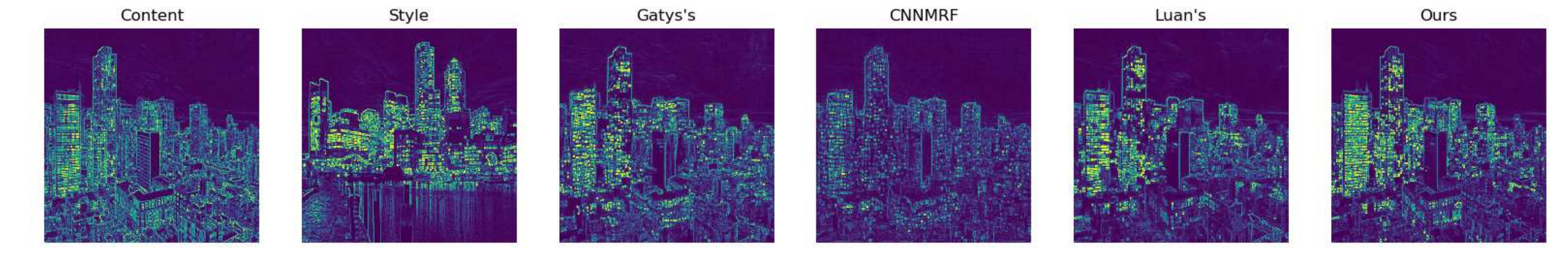}
	\end{minipage}
	\quad
	\begin{minipage}[t]{1\textwidth}
		\centering
		\includegraphics[width=6in,height=0.8in]{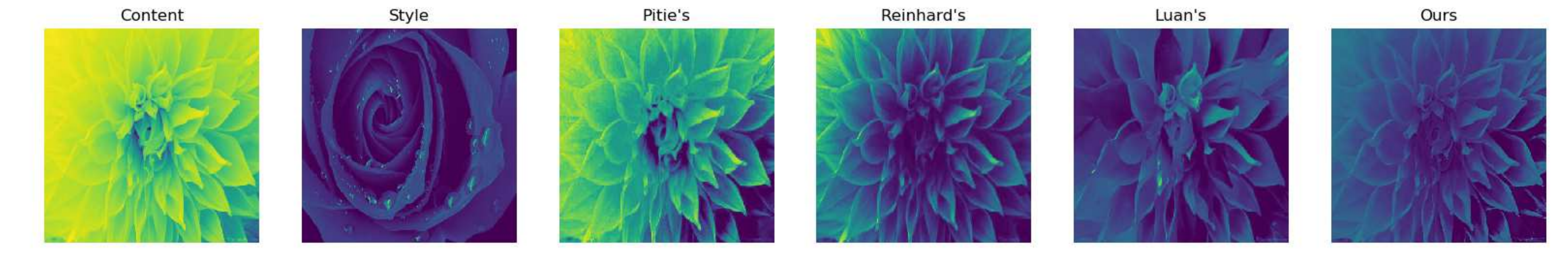}
	\end{minipage}
	\quad
	\begin{minipage}[t]{1\textwidth}
		\centering
		\includegraphics[width=6in,height=0.8in]{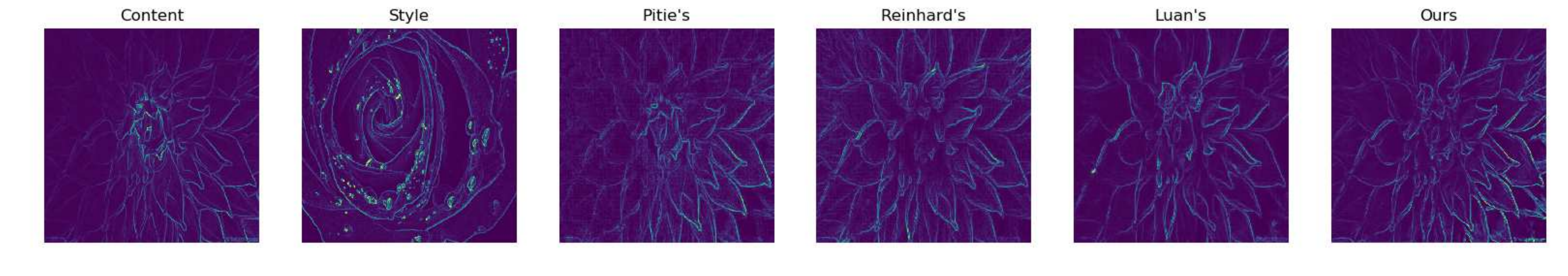}
	\end{minipage}
	\quad
	\begin{minipage}[t]{1\textwidth}
		\centering
		\includegraphics[width=6in,height=0.8in]{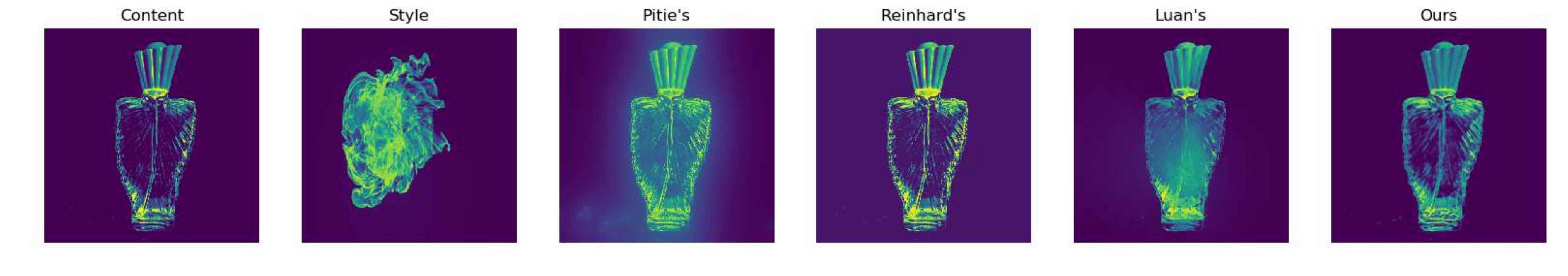}
	\end{minipage}
	\quad
	\begin{minipage}[t]{1\textwidth}
		\centering
		\includegraphics[width=6in,height=0.8in]{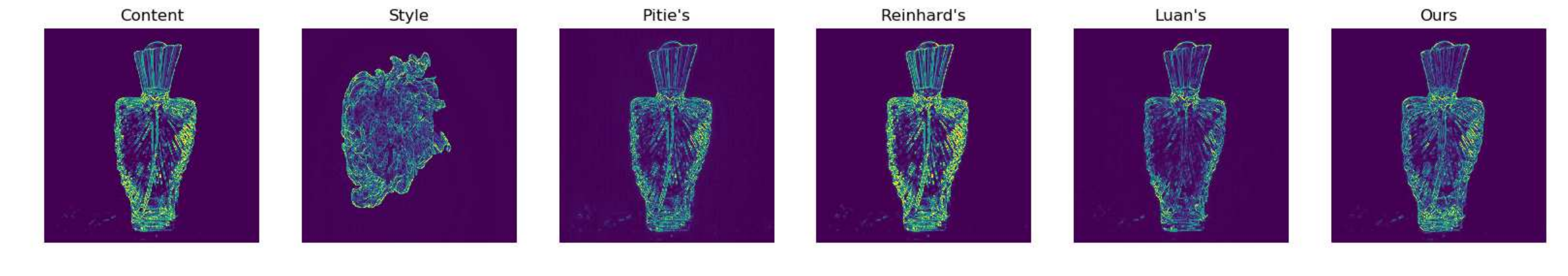}
	\end{minipage}
	\quad
	\begin{minipage}[t]{1\textwidth}
		\centering
		\includegraphics[width=6in,height=0.8in]{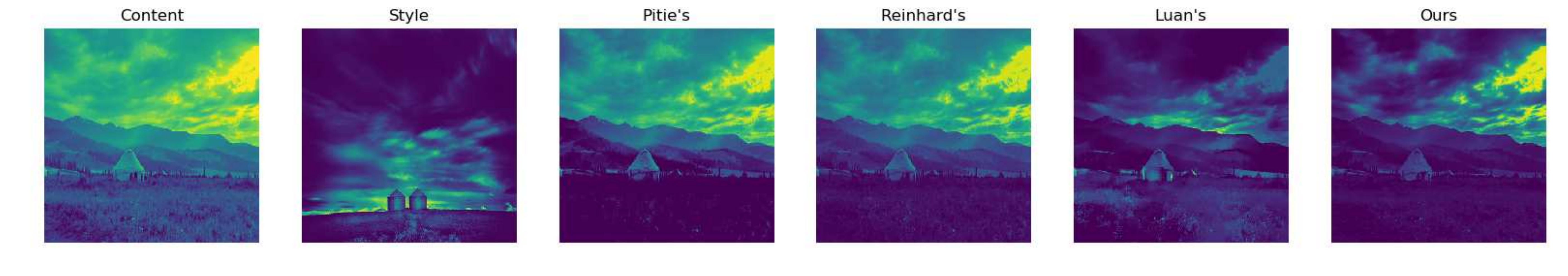}
	\end{minipage}
	\quad
	\begin{minipage}[t]{1\textwidth}
		\centering
		\includegraphics[width=6in,height=0.8in]{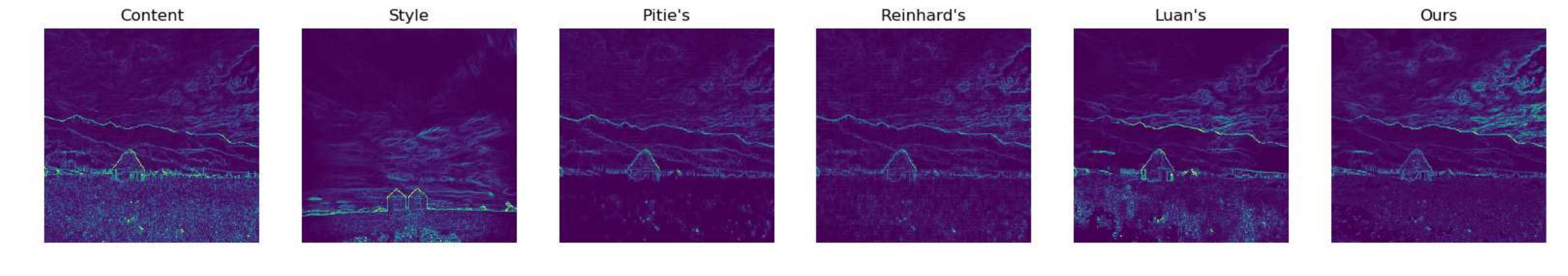}
	\end{minipage}
	\quad
	\begin{minipage}[t]{1\textwidth}
		\centering
		\includegraphics[width=6in,height=0.8in]{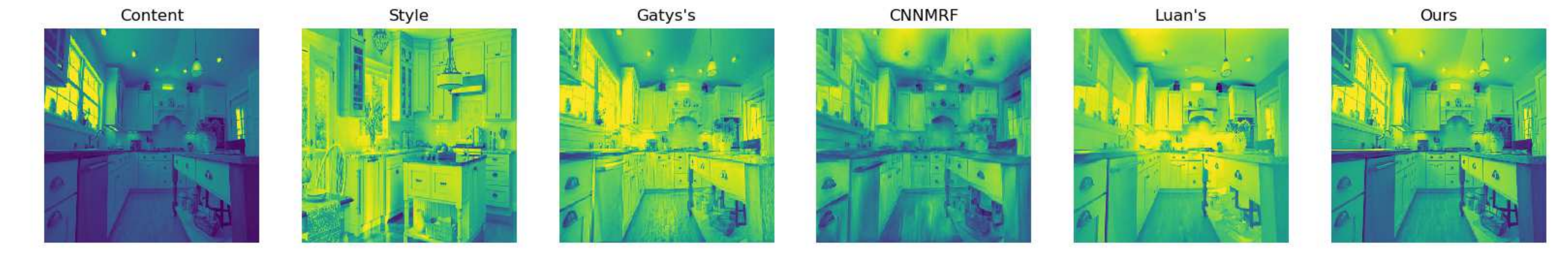}
	\end{minipage}
	\quad
	\begin{minipage}[t]{1\textwidth}
		\centering
		\includegraphics[width=6in,height=0.8in]{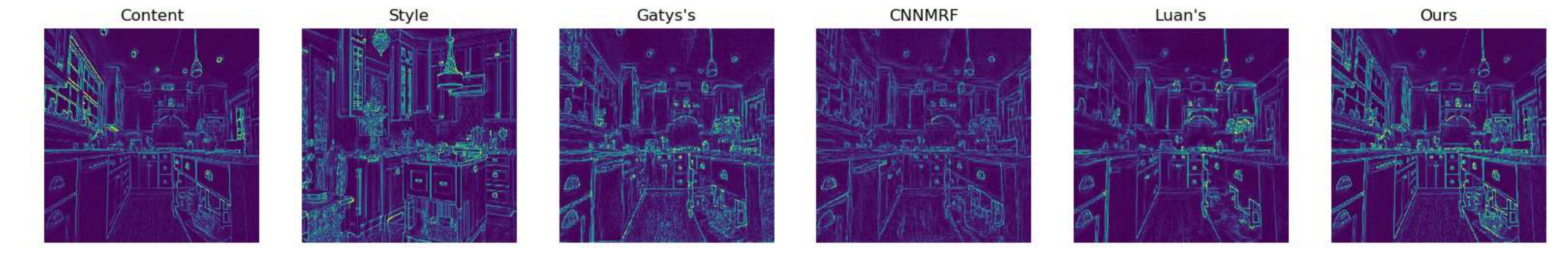}
	\end{minipage}
        \caption{Grayscale and contour images}
        \label{fig:grayscale}
\end{figure}

More results of our approach can be referred to
Figure~\ref{fig:moreresult}. We also compare the contour of resulting images
produced by different solutions through image graying, as shown in
Figure~\ref{fig:grayscale}. By using the sobel operator to extract the contour
of the resulting images and comparing them with those of the content images, we
can observe that the images after stylization based on our approach preserve
more semantic information and have a finer structure than existing solutions.

\end{document}